\documentclass[11pt]{article}
\usepackage{mathrsfs,amsmath,amsfonts,amssymb,bm,bbm,dsfont,pifont,amscd,
amsthm,stmaryrd,euscript,color}
\usepackage{algorithmic}
\usepackage{algorithm}
\usepackage{tikz}
\newcommand{\eu}{\EuScript}

\newcommand{\Cal}{\mathcal}

\newlength{\myleftmargin}
\usepackage{amsmath}
\usepackage{amssymb}
\DeclareSymbolFontAlphabet{\Bbb}{AMSb}

\newtheorem*{theorem*}{Theorem}
\newtheorem{theorem}{Theorem}[section]
\newtheorem{lemma}[theorem]{Lemma}
\newtheorem{proposition}[theorem]{Proposition}
\newtheorem{corollary}[theorem]{Corollary}

\newtheorem{claim}[theorem]{Claim}

\theoremstyle{definition}
\newtheorem{definition}[theorem]{Definition}

\newcommand{\atob}[2]{\emph{#1)} $\Rightarrow$ \emph{#2)}.} 
\newcommand{\ada}[1]{\emph{#1).}} 

\newenvironment{proofof}[1]{\noindent{\bf Proof of #1:}}{\qed\medskip}

\newenvironment{assumption}[1]{\medskip\noindent{\bf Assumption #1.}}{\medskip}
\newcommand{\assx}[1]{\emph{Assumption #1}}

\newtheorem{innercustomthm}{Theorem}
\newenvironment{customthm}[1]
  {\renewcommand\theinnercustomthm{#1}\innercustomthm}
  {\endinnercustomthm}

\newlength{\fixboxwidth}
\newcommand{\ca}[1]{{\cal #1}}

\newcommand{\qspace}{\, , \qquad\qquad}

\newcommand{\R}{\mathbb{R}}
\newcommand{\Rd}{{\mathbb{R}^d}}

\newcommand{\E}{\mathbb{E}}

\renewcommand{\a}{\alpha}

\newcommand{\g}{\gamma}

\renewcommand{\d}{\delta}
\newcommand{\D}{\Delta}
\newcommand{\e}{\varepsilon}
\newcommand{\eps}{\epsilon}
\newcommand{\z}{\zeta}

\newcommand{\vt}{\vartheta}
\renewcommand{\k}{\kappa}
\newcommand{\lb}{\lambda}

\renewcommand{\r}{\rho}
\newcommand{\vr}{\varrho}
\newcommand{\s}{\sigma}
\newcommand{\vs}{\varsigma}
\renewcommand{\t}{\tau}

\newcommand{\om}{\omega}

\DeclareMathOperator{\diam}{diam}
\DeclareMathOperator{\inrad}{inrad}
\DeclareMathOperator{\supp}{supp}

\DeclareMathOperator{\vol}{vol}
\DeclareMathOperator{\vold}{vol_d}

\newcommand{\symdif}{\vartriangle}
\newcommand{\Leq}{\,\,\leq\,\,}

\newcommand{\eins}{\boldsymbol{1}}

\newcommand{\snorm}[1]{\Vert #1 \Vert}

\newcommand{\inorm}[1]{\Vert #1 \Vert_\infty}
\newcommand{\tnorm}[1]{\Vert #1 \Vert_2}
\newcommand{\snorme}{{\snorm\cdot}}

\newcommand{\Lx}[2]{{L_{#1}(#2)}}

\newcommand{\tridia}[6]{
\setlength{\unitlength}{1ex}
\begin{picture}(60,20)
\put(20,14){\makebox(9,2)[rt]{$#1$}}
\put(51,14){\makebox(9,2)[lt]{$#2$}}
\put(35.5,2){\makebox(9,2){$#3$}}
\put(30,15){\vector(1,0){20}}
\put(30,14){\vector(1,-1){9}}
\put(41,5){\vector(1,1){9}}
\put(37,16){\makebox(6,2){$#4$}}
\put(27.5,8){\makebox(6,2)[rt]{$#5$}}
\put(47.5,8){\makebox(6,2)[lt]{$#6$}}
\end{picture}}

\newcommand{\quadiann}[8]{
\setlength{\unitlength}{1ex}
\begin{picture}(60,20)
\put(20,15){\makebox(9,2)[rt]{$#1$}}
\put(45,15){\makebox(9,2)[lt]{$#2$}}
\put(20,2){\makebox(9,2)[rt]{$#3$}}
\put(45,2){\makebox(9,2)[lt]{$#4$}}
\put(30,16){\vector(1,0){14}}
\put(25,5){\vector(0,1){9}}
\put(49,5){\vector(0,1){9}}
\put(30,3){\vector(1,0){14}}
\put(34,17){\makebox(6,2){$#5$}}
\put(18,8.5){\makebox(6,2)[rt]{$#6$}}
\put(50,8.5){\makebox(6,2)[lt]{$#7$}}
\put(34,0){\makebox(6,2){$#8$}}
\end{picture}}

\newcommand{\dsuf}{\d_{\mathrm{up},n}}
\newcommand{\dnec}{\d_{\mathrm{low},n}}

\newcommand{\rsuf}{\r_{\mathrm{up},n}}
\newcommand{\rnec}{\r_{\mathrm{low},n}}

\newcommand{\riin}{\r_{\mathrm{ii},n}}
\newcommand{\diin}{\d_{\mathrm{ii},n}}

\newcommand{\rsufnu}{\r_{\mathrm{up},\om, n}}

\newcommand{\hpd}{h_{P,\d}}
\newcommand{\hdd}{h_{D,\d}}
\newcommand{\hdpd}{h_{D',\d}}

\newcommand{\pdex}[1]{^{+#1}}
\newcommand{\pde}{^{+\d}}

\newcommand{\mde}{^{-\d}}

\newcommand{\pds}{^{+\s}}
\newcommand{\pdds}{^{+2\s}}
\newcommand{\mdds}{^{-2\s}}

\newcommand{\rmax}{\r_{\mathrm{max}}}

\newcommand{\rds}{{\r^*_D}}

\newcommand{\rs}{{\r^*}}

\newcommand{\rso}{{\r_1^*}}
\newcommand{\rst}{{\r_2^*}}

\newcommand{\rss}{{\r^{**}}}

\newcommand{\rsso}{{\r_1^{**}}}

\newcommand{\rls}{{\r_{*}}}
\newcommand{\eda}{{\e^\dagger}}
\newcommand{\rda}{{\r^\dagger}}

\newcommand{\rdao}{{\r_1^\dagger}}

\newcommand{\rdda}{{\r^{\dagger\dagger}}}

\newcommand{\rddao}{{\r_1^{\dagger\dagger}}}

\newcommand{\rout}{{\r_{\mathrm{out}}}}

\newcommand{\routo}{{\r_{1, \mathrm{out}}}}
\newcommand{\routt}{{\r_{2, \mathrm{out}}}}

\newcommand{\routD}{{\r_{D, \mathrm{out}}}}

\newcommand{\routDd}{{\r_{D, \d, \mathrm{out}}}}

\newcommand{\rdcp}{{\r^\ddagger}}
\newcommand{\setcp}{\mathfrak P(h)}
\newcommand{\setsl}{\mathfrak S(h)}

\newcommand{\ts}{{\t^*}}

\newcommand{\psis}{\psi^*}

\newcommand{\dthick}{\d_{\mathrm{thick}}}
\newcommand{\cthick}{c_{\mathrm{thick}}}

\newcommand{\csepl}{\underbar c_{\mathrm{sep}}}
\newcommand{\csepu}{\overline c_{\mathrm{sep}}}

\newcommand{\cflat}{c_{\mathrm{flat}}}
\newcommand{\cbound}{c_{\mathrm{bound}}}

\newcommand{\cc}[1]{\ca C(#1)}
\newcommand{\cct}[1]{\ca C_\t(#1)}
\newcommand{\ccth}[1]{\widehat{\ca C}_\t(#1)}

\newcommand{\ccp}[1]{\ca C_{\mathrm{path}}(#1)}

\newcommand{\lr}{{L_\r}}
\newcommand{\lrp}{L_{\r+2\e}}
\newcommand{\mr}{{M_\r}}
\newcommand{\mrx}[1]{{M_{\r#1}}}
\newcommand{\mrmde}{{M_\r^{-\d}}}
\newcommand{\mrmdex}[1]{{M_{\r#1}^{-\d}}}
\newcommand{\mrpdex}[1]{{M_{\r#1}^{+\d}}}
\newcommand{\mrpde}{{M_\r^{+\d}}}
\newcommand{\mrs}{{M_{\r+\e}^{-\d}}}
\newcommand{\mrl}{{M_{\r-\e}^{+\d}}}

\newcommand{\mx}[1]{{M_{#1}}}
\newcommand{\mri}[1]{{M_{#1,\r}}}
\newcommand{\mrimde}[1]{{M_{#1,\r}^{-\d}}}
\newcommand{\mripde}[1]{{M_{#1,\r}^{+\d}}}

\newcommand{\mxpde}[1]{{M_{#1}\pde}}
\newcommand{\mxmde}[1]{{M_{#1}\mde}}

\newcommand{\mhx}[1]{{\{h\geq #1\}}}
\newcommand{\mhxg}[1]{{\{h> #1\}}}

\newcommand{\mycdot}{\,\cdot\,}

\newcommand{\comparable}{\sqsubset}
\newcommand{\persist}{\sqsubseteq}

\newcommand{\lbd}{\lb^d}
\newcommand{\dP}{\, \mathrm d P}
\newcommand{\dld}{\, \mathrm{d}\lbd}

\newcommand*{\flip}[1]{\scalebox{-1}[1]{\rotatebox[origin = c]{180}{$\mathrm{#1}$}}}

\newcommand{\hdr}[2]{{\flip W (#1, #2)}}
\newcommand{\hdra}[1]{\hdr {#1}{A_1\cup A_2}}
\newcommand{\hdrar}[1]{\hdr {#1}{A_{\r,1}\cup A_{\r,2}}}
\newcommand{\hdrax}[2]{\hdr {#1}{A_{#2,1}\cup A_{#2,2}}}

\newcommand{\hds}[2]{{\mathrm V(#1, #2)}}
\newcommand{\hdsa}[1]{\hds {#1}{A_1, A_2}}
\newcommand{\hdsar}[1]{\hds {#1}{A_{\r,1}, A_{\r,2}}}
\newcommand{\hdsax}[2]{\hds {#1}{A_{#2,1}, A_{#2,2}}}
\newcommand{\hdsb}[1]{\hds {#1}{B_1, B_2}}

\newcommand{\kxd}{k_{x,\d}}
\newcommand{\fxd}{f_{x,\d}}

\newcommand{\kt}[2]{\kappa_{#1}(#2)}
\newcommand{\kto}[1]{\kt 1{#1}}
\newcommand{\kti}[1]{\kt \infty{#1}}
\newcommand{\ktof}{\kt 1{\cdot}}
\newcommand{\ktif}{\kt \infty{\cdot}}

\newcommand{\eul}{\mathrm{e}}

\newcommand{\ccr}[2]{A_{#1,#2}}

\begin{document}
\title{Adaptive Clustering Using Kernel Density Estimators}
\author{Ingo Steinwart\footnote{Corresponding author}\\
University of Stuttgart\\
Faculty 8: Mathematics and Physics\\
Institute for Stochastics and Applications\\
D-70569 Stuttgart Germany \\
\texttt{\small ingo.steinwart@mathematik.uni-stuttgart.de}\\[2ex]
Bharath K. Sriperumbudur\\
Pennsylvania State University\\
Department of Statistics\\
University Park, PA 16802, USA\\
\texttt{\small bks18@psu.edu}\\[2ex]
Philipp Thomann\\
D ONE Solutions AG\\
Sihlfeldstrasse 58\\
8003 Z\"urich\\
\texttt{\small philipp.thomann@d1-solutions.com}
}
\date{}

\maketitle


\begin{abstract}
We  derive and analyze a generic, recursive  algorithm for estimating all splits in a finite cluster tree
as well as  the corresponding clusters. We further 
investigate statistical properties of this generic clustering algorithm when it receives 
level set estimates from a kernel density estimator. In particular, we derive
finite sample guarantees, consistency, rates of convergence, and an adaptive data-driven 
strategy for choosing the kernel bandwidth.
 For these results
we do not need continuity assumptions on the density such as H\"older continuity,
but only require intuitive geometric assumptions of non-parametric nature.
\end{abstract}
\textbf {MSC2010 Subject Classification:} \textbf{Primary:} 62H30; \textbf{Secondary:} 62H12, 62G05 





%
%

\section{Introduction}

A widely acknowledged problem in cluster analysis is 
the definition of a learning goal that describes  
a conceptually and mathematically convincing definition of clusters.
One such definition, which goes back to Hartigan \cite{Hartigan75} and is known as \emph{density-based clustering},
   assumes i.i.d.~data $D=(x_1,\dots,x_n)$  generated by 
some unknown distribution $P$. 
Given some $\r\geq 0$, the clusters of $P$  are then defined to be the connected components of the  level set
$\{h\geq  \r\}$, where $h$ is the density associated with $P$ w.r.t.~the Lebesgue measure. 
%
%
This \emph{single level approach}
has been studied, for example in 
\cite{Hartigan75,CuFr97a,Rigollet07a,MaHeLu09a,RiWa10a}.
However, one of the conceptual drawbacks of the single level approach is that 
different values of $\r$ may lead to different (numbers of) clusters, and there is also 
 no general rule for choosing $\r$, either.
To address this conceptual shortcoming, one often considers the 
so-called  \emph{cluster tree approach} instead,
which tries to consider 
all levels and the corresponding connected components simultaneously.

If the focus 
 lies   on the identification of 
the \emph{hierarchical tree structure}
of the connected components, then one can find a 
variety of articles investigating properties of the cluster tree approach, see 
e.g.~\cite{Hartigan75,Stuetzle03a,ChDa10a,StNu10a,KpLu11a,ChDaKpLu14a,WaLuRi19a} for details.
For example, \cite{ChDa10a} shows, under some assumptions on  $h$, that a modified single linkage 
algorithm recovers this tree in the sense of \cite{Hartigan81a},
and 
%
\cite{KpLu11a,ChDaKpLu14a} obtain similar results for an underlying $k$-NN density estimator. In addition, \cite{KpLu11a,ChDaKpLu14a} propose
a simple pruning strategy,
that removes connected components that  artificially occur because of finite sample variability. 
However, the notion of recovery taken from \cite{Hartigan81a}
only focuses on the correct estimation of the cluster tree structure and not on the 
estimation of the clusters itself, cf.~the discussion in \cite{Steinwart11a}.
Finally, the most recent paper \cite{WaLuRi19a} establishes guarantees including rates of convergence
for each fixed level set, provided that a kernel-density estimator is used to produce level set estimates
and the density has a certain behavior such as $\a$-H\"older continuity.

A third approach taken in \cite{Steinwart11a, SrSt12a, Steinwart15a} tries to estimate both the \emph{first 
split $\rho^*$ in the cluster tree}, and the corresponding
clusters. As in the previously discussed papers, 
finite sample bounds are derived, which in \cite{Steinwart15a} are extended  to learning rates.
Moreover,  \cite{Steinwart15a} shows that these learning rates   can also 
be obtained by an adaptive, fully data-driven hyper-parameter selection strategy. Unfortunately, however, 
\cite{Steinwart11a, Steinwart15a} only consider the simplest possible density estimator,
namely a histogram approach, and \cite{SrSt12a}
restricts its considerations to compactly supported moving window density estimates for $\a$-H\"older-continuous densities.
In addition, the method in \cite{SrSt12a} requires to know $\a$, and in particular, it is not data-driven.
Finally, all three papers completely ignore the behavior of the considered algorithm 
for \emph{single cluster} distributions, i.e.~for
 distributions that do not have a split in the cluster tree. As a consequence, it is unclear whether and how 
a suitably modified version of this algorithm can be used to estimate the
\emph{split-tree}, i.e.~all levels at which a split in the cluster tree occurs, as well as
  the resulting clusters at these splits

The goal of this paper is to address the discussed issues of 
\cite{Steinwart11a, SrSt12a, Steinwart15a}. To be more precise, compared to these articles, 
we establish the following new results:
\begin{enumerate}
	\item For single cluster distributions,  we propose a  new set of 
	regularity
 assumptions
	for levels $\rho$ at which the level set $\{h\geq \r\}$ is small. For example, 
for bounded densities,   these 
   assumptions roughly speaking guarantee, that the level sets do not frazzle for levels $\r$ close to the maximum $\inorm h$ of the 
   density $h$. Such assumptions were missing in \cite{Steinwart11a, SrSt12a, Steinwart15a}.

\item 
We  present a simple modification of the output behavior of the
  generic cluster algorithm of \cite{Steinwart15a} to deal with distributions that do not have a split
in the cluster tree. Based on our new regularity 
  assumptions in \emph{i)} and the ones from  
\cite{Steinwart15a}, we then 
	show that this new cluster  algorithm
  is able to:
  \emph{a)}   provide an estimate $\rout$ of  the first split $\r^*$ in the cluster tree whenever there is one;
    \emph{b)} correctly detect distributions for which there is no such split $\r^*$,
    and   \emph{c)} construct estimates $B_i$ of the clusters $A_i^*$ occurring at the first split level. 
  Note that from a technical side both \emph{a)} with finite sample bounds on $|\rout - \rs|$  and \emph{c)} with finite sample bounds on $\lb^d(B_i \symdif A_i^*)$ directly follows 
  from \cite{Steinwart15a}, since our modification
  of the generic cluster algorithm scans through candidate levels $\r$ in exactly the same way 
  as the original algorithm of \cite{Steinwart15a} does. 
	Therefore, the  surprising, and compared to \cite{Steinwart15a} new   part of our finite sample guarantees  is the fact
 that this scanning procedure 
  does not need to be changed for correctly detecting  single cluster distributions in \emph{b)}.
Note that a highly beneficial side-effect of this fact is
that our analysis in \emph{b)} as well as in \emph{iii)} and \emph{iv)} below, can rely on the extensive set of tools 
developed in \cite{Steinwart15a}.

\item We then show  how the results of \emph{ii)} can be used to estimate  the entire 
		\emph{split-tree} by recursively applying the new generic cluster algorithm.
	While from a higher perspective this result does not seem to be too surprising, it turns
out that there are still a couple of serious technical difficulties involved. In a nutshell, these difficulties relate 
to the fact, that the generic algorithm may return an estimate $\rout$ for $\rho^*$ 
for which the connected components of $\{h\geq \rout\}$ are not yet sufficiently apart from each other. 
While such an estimate $\rout$ for $\r^*$ is desirable, it also prohibits a direct recursive application of the results 
of \emph{ii)}. To address this issue, we analyze the behavior of the generic cluster algorithm 
above the returned level $\rout$. In this analysis, which also goes beyond \cite{Steinwart11a, SrSt12a, Steinwart15a}, it turns out, that the algorithm 
 behaves correctly until it reaches 
a level $\r$ for which the connected components of $\{h \geq \r\}$ are sufficiently apart from each other. 
Above this level $\r$, the results of \emph{ii)} can then be recursively applied, leading to guarantees for 
the entire split-tree. We refer to Figure \ref{figure:split-tree} for a detailed description on how the different guarantees can be combined.

\item We show that the new generic cluster algorithm does not only work with 
  an underlying histogram density estimator (HDEs) as in \cite{Steinwart11a, Steinwart15a},
  but also for a variety of kernel density estimators (KDEs). Here it turns out that the results 
of \cite{Steinwart15a}, including those for the  
adaptive, fully data-driven hyper-parameter selection strategy,
remain valid for the resulting new clustering algorithm, provided that the kernel has a bounded 
support. Moreover, if the kernel has an exponential tail behavior, then the results remain true 
modulo an extra logarithmic term, while in the case of even heavier tails, we show that the 
rates become worse by a polynomial factor. Note that compared to \cite{Steinwart15a}, all the results
for   KDEs are new. Moreover, the results for KDEs substantially extend the results of 
\cite{SrSt12a}, since there \emph{a)} only moving window kernels were treated,  and \emph{b)}
only  $\a$-H\"older continuous densities with known $\a$ were considered. In contrast, our new results
do not even require continuous densities, and for this reason, we also obtain significantly more
general results than the currently best results for KDE-based clustering achieved in \cite{WaLuRi19a}.
The latter improvement 
is partially made possible,  because we can rely on the tools of \cite{Steinwart15a}. However, 
compared to the HDEs in \cite{Steinwart15a}
considering KDEs still requires significant technical efforts such as finite sample 
bound for the $\inorm\cdot$-distance between a KDE and its population version.

\end{enumerate}


The rest of this paper is organized as follows:
In Section \ref{sec-prelim} we briefly recall the key concepts of \cite{Steinwart15a}. In Section 
\ref{sec:smg} we first introduce the new regularity assumptions mentioned in \emph{i)}. We then 
introduce and analyze the new generic cluster algorithm as described in \emph{ii)}. Moreover, the 
recursive approach described in \emph{iii)} is analyzed in detail.
 Section \ref{sec:kde}
then presents key uncertain guarantees for level sets generated by KDEs,
and Section \ref{sec:algo} contains 
the material mentioned in \emph{iv)}, namely 
finite sample bounds as well as consistency results, rates of convergence,
and an adaptive data-driven 
strategy for choosing the kernel bandwidth.
In Section \ref{sec:comparison} we present a detailed comparison to the most closely related articles 
\cite{ChDaKpLu14a} and \cite{WaLuRi19a} on cluster tree estimation, and in Section \ref{sec:experiments} a couple of experiments on artificial data sets are 
reported.
All proofs can be found in Section \ref{sec:proofs}.

\section{Preliminaries}\label{sec-prelim}

%
%
%

In this section we recall  the setup for defining density-based clusters
in  a general context from \cite{Steinwart15a}.
To this end, let 
  $\snorm\cdot$ be  a norm on $\Rd$. Then we denote the closed unit ball of this norm by $B_{\snorm\cdot}$ and 
 write $B_{\snorm\cdot}(x,\d) := x + \d B_{\snorm\cdot}$ for the closed ball with center $x\in \Rd$ and 
radius $\d>0$. If the   norm is known from the context, we usually write
$B(x,\d)$ instead. Moreover, the Euclidean norm on $\Rd$ is denoted by $\tnorm\cdot$ and for the Lebesgue
volume of its unit ball we write $\vold$. Finally, $\inorm\cdot$ denotes the supremum norm for functions.

Let us now assume that we have some $A\subset X\subset \Rd$ as well as some norm $\snorm\cdot$ on $\Rd$.
Then, for $\d\geq0$ we define the $\d$-tube and $\d$-trim of $A$ in $X$ by
%
%
\begin{align*}
   A\pde := A_X\pde :=\{x\in X: d(x,A) \leq \d\}\, ,\qquad  \mbox{ and } \qquad  
   A\mde := A_X\mde :=X\setminus (X\setminus A)\pde\, ,
\end{align*}
 where $d(x,A) := \inf_{x'\in A}\snorm{x -x'}$. For later use we note that these definitions can actually be made
 in arbitrary metric spaces $(X,d)$, see also  \cite{Steinwart15a}.
We further write $\mathring A$ for the interior of $A$ and $\overline A$ for the closure of $A$.
Moreover, $\partial A := \overline A\setminus \mathring A$ denotes the boundary of $A$.
Obviously, we have $A^{+0} = \overline A$, and hence also $A^{-0} = \mathring A$.
Furthermore, since $x\mapsto d(x,A)$ is continuous, $A\pde$ is always closed in $X$ and $A\mde$ is always open in $X$.
Note that if $A$ is bounded, we always find compact and convex $X\subset \Rd$ with 
$A_X\pde = A_\Rd\pde$ and $A_X\mde = A_\Rd\mde$. In fact, for the first identity we may choose 
$X:= (2\d + \snorm A)B_{\snorm\cdot}$, where $\snorm A := \sup_{x\in A}\snorm x$, while
for the second identity $X:= (1 + \snorm A)B_{\snorm\cdot}$ is a suitable choice.
Based on this observation and the fact that we usually consider $\d\in (0,1]$ in combination 
with some suitably chosen $X$, see \assx P below for details, 
we often ignore the surrounding set $X$.
In addition to this notations, we denote the inradius and diameter of a bounded $A\subset \Rd$ by $\inrad A$ and $\diam A$, respectively, that is 
\begin{align*}
 \inrad A &:= \sup\bigl\{ r>0: \exists x\in A\mbox{ with } B(x,r)\subset A  \bigr\} \\
 \diam A &:= \sup\bigl\{ \snorm{x-x'}: x,x'\in A \bigr\} \, .
\end{align*}
For later use we note that for all $\d>0$ we have $A_\Rd\mde\neq \emptyset$ if and only if $\d<\inrad A$, and 
by the observation above, the same is true for all sufficiently large compact and convex sorrounding $X\subset \Rd$. 
In addition, it holds $\diam A_\Rd\pde = 2\d + \diam A$ for all $\d>0$. For details, we refer to Lemma \ref{lemma:inrad}
and Lemma \ref{lemma:diam}, respectively.

Throughout this work, $\eins_A$ denotes the indicator function of a set $A$,
and $A \symdif B$ 
the symmetric difference of two sets $A$ and $B$.
Let us now assume that $P$ is a non-zero finite  measure\footnote{In \cite{Steinwart15a} only probability measures were considered, but for later use it is more convenient to consider finite measures instead. It it easy to check that all results mentioned in the following remain true for finite measures by a simple reparametrization of the levels $\r$.} on a closed $X\subset \Rd$ that is absolutely continuous 
with respect to the Lebesgue measure $\lbd$. Then $P$ has a $\lbd$-density $h$ and 
as explained in the introduction, one could define the clusters of $P$ to be
the connected components of the level set $\{h\geq \r\}$, where $\r\geq 0$ is some user-defined 
threshold. Unfortunately, however, this notion leads to serious issues if there is no canonical choice 
of $h$ such as a continuous version, see the illustrations in \cite[Section 2.1]{Steinwart15a}. 
To address this issue, \cite{Steinwart15a} considered, for $\r\geq 0$, the measures
\begin{displaymath}
\mu_\r(A):= \lbd(A\cap \{h\geq \r\}) \qspace A\in \ca B(\Rd)\, .
\end{displaymath}
Since $\mu_\r$ is independent of the choice of $h := \dP/\dld$, the set
\begin{align}\label{Def-Mr}
  M_\r & := \supp \mu_\r\, ,
\end{align}
where $\supp \mu_\r$ denotes the support of the measure $\mu_\r$, is independent of this choice, too.
For any  $\lbd$-density $h$ of $P$, the definition immediately gives
\begin{equation}\label{reg-half}
\lbd\bigl( \{h\geq \r\} \setminus M_\r\bigr) = \lbd\bigl( \{h\geq \r\} \cap (\Rd\setminus M_\r)\bigr) = \mu_\r(\Rd\setminus M_\r) = 0\, ,
\end{equation}
i.e.~modulo $\lbd$-zero sets, the level sets $\{h\geq \r\}$ are not larger than $M_\r$.
In fact, $M_\r$ turns out to be the smallest closed  set satisfying \eqref{reg-half} 
and 
it is shown in 
\cite[Lemma A.1.2]{SteinwartXXb1}, that  
 \begin{equation}\label{Mr-diff-closure}
  {\mathring{\{h\geq \r\}}} \subset M_\r  \subset \overline{\{h\geq \r\}}
  \qquad 
  \mbox{ and }
  \qquad 
   M_\r \symdif \{h\geq \r\} \subset \partial \{h\geq \r\}
  \, .
\end{equation}
In order to ensure  inclusions that are ``inverse'' to \eqref{reg-half}, \cite{Steinwart15a} said that $P$ is normal at 
level $\r$ if   
 there exist two 
 $\lbd$-densities $h_1$ and $h_2$  of $P$ such that 
 \begin{displaymath}
    \lbd (M_\r\setminus \{h_1\geq \r\}  ) = \lbd(\{h_2>\r\}  \setminus \mathring M_\r) = 0\, .
 \end{displaymath}
It is shown in 
\cite[Lemma A.1.3]{SteinwartXXb1}\footnote{In this lemma, the term ``upper normal at level $\r$'' means that 
$\lbd (M_\r\setminus \{h_1\geq \r\}  ) = 0$ for some density $h_1:= \dP/\dld$ while ``lower normal at level $\r$'' 
means $ \lbd(\{h_2>\r\}  \setminus \mathring M_\r) = 0$ for some density $h_2:= \dP/\dld$.}
that $P$ is  normal at every level, if it has both an upper semi-continuous $\lbd$-density $h_1$ and a
lower semi-continuous $\lbd$-density $h_2$.
Moreover, if 
$P$ has a $\lbd$-density $h$ such that $\lbd(\partial\{h\geq \r\})=0$, then 
the same lemma shows that 
$P$ is 
  normal at level $\r$.
Finally, note that if the conditions of normality at level $\r$ are satisfied for some $\lbd$-densities $h_1$ and $h_2$ of $P$, then 
they are actually satisfied for all  $\lbd$-densities $h$ of $P$
and we have $\lbd(M_\r \symdif \{h\geq \r\})=0$. The next assumption collects
the concepts introduced so far.

\begin{assumption}{P}
We have a $\lbd$-absolutely continuous, finite, non-zero measure $P$ 
that is normal at every level. In addition,  $\supp P$ is compact, and $X\subset \Rd$ is 
compact and connected set with $(\supp P)_\Rd^{+2} \subset X$. Finally, 
$\mu$ denotes the Lebesgue measure on $X$.
%
\end{assumption}

Note that the assumption $(\supp P)_\Rd^{+2} \subset X$ ensures both 
 $A_X\pde = A_\Rd\pde$ and $A_X\mde = A_\Rd\mde$
for all $\d\in (0,\dthick]$ and $A\subset \supp P$. In the following, we therefore usually ignore the sorrounding $X$
when dealing with $\d$-tubes and $\d$-trims of such $A$.

Let us now recall the definition of clusters from \cite{Steinwart15a}. We begin with the following
definition that compares different partitions.

\begin{definition}\label{comp-partitions}
   Let $A\subset B$ be   non-empty sets, and $\ca P(A)$ and $\ca P(B)$ be   partitions of $A$ and $B$, respectively.
   Then $\ca P(A)$ is comparable to $\ca P(B)$, write  $\ca P(A) \comparable \ca P(B)$,
   if, for all $A'\in \ca P(A)$, there is a $B'\in \ca P(B)$
   with $A'\subset B'$.
\end{definition}

Informally speaking, $\ca P(A)$ is comparable to $\ca P(B)$, if no cell $A'\in \ca P(A)$ is broken into pieces in 
$\ca P(B)$. In particular, if $\ca P_1$ and $\ca P_2$ are two partitions of $A$, then 
$\ca P_1 \comparable \ca P_2$ if and only if $\ca P_1$ is finer than $\ca P_2$.
Let us now assume that we have have two partitions $\ca P(A)$ and $\ca P(B)$ 
with $\ca P(A) \comparable \ca P(B)$. Then \cite[Lemma A.2.1]{SteinwartXXb1} shows that 
there exists a unique map $\z: \ca P(A) \to \ca P(B)$ such that, for all $ A'\in \ca P(A)$, we have 
\begin{equation*}
 A' \subset \z(A')\, .
\end{equation*}
Following \cite{Steinwart11a,Steinwart15a}, we call 
 $\z$ the cell relating map (CRM) between $A$ and $B$. Moreover, if $\z$ is bijective, we say that $\ca P(A)$ is \emph{persistent} in $\ca P(B)$ and write 
 $\ca P(A) \persist \ca P(B)$.

The first example of comparable partitions come from connected components.
To be more precise, let $A\subset \Rd$ be a closed subset and $\ca C(A)$
be the collection of its  connected components. By definition, $\ca C(A)$ forms a partition of $A$, and 
if $B\subset \Rd$ is another closed subset with $A\subset B$ and
$|\ca C(B)|<\infty$ then we have $\ca C(A) \comparable\ca C(B)$, see 
\cite[Lemma A.2.3]{SteinwartXXb1}.

Following \cite{Steinwart15a}, another class of partitions arise from a discrete notion of 
path-connectivity.
To recall the latter, we fix a $\t>0$, an $A\subset \Rd$, and a norm $\snorm\cdot$ on $\Rd$.
Then $x,x'\in A$ are $\t$-connected in $A$, 
if 
  there exist
$x_1,\dots,x_n\in A$ such that  $x_1=x$, $x_n=x'$ and $\snorm{x_i-x_{i+1}} < \t$ for all $i=1,\dots,n-1$.
Clearly, 
being $\t$-connected  gives an equivalence relation on $A$.
We write $\ca C_\t(A)$ for the resulting partition and call its cells 
 the {\em $\t$-connected components of $A$}.
It has been shown in   \cite[Lemma A.2.7]{SteinwartXXb1},
that $\ca C_\t(A) \comparable\ca C_\t(B)$ for all $A\subset B$ and $\t>0$.
Moreover, if $|\ca C(A)|<\infty$ then $\ca C(A)=\ca C_\t(A)$
for all sufficiently small $\t>0$, see \cite[Section 2.2]{Steinwart15a} for details.

Following \cite{Steinwart15a},
we can now describe finite measures that can be clustered.

\begin{definition}\label{top-clust-def1-neu}
Let \assx P be satisfied. Then $P$
 can be   clustered between 
$\r^*\geq 0$ and $\r^{**}>\r^*$, if 
for all $\r\in [0,\r^{**}]$,   
the following three
conditions are satisfied:
\begin{enumerate}
  \item We  either have $|\ca C(M_\r)|= 1$ or $|\ca C(M_\r)|= 2$.
  \item If we have $|\ca C(M_\r)|=1$, then $\r\leq \r^*$.
  \item If we have $|\ca C(M_\r)|=2$, then $\r\geq \r^*$  and $\ca C (M_{\r^{**}})\persist \ca C(M_\r)$.
\end{enumerate}
Using the CRMs $\z_\r:\ca C (M_{\r^{**}})\to \ca C(M_\r)$, we then define 
the clusters of $P$ by
\begin{displaymath}
 A^{*}_i := \bigcup_{\r\in (\r^*,\r^{**}]} \z_\r(A_i) \qspace i\in \{1,2\}\, ,
\end{displaymath}
where 
$A_1$ and $A_2$ are the topologically connected components of $M_{\r^{**}}$. Finally, we define
\begin{equation}\label{reg-cluster-lem-def-tds-new}
   \ts(\e) := \frac 1 3 \cdot d\bigl(\z_{\rs+\e}(A_1) , \z_{\rs+\e}(A_2) \bigr)\, , \qquad \qquad \e\in (0,\rss-\rs].
\end{equation}
\end{definition}

Definition \ref{top-clust-def1-neu} ensures that the level sets below $\rs$ are connected, while 
for a certain range above $\rs$
the level sets  have exactly two components, which, in addition, are assumed to be persistent.
Consequently, the topological structure between $\rs$ and $\rss$ is already determined by that of 
$M_\rss$, and we can use the connected components of $M_\rss$ to number the connected components  of $M_\r$ for $\r\in (\rs,\rss)$.
This is done in the definition of the clusters $A_i^*$ as well as in the definition of the function
$\ts$, which essentially measures the distance between the two connected components at level $\rs+\e$.

The major goal of \cite{Steinwart11a, Steinwart15a} was to design an algorithm that is able to 
asymptotically estimate both the correct value of $\rs$ and the clusters $A_1^*$ and $A_2^*$.
Moreover, \cite{Steinwart15a} established rates of convergence for both 
estimation problems, and these rates did depend on the behavior of the function $\ts$.
However, this algorithm required that the level sets do not have bridges or cusps that are too thin.
To make this precise, let us recall that for a closed $A\subset \Rd$, 
\cite{Steinwart11a, Steinwart15a}  considered the 
function $\psis_A:(0,\infty)\to [0,\infty]$ defined by 
\begin{displaymath}
\psis_A(\d) :=  \sup_{x\in A} d(x, A\mde) \, , \qquad \qquad \d>0.
\end{displaymath}
Roughly speaking, $\psis_A(\d)$ describes the smallest radius $\e$ needed to ``recover'' $A$ 
from $A\mde$ in the sense of $A\subset (A\mde)^{+\e}$, see \cite[Section A.5]{SteinwartXXb1} for this and 
various other results on $\psis_A$. In particular, we have $\psis_A(\d) \geq \d$ for all $\d>0$ and 
 $\psis_A(\d) = \infty$ if $A\mde=\emptyset$. Moreover, $\psis_A$ behaves linearly, if bridges and cusps 
of $A$ are not too thin, and even thinner cusps and bridges can be included by considering sets with 
$\psis_A(\d) \leq c \d^\g$ for some constants $c\geq 1$, $\g\in (0,1]$ and all sufficiently small $\d>0$.
Finally,  
for our later results we need to recall from  \cite[Lemma A.4.3]{SteinwartXXb1} that 
for all $\d>0$ with $A\mde\neq \emptyset$ and all $\t>2\psis_A(\d)$ we have 
\begin{equation}\label{tmde}
   |\cct{A\mde}| \leq |\cc A|
\end{equation}
whenever $A$ is contained in some compact $X\subset \Rd$  and $|\cc A|<\infty$.

With the help of these preparations we can now recall the 
following definition taken from 
\cite{Steinwart15a}, which categorizes the behavior of $\psis_\mr$.

\begin{definition}\label{def-reg}
Let \assx P be satisfied.  Then we
 say that  $P$ 
has thick level sets of order  $\g\in (0,1]$ up to the level $\rss>0$,
if there 
exist constants $\cthick\geq 1$ and  $\dthick\in (0,1]$ such that, for all $\d\in(0, \dthick]$ and  $\r\in [0,\r^{**}]$, we have 
\begin{equation}\label{def-thick-h0}
 \psis_\mr(\d)
 \leq \cthick \,\d^\g \, .
\end{equation}
In this case, we call  
$\psi(\d) := 3\cthick\d^\g$ 
the thickness function of $P$.  
\end{definition}

The following assumption, which collects all concepts introduced so far, describes the 
finite measures we wish to cluster.

\begin{assumption}{M}
The finite measure 
$P$  
 can be clustered between   $\r^*$ and $\r^{**}$.
In addition, $P$ has 
thick level sets of order  $\g\in (0,1]$
up to the level $\rss$. We denote
the corresponding thickness function by $\psi$ and write $\ts$
for the function defined in \eqref{reg-cluster-lem-def-tds-new}.
\end{assumption}

Recall that 
the theory developed in \cite{Steinwart11a, SrSt12a, Steinwart15a}
focused on the question, whether it is possible to estimate $\rs$ and the 
resulting clusters for distributions that can be clustered.
To this end, a generic cluster algorithm 
 was developed, which receives   some level set estimates $L_{D,\r}$
of $\mr$ satisfying 
\begin{equation}\label{uq}
   M_{\r+\e}\mde \subset L_{D,\r} \subset M_{\r-\e}\pde
\end{equation}
for all $\r\in [0,\rss]$ and some $\e,\d>0$. The key result
\cite[Theorem 2.9]{Steinwart15a} then specified in terms of $\e$ and $\d$
how well this algorithm estimates both $\rs$ and the clusters $A_1^*$ and $A_2^*$.
What is missing in this analysis, however, is an investigation of the behavior 
of the generic cluster algorithm in situations in which $P$ cannot be clustered because
 all level sets are connected.

Now observe that 
the reason for this gap was the notion of thickness: 
Indeed, if $P$ is a \emph{single-cluster finite measure}, i.e.~$|\ca C(\mr)|\leq 1$ for 
all $\r\geq 0$, and $P$ has thick 
level sets of the order $\g$ up to the level $\rss := \sup\{\r: \r\geq 0 \mbox{ and } |\ca C(\mr)| = 1 \}$,
then the proof of \cite[Theorem 2.9]{Steinwart15a} can be easily extended to show that 
at each visited level $\r$
the algorithm 
correctly detects  exactly one connected component.
Unfortunately, however, the assumption of having thick levels up to the height $\rss$ of the peak of $h$
is too unrealistic, as it requires $\mrmde\neq \emptyset$ for all $\r\in [0,\rss]$ and $\d\in (0,\dthick]$,
that is, \emph{the 
 peak needs to be a plateau that contains a ball of radius $\dthick$.}

\section{A Generic Algorithm for Estimating the Split-Tree}\label{sec:smg}

The overall goal of this section is to present a generic algorithm for estimating the 
entire split-tree. To this end, we first introduce a new set of assumptions for single-cluster distributions 
that rule out irregular behavior of the level sets  in the vicinity of the peak of the density.
Unlike the na\"ive approach for extending the results of \cite{Steinwart15a} to single-cluster distributions,
which we have discussed at the end of Section \ref{sec-prelim}, this new set of assumptions 
includes a variety of realistic behaviors.
In the second step we then present a generic cluster algorithm, whose only difference to the one 
in \cite{Steinwart15a} is its output behavior in situations in which no split has been detected. 
We then show that 
this new cluster algorithm, like its 
predecessor
in \cite{Steinwart15a},
  correctly identifies a  split in the cluster tree. Moreover, 
we demonstrate that, unlike the one in \cite{Steinwart15a}, 
the new cluster algorithm also correctly identifies single-cluster distributions.
Finally, we combine these insights to develop a new generic algorithm for estimating 
the entire cluster tree.

Let us begin by   introducing  the already mentioned  new   assumption 
for dealing with single-cluster distributions.
%

\begin{assumption}{S}
Assumption P is satisfied and there are $\rls\geq 0$,
$\g\in (0,1]$, $\cthick\geq 1$ and $\dthick\in (0,1]$ such that 
 for all $\r\geq \rls$ and  $\d\in(0,\dthick]$,  we have:
\begin{enumerate}
   \item $|\cc \mr|\leq 1$.
	\item If $\mrmde\neq\emptyset$ then 
		   $\psis_\mr(\d)  \leq \cthick \d^\g$.
	\item If $\mrmde=\emptyset$, then, for all  $\emptyset \neq A\subset \mrpde$ and 
$\t>2\cthick \d^\g$, 
we have $|\cct A| = 1$.
\end{enumerate}
\end{assumption}

Note that condition \emph{i)} simply means that the level sets of $P$ above $\rls$ 
are either empty or connected. If they are empty, there is nothing more to assume 
and in the other case, we can either have $\mrmde\neq \emptyset$ or $\mrmde=\emptyset$.
If  $\mrmde\neq \emptyset$ , then condition \emph{ii)} ensures that the level set $\mr$ is still thick
in the sense of Definition \ref{def-reg}, while in the other case 
$\mrmde= \emptyset$, condition \emph{iii)}  guarantees that the larger sets $\mrpde$
cannot have multiple $\t$-connected components as long as we choose $\t$ in a way that is required 
in the case of multiple clusters, too. 
In this respect note that Lemma \ref{lemma:inradius-decreases} shows that 
for all $\d\in(0,\dthick]$, there exists a $\r\geq \rls$ with $\mrmde=\emptyset$.
In other words, the situation $\mrmde=\emptyset$ does occur for all sufficiently high
level sets.

\begin{figure}
\includegraphics[width=0.45\textwidth]{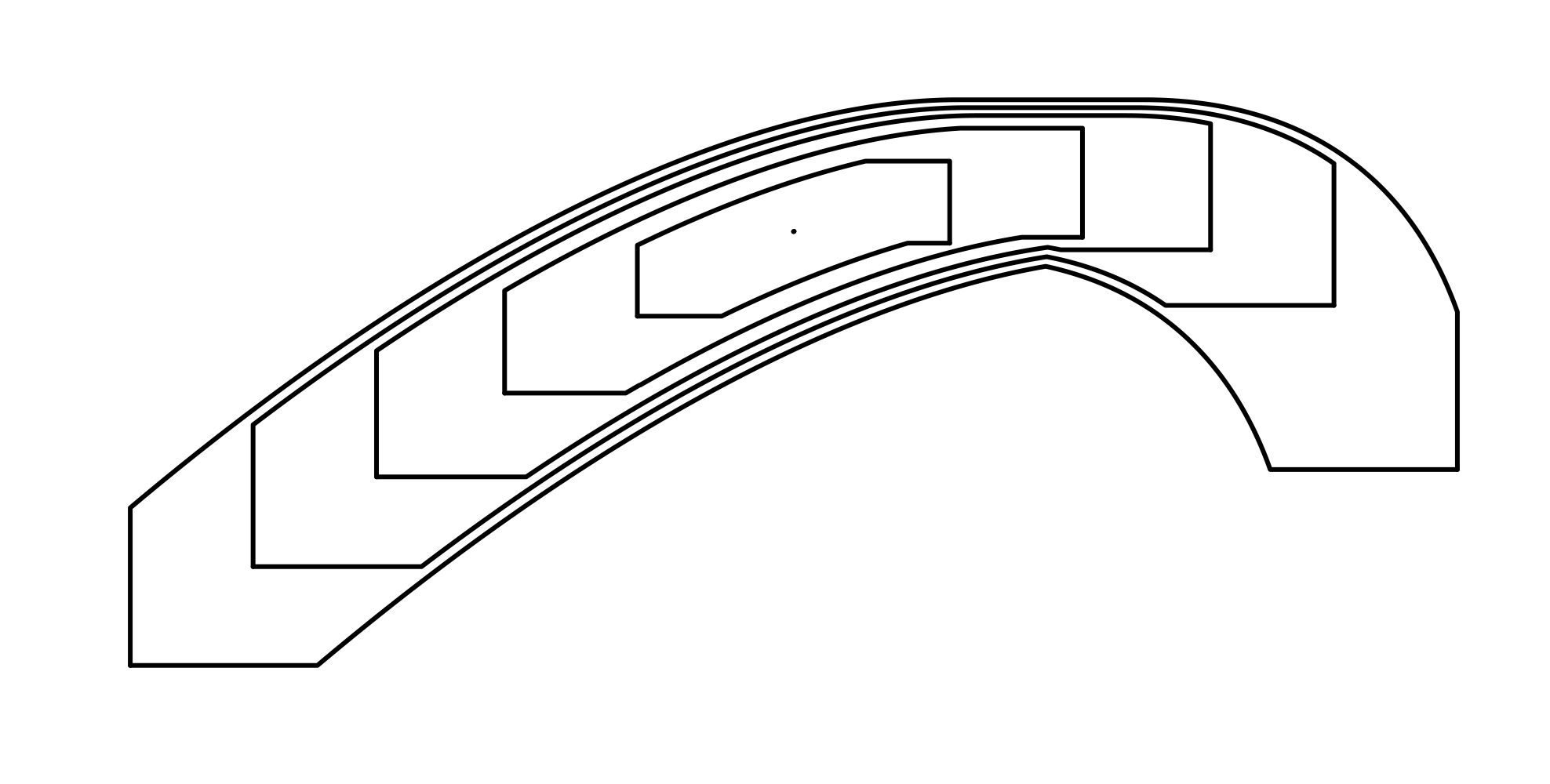}
\hspace*{0.03\textwidth}
\includegraphics[width=0.45\textwidth]{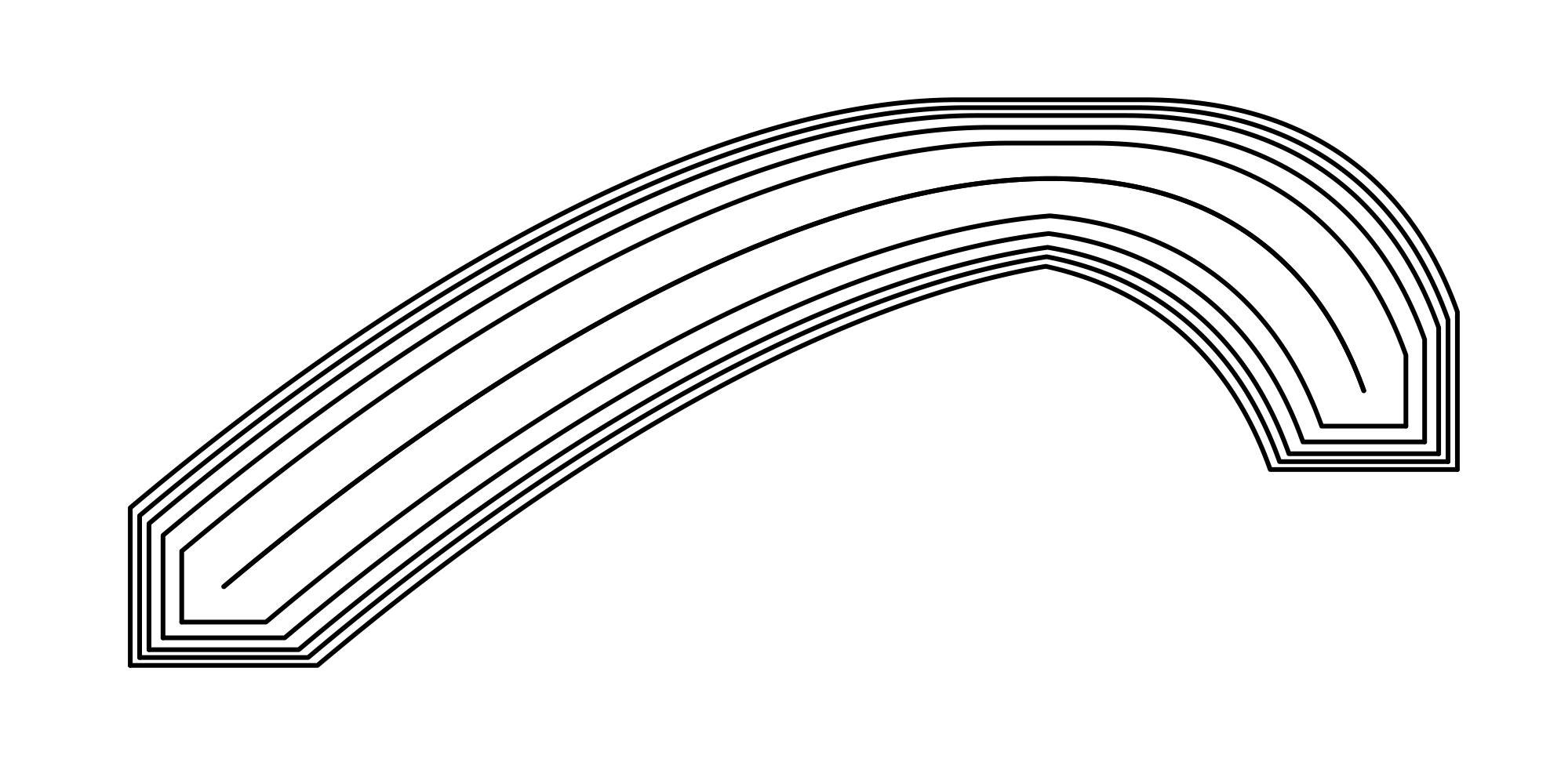}
\caption{Contour lines of two continuous densities. The levels $\rho$ were chosen such that 
in both cases we have $\inrad \mr \in \{0, 1, 2, 3, 4, 5\}$ with respect to the $\inorm \cdot$.
\textbf{Left.}  \assx S is satisfied for $\g=1$, since, for 
increasing $\r$, $\diam \mr$ decays faster than $\inrad \mr$ does, and hence \eqref{assS:reform} holds. 
\textbf{Right.} \assx S is not satisfied for $\g=1$, since the highest level set 
satisfies $\inrad \mr =0$ and $\diam \mr >0$, which violates \eqref{assS:reform}.
%
}
\label{figure:ass-s}
\end{figure}

Condition \emph{iii)} can also be interpreted in terms on inradius and diameter.
Indeed, by choosing the surrounding $X\subset \Rd$ sufficiently large, Lemma \ref{lemma:inrad} shows that $\mrmde=\emptyset$
if and only if $\d\geq \inrad \mr$. Moreover, Lemma \ref{lemma:tau-connect-subsets} shows that 
$|\cct A| = 1$ for all $\emptyset \neq A\subset \mrpde$ if and only if $\t> \diam \mrpde = 2\d + \diam \mr$, where for the latter
identity we  used Lemma \ref{lemma:diam}. Combining these, we see that in the most interesting case $\g=1$, condition \emph{iii)} is satisfied if and only if
\begin{align}\label{assS:reform}
 2 (\cthick -1)\inrad \mr \geq \diam \mr
\end{align}
for all $\r\geq \rs$ with $\inrad \mr \leq \dthick$. Inequality 
\eqref{assS:reform} essentially  states that the diameter-inradius ratio must be bounded for increasing $\r$. 
Consequently,  \eqref{assS:reform} is satisfied for $\cthick := 2$ if all $\mr$ 
above $\rls$ are balls with respect to the considered norm
since in this case we have $2\inrad \mr = \diam \mr$.
Moreover, by increasing $\cthick$ we see that \eqref{assS:reform} remains true 
if we distort these balls by
bi-Lipschitz continuous maps with constants that can be bounded independently of $\r$.
Analogously, \eqref{assS:reform} is satisfied, if $\mr$ are balls with respect to a \emph{fixed} norm
that is different to the one used for the $\d$-tubes and $\d$-trims in condition \emph{iii)}.
In addition, \eqref{assS:reform} is violated, if, for example, 
$\inrad \mx {\rmax} =0$ and $\diam \mx {\rmax} >0$ for $\rmax := \sup\{\r>0: \mr \neq \emptyset\}<\infty$.
In contrast, if we have a flat plateau at the highest level $\rmax$, that is 
$\inrad \mx {\rmax} >0$, then \eqref{assS:reform}  is always satisfied for some $\cthick>1$
because of $\diam \mr \leq \diam X<\infty$.
Finally, for $d=1$, we always have $2 \inrad \mr = \diam \mr$ for all non-empty level sets $\mr$, since for these
$|\cc\mr| = 1$ ensures that $\mr$ is an interval. Consequently, \assx S is satisfied for $\g=1$, $\cthick = 2$, and $\dthick = 1$ whenever \assx P is satisfied.
For two-dimensional examples we refer to Figure \ref{figure:ass-s} for an illustration. 

%

\begin{algorithm}[t]
\caption{Clustering with the help of a generic level set estimator}
\label{cluster-algo-generic}
\begin{algorithmic}[1]
\REQUIRE{ Some $\t>0$ and  $\e>0$ and a start level $\r_0\geq 0$.\\
  \hspace*{2.7em} 
  A decreasing family 
   $(L_{\r})_{\r\geq 0}$ of 
    subsets of $X$.}
\ENSURE An estimate of $\rls$ or $\rs$ and the corresponding   clusters.
\STATE $\r\gets \r_0$
\REPEAT 
	\STATE {Identify the $\t$-connected components $B_1,\dots,B_M$ of $L_{\r}$  
	       satisfying \begin{displaymath}
	                   B_i\cap L_{\r+2\e}\neq \emptyset.
	                  \end{displaymath}}
	\STATE $\r \gets \r+\e$
\UNTIL{$M \neq 1$}
	\STATE $\r \gets \r+2\e$
	\STATE {Identify the $\t$-connected components $B_1,\dots,B_M$ of $L_{\r}$  
	       satisfying \begin{displaymath}
	                   B_i\cap  L_{\r+2\e} \neq \emptyset.
	                  \end{displaymath}}
 \IF{$M>1$}
	\STATE	{\textbf{return} $\rout=\r$ and the sets $B_i$ for $i=1,\dots,M$.} 
\ELSE
	\STATE {\textbf{return} $\rout=\r_0$ and the set $L_{\r_0}$. }
\ENDIF
\end{algorithmic}
\end{algorithm}

The next task is to formulate a generic algorithm that is able to estimate $\rs$ and the resulting 
clusters if $P$ can be clustered in the sense of Assumption M
and that is able to detect distributions that cannot be clustered in the sense of Assumption S.
We will see in the following that Algorithm \ref{cluster-algo-generic} is such an algorithm.
Before we present the corresponding results we first note that the
only difference of  Algorithm \ref{cluster-algo-generic} to the algorithm considered
in \cite{Steinwart15a} is the more flexible start level $\r_0$, compared to $\r_0 = 0$ in \cite{Steinwart15a}, and the 
modified output
in Lines 8-12. Indeed, the algorithm in
\cite{Steinwart15a} always produces the return values of Line 9. In contrast, 
Algorithm \ref{cluster-algo-generic} distinguishes between the cases $M>1$ and $M=0$. While for 
$M>1$ the output of both algorithms exactly coincide, the new 
Algorithm \ref{cluster-algo-generic} now returns $\r_0$ and $L_{\rho_0}$ in the case of $M=0$. 
We will see in Theorem \ref{main-generic-single} that the latter case typically occurs for distributions 
satisfying Assumption S. In this respect recall that $L_{\rho_0}$ can be viewed as an estimate of 
$M_{\r_0}$ and therefore returning  $L_{\rho_0}$  makes sense for such distributions.


With these preparations we can now formulate the following 
adaptation of \cite[Theorem 2.9]{Steinwart15a} to the new Algorithm \ref{cluster-algo-generic}.
Since the proof of  \cite[Theorem 2.9]{Steinwart15a}  can be easily adapted to 
arbitrary   start levels $\r_0\geq 0$ and this 
proof also shows that 
the case $M\leq 1$ is not occurring under the assumptions of this theorem, we omit the proof of 
Theorem \ref{analysis-main-combined-new}.

\begin{theorem}\label{analysis-main-combined-new}
Let \assx  M  be satisfied. Furthermore, let 
  $\e^*\leq (\rss - \rs)/9$ , 
 $\d\in (0,  \dthick]$,  
$\t\in (\psi(\d),\ts(\e^*)]$, and  $\e\in (0 , \e^*]$, and $\r_0 \leq \rs$.
In addition, let   $(L_{\r})_{\r\geq 0}$ be a 
decreasing family satisfying \eqref{uq}
for all $\r\geq \r_0$. 
Then we have:
\begin{enumerate}
 \item The  level $\rout$ returned by  Algorithm \ref{cluster-algo-generic} satisfies  $\rout \in [\rs+2\e , \rs+\e^*+5\e]$
  and 
  \begin{equation}\label{bound-rds-second}
     \t-\psi(\d) < 3\t^*\bigl(\rout-\rs+\e\bigr)\, .
  \end{equation}
  \item Algorithm \ref{cluster-algo-generic} returns two sets $B_1$ and $B_2$ and these  
   can be ordered such that we have 
          \begin{equation} 
   \sum_{i=1}^2 \mu\bigl(B_i \symdif A_i^*\bigr)   \label{cluster-chunk-generic-bound}
   \leq 
 2\sum_{i=1}^2 \mu \bigl(A_i^* \setminus  (A^i_{\rout + \e})\mde\bigr)  
  +
  \mu\bigl ( M_{\rout-\e}\pde \setminus \{ h>\rs \}\bigr)\, .
\end{equation}
   Here, $A^i_{\rout+\e} \in \ca C(M_{\rout+\e})$ are ordered in the sense of  $A^i_{\rout+\e}\subset A_i^*$.
\end{enumerate}
\end{theorem}

Theorem \ref{analysis-main-combined-new} shows that the modified Algorithm \ref{cluster-algo-generic}
is still able to estimate $\rs$ and the corresponding clusters 
if the distribution can be clustered in the sense of Assumption M.
The main motivation for this section was, however, to have an algorithm that also behaves correctly 
for distributions that cannot be clustered in the sense of Assumption S.
The next theorem shows that Algorithm \ref{cluster-algo-generic} does indeed have such a behavior.

\begin{theorem}\label{main-generic-single}
      Let \assx S be satisfied and
 $(\lr)_{\r\geq 0}$ be a decreasing family of sets $\lr\subset X$ such that 
\eqref{uq} holds
for some fixed $\e,\d>0$ and all $\r\geq \r_0$. If $\r_0\geq \rls$,
$\d\in (0,\dthick]$, 
and  $\t>2\cthick\d^\g$, then Algorithm \ref{cluster-algo-generic} returns $\r_0$ and $L_0$.
\end{theorem}

Note that Theorem \ref{analysis-main-combined-new} requires $\t > \psi(d) = 3\cthick \d^\g$,
while Theorem \ref{main-generic-single} even holds under the milder assumption 
$\t>2\cthick \d^\g$. Consequently, if we choose a $\t$ with 
$\t > 3\cthick \d^\g$, then the corresponding assumptions of both theorems are satisfied.
Moreover, the additional assumption $\t< \ts(\e^*)$ in Theorem \ref{analysis-main-combined-new}
is actually more an assumption on $\e^*$ than on $\t$ as we will see later when we apply 
Theorems \ref{analysis-main-combined-new} and 
\ref{main-generic-single}. 

%
%
%
%

Roughly speaking, Theorem \ref{analysis-main-combined-new} shows that Algorithm \ref{cluster-algo-generic} correctly detects 
the next split $\r^*$ above the start level $\r_0$ whenever there is such a split, 
while Theorem \ref{main-generic-single} shows that  
Algorithm \ref{cluster-algo-generic} also correctly identifies situations, in which there is no split above 
the start level $\rho_0$.

\begin{algorithm}[t]
\caption{Estimating the split-tree with the help of a generic level set estimator}
\label{split-tree-algo-generic}
\begin{algorithmic}[1]
\REQUIRE{ Some $\t>0$,   $\e>0$ and a start level $\r_0\geq 0$.\\
   decreasing family 
   $(L_{\r})_{\r\geq 0}$ of 
    subsets of $X$.}
\ENSURE Estimates of all splits  of the cluster tree and the corresponding   clusters.
\STATE {Call Algorithm \ref{cluster-algo-generic} with $\r_0$ and $(L_{\r})_{\r\geq 0}$}
\IF {$\rout > \r_0$}
\STATE{Store the return values of Algorithm \ref{cluster-algo-generic} in the split-tree}
\STATE {Call Algorithm \ref{split-tree-algo-generic} with $\rout + \e$ and $(L_{1, \r})_{\r\geq 0}$} 
\STATE {Call Algorithm \ref{split-tree-algo-generic} with $\rout + \e$ and $(L_{2, \r})_{\r\geq 0}$} 
\ELSIF{$\rout = 0$}
\STATE{Store the return values of Algorithm \ref{cluster-algo-generic} in the split-tree}
\ENDIF

	\STATE {\textbf{return} split-tree}

\end{algorithmic}
\end{algorithm}

%
%

Now assume that the assumptions of Theorem \ref{analysis-main-combined-new} are satisfied 
and that Algorithm \ref{cluster-algo-generic} returned 
$\rout$  and the cluster estimates  $B_1, B_2$.
Our goal is to apply Algorithm \ref{cluster-algo-generic} on the two detected clusters 
$B_1$ and $B_2$ separately in a recursive fashion, see Algorithm \ref{split-tree-algo-generic}.
To this end, we define the new level set estimates 
\begin{align*}
 L_{i,\r} := L_\r \cap B_i\, , \qquad \qquad i=1,2, \, \r\geq \rout,
\end{align*}
and let the Algorithm \ref{cluster-algo-generic} run on both families of level set 
estimates separately. 
Of course, we want to  use our insights into  Algorithm \ref{cluster-algo-generic} as 
much as possible. For this reason, we need to replace \eqref{uq} by a suitable
new horizontal and vertical control.

To find such a new control, 
let us assume that \assx  M  is satisfied and that we have fixed a $\rda \in  (\rs, \rss]$.
Moreover, let 
$\ccr 1\rda$, and $\ccr 2\rda$ be the two connected components of $\mx \rda$, 
i.e.~$\cc{\mx\rda} = \{\ccr 1\rda,\ccr 2\rda\}$.
For $i=1,2$ we then define two new ``children'' probability measures $P_1$ and $P_2$ by 
\begin{align}\label{child-measures}
 P_i (B) := \frac {P(B \cap \ccr i\rda)} {P(\ccr i\rda)} 
\end{align}
for all measurable sets $B\subset X$. Moreover, for $\r\geq 0$  we denote the level sets of $P_1$ and $P_2$
by $\mri 1$ and 
$\mri 2$, respectively. With the help of these notations we can now introduce distributions having a finite 
split tree.

\begin{definition}
 Let $P$ be a  distribution satisfying \assx P and $|\cc \mr| < \infty$ for all $\r\geq 0$. Moreover, assume that 
 there is a $\rmax>0$ such that $\mr = \emptyset$ for all $\r\geq \rmax$.
 Then we say that $P$ has a finite split-tree with minimal step size $\eps>0$, if one of the following two conditions are satisfied:
 \begin{enumerate}
  \item $P$ satisfies \assx S.
  \item $P$ satisfies \assx M with $\rss-\rs \geq \eps$, and for $\rda := (\rss + \rs)/2$ the two measures $P_1$ and $P_2$ defined by 
  \eqref{child-measures} have a finite split-tree  with minimal step size $\eps>0$.
 \end{enumerate}
\end{definition}


\begin{figure}
\includegraphics[width=0.45\textwidth]{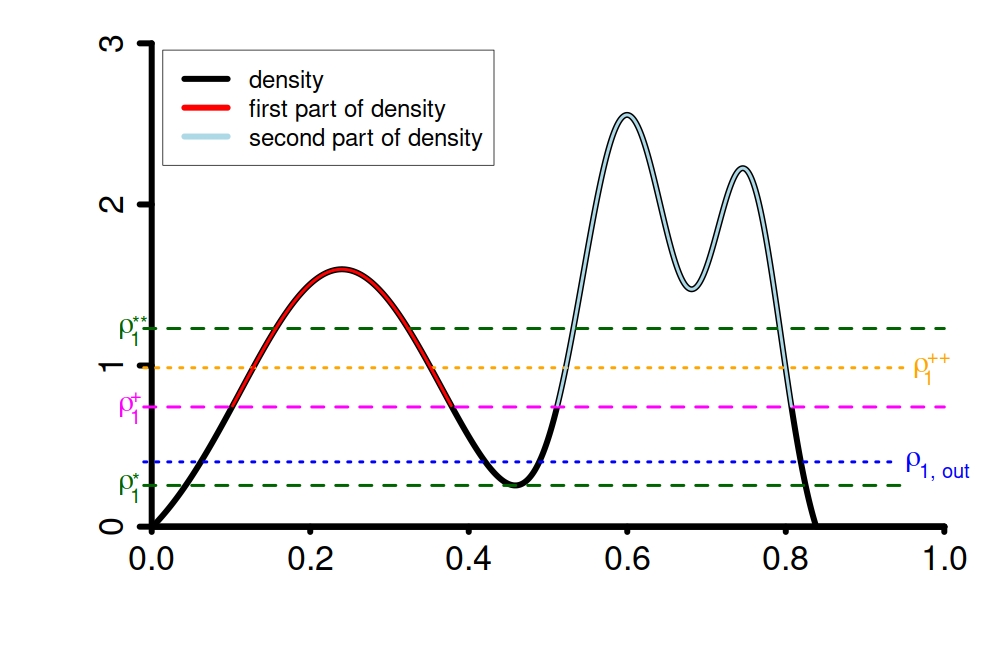}
\hspace*{0.03\textwidth}
\includegraphics[width=0.45\textwidth]{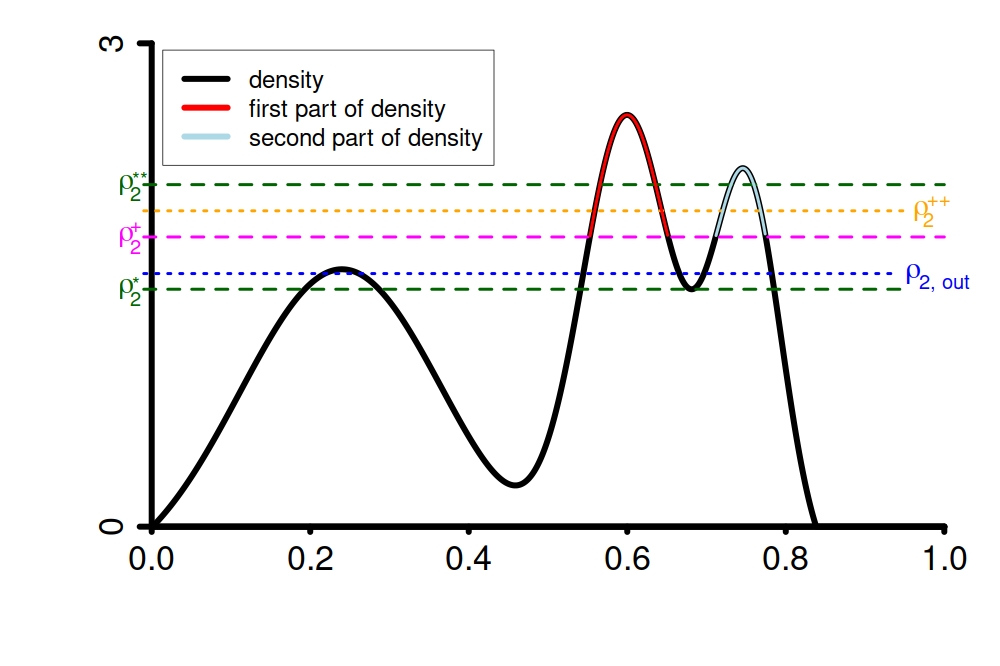}
\caption{A distribution $P$ with depicted density that has a finite split-tree  with minimal step size $\eps>0$ and the resulting guarantees.
\textbf{Left.} Situation at the lowest split level $\r_1^*$. The distribution $P$ can be clustered between
$\r_1^*$ and $\r_1^{**}$ and the 
(unnormalized) densities  of the ``children'' distributions $P_{1,1}$ (red) and $P_{1,2}$ (light blue) above $\rdao := (\rsso + \rso)/2$
 are shown. 
 \assx S is satisfied by $P_{1,1}$, while   $P_{1,2}$ satisfies 
\assx M.
 Algorithm \ref{split-tree-algo-generic} initialized with $\r_0 := 0$ calls Algorithm \ref{cluster-algo-generic} in its Line 1,
 which in turn returns $\routo\approx \rso>0$ and two  corresponding clusters denoted by $B_{1,1}$ and $B_{1,2}$ as guaranteed by Theorem \ref{analysis-main-combined-new}.
 Algorithm \ref{split-tree-algo-generic} stores these in its Line 3 and continues by calling a second instance of
 Algorithm \ref{split-tree-algo-generic} with start level 
 $\routo + \e$ and the familiy $(L_{1,\r})_{\r\geq 0}$ in Line 4. This instance  of  Algorithm \ref{split-tree-algo-generic} in turn 
calls Algorithm \ref{cluster-algo-generic} with $\routo + \e$ and the familiy $(L_{1,\r})_{\r\geq 0}$.
 For $\r\in [\routo, \rddao]$ with $\rddao := 0.75 \rso + 0.25 \rsso$, Part \emph{ii)} of Theorem \ref{result:acting-in-the-gray-zone} ensures that Algorithm \ref{cluster-algo-generic}
 only detects one cluster $B_1^\r$ in its loop, see also the calculations following Theorem \ref{result:acting-in-the-gray-zone}. 
 Consequently, this loop is not left below $\rddao$.
 Moreover, for $\r\geq \rddao$,
 Part \emph{iii)} of Theorem \ref{result:acting-in-the-gray-zone} ensures 
 $\mxmde {1,\r+\e}  \subset L_{1,\r} \subset \mxpde {1,\r-\e}$, and hence Theorem \ref{main-generic-single} shows that Algorithm 
 \ref{cluster-algo-generic} actually returns its start level $\routo + \e$  and $L_{\routo + \e}$. As a consequence, both if-clauses of the second
 instance of  Algorithm \ref{split-tree-algo-generic} are not satisfied and the overall program continues at Line 5 of the first instance of  
  Algorithm \ref{split-tree-algo-generic}. There, a new, third instance of Algorithm \ref{split-tree-algo-generic} is called with 
  $\routo + \e$ and the familiy $(L_{2,\r})_{\r\geq 0}$, which in turn begins by calling Algorithm \ref{cluster-algo-generic} with the same values.
  Again, Theorem \ref{result:acting-in-the-gray-zone} ensures, that for $\r\in [\routo, \rddao]$ Algorithm \ref{cluster-algo-generic}
  only detects one cluster in its loop and that for  $\r\geq \rddao$, the crucial inclusions
  $\mxmde {2,\r+\e}  \subset L_{2,\r} \subset \mxpde {2,\r-\e}$ are satisfied. As a consequence, Theorem 
   \ref{analysis-main-combined-new} ensures that Algorithm \ref{cluster-algo-generic} returns a 
   $\routt\approx \rst> \routo+\e$ with corresponding clusters denoted by $B_{2,1}$ and $B_{2,2}$ to the third instance of  \ref{split-tree-algo-generic}. 
   These values are then stored 
   in the split tree  and a  fourth and fifth instance of
   Algorithm   \ref{split-tree-algo-generic} are called  with $\routt + \e$ and
   the newly defined 
   $(L_{(2,1),\r})_{\r\geq 0}$, respectively  $(L_{(2,2),\r})_{\r\geq 0}$, given by  $L_{(2,i),\r} := L_{2,\rho} \cap B_{2,i}$. 
\textbf{Right.} \assx S is satisfied for both children measures occurring above the split $\rst$. As for $B_1$ on the left image,
the combination of Theorem \ref{result:acting-in-the-gray-zone} and  Theorem \ref{main-generic-single} ensures that 
Algorithm \ref{cluster-algo-generic} called by the fourth and fifth instance of Algorithm \ref{split-tree-algo-generic}
with $\routt + \e$ and
   $(L_{(2,i),\r})_{\r\geq 0}$ returns these values  to the these instances of Algorithm \ref{split-tree-algo-generic}.
   As a consequence, the fourth and fifth instance of  Algorithm \ref{split-tree-algo-generic} return to 
   the third instance of Algorithm \ref{split-tree-algo-generic} without any further action, and in turn the third instance 
   returns to the first instance of Algorithm \ref{split-tree-algo-generic}. 
   This instance then reaches its Line 9, and therefore the overall program terminates with $(\routo, B_1, B_2)$
   and $(\routt, B_{2,1}, B_{2,2})$ being stored in its split tree.
}
\label{figure:split-tree}
\end{figure}

Our next goal is to show that Algorithm \ref{split-tree-algo-generic} can be used to estimate the split-tree 
for distributions having a finite  split-tree with some unknown  minimal step size $\eps>0$. 
To this end, we need Theorem \ref{result:acting-in-the-gray-zone} below, which in its 
  formulation requires the sets
 \begin{align*}
  \ccth{L_\r} :=  \bigl\{ B\in \cct{L_\r}:  B\cap L_{\r+2\e} \neq \emptyset  \bigr\} \, , \qquad \qquad \r\geq 0.
 \end{align*}
Note that this set consists of exactly those $\t$-connected components of $L_\r$ that are identified in 
Lines 4 and 7 of Algorithm \ref{cluster-algo-generic}.

\begin{theorem}\label{result:acting-in-the-gray-zone}
 Let \assx  M  be satisfied. Furthermore, let 
  $\e^*\leq (\rss - \rs)/16$, 
 $\d\in (0,  \dthick]$,  
$\t\in (\psi(\d),\ts(\e^*)]$, and  $\e\in (0 , \e^*]$, and $\r_0 \leq \rs$.
In addition, let   $(L_{\r})_{\r\geq 0}$ be a 
decreasing family satisfying \eqref{uq}
for all $\r\geq \r_0$. 
Finally, let  $\rout$ be the estimate of $\rs$ and $B_1, B_2$ be the cluster estimates returned by  Algorithm \ref{cluster-algo-generic}.
Then the following statements are true:
\begin{enumerate}
 \item We have $|\cct {\mxmde\rss}| = 2$ and the sets $V_1 := \ccr 1\rss\mde$ and $V_2 := \ccr 2\rss\mde$ are the two $\t$-connected 
 components of $\mxmde\rss$.
 \item For all $\r\in [\rout, \rss-3\e]$ we have $|\ccth{L_\r}| = 2$.
 Moreover, we can order the two elements $B_1^\rho$ and $B_2^\rho$ of  $\ccth{L_\r}$
 such that 
 \begin{align}\label{result:acting-in-the-gray-zone-r1}
  V_i \subset B_i^\r \subset B_i\, , \qquad \qquad i=1,2\, .
 \end{align}
 
 \item If $\rda \in [ \rs+\e^* + 6\e, \rss - 5\e]$, then for all $\r\geq \rda + 4\e$ we have $L_{i,\r} \subset B_i^{\rda + 2\e}$
 and
 \begin{align}\label{new-uq}
  \mxmde {i,\r+\e}  \subset L_{i,\r} \subset \mxpde {i,\r-\e}.
 \end{align}
\end{enumerate}
\end{theorem}

To illustrate Theorem \ref{result:acting-in-the-gray-zone}, we now define $\rda := (\rss + \rs)/2$ and assume 
$\e^*\leq (\rss - \rs)/16$. For $\e\in (0,\e^*]$ we then find 
 $\rda \in [ \rs+\e^* + 6\e, \rss - 5\e]$ and  
\begin{displaymath}
 \rda + 4\e \leq  \frac{\rss + \rs}2 +  \frac{\rss - \rs}{4} = \frac {3\rss}{4} + \frac {\rs}{4} =: \rdda\, . 
\end{displaymath}
Then we have, 
$\rss - 3\e > \rss - 4\e^* \geq \rss - (\rss - \rs)/4 =   \rdda$, and therefore  
 part \emph{ii)} of Theorem \ref{result:acting-in-the-gray-zone} shows that for all $\r\in [\rout, \rdda]$
we have \eqref{result:acting-in-the-gray-zone-r1}. Consequently, 
Algorithm  \ref{cluster-algo-generic}, when working with the level sets $(L_{i,\r})_{\r\in [\rout, \rdda]}$,
does  identify exactly one connected component in its Line 3. In other words, the loop between its Lines 2 and 5 is not left
for such $\r$. Moreover, for $\r\geq \rdda \geq \rda + 4\e$
part \emph{iii)} of Theorem \ref{result:acting-in-the-gray-zone} ensures \eqref{new-uq}.
 Consequently, Theorems  \ref{analysis-main-combined-new}  and \ref{main-generic-single} can be applied 
 to Algorithm  \ref{cluster-algo-generic} when working with the level sets $(L_{i,\r})_{\r\geq \rdda}$ for the 
 distribution $P_i$. We refer to Figure \ref{figure:split-tree} for a detailed description of how these guarantees work together.
In summary, these considerations show that Algorithm \ref{split-tree-algo-generic} can be recursively analyzed 
with the help of Theorems  \ref{analysis-main-combined-new}  and \ref{main-generic-single} to show 
that Algorithm \ref{split-tree-algo-generic} indeed estimates the split-tree for 
all distributions $P$ having a finite split-tree with some unknown minimal step size $\eps>0$. In particular, for all quantitative 
guarantees 
it actually suffices to describe the behavior of Algorithm \ref{cluster-algo-generic} for distributions satisfying 
\assx S and \assx M. This insight will be adopted later in the statistical analysis of  Section \ref{sec:algo}.

\section{Uncertainty Control for Kernel Density Estimators}\label{sec:kde}

The results of Section \ref{sec:smg} provide 
guarantees for Algorithm \ref{cluster-algo-generic} as soon as the input level sets 
satisfy \eqref{uq}. In \cite{Steinwart15a} it has been shown that guarantees of the 
form \eqref{uq} can be established for the level sets of histogram-based density estimators.
The goal of this section is to show that \eqref{uq} can also be established for a variety of 
kernel density estimators.

Our first definition introduces the kernels we are considering in the following.

\begin{definition}\label{kde}
   A bounded, measurable function $K:\R^d\to [0,\infty)$ is called symmetric kernel, if 
$K(x)>0$ in some neighborhood of $0$,
$K(x) = K(-x)$ for all $x\in \Rd$, and 
\begin{equation}\label{int-kern}
   \int_{\Rd} K(x) \dld(x) = 1\, .
\end{equation}
For $\d>0$ we write $K_\d  := \d^{-d} K(\d^{-1}\, \cdot\,)$, and for 
$r>0$ and a norm $\snorm\cdot$ on $\Rd$ we define
\begin{align*}
	\kto r &:= \int_{\Rd\setminus B(0,r)} K(x) \,\dld(x)\, , &
	\kti r &:= \sup_{x \in \Rd\setminus B(0,r)} K(x) \, .
\end{align*}
Moreover,   $\ktof$ and $\ktif$ are called tail functions. Finally, we say that $K$ has:
\begin{enumerate}
   \item a bounded support if $\supp K\subset B_{\snorm\cdot}$.
	\item an exponential tail behavior, if  there 
exists a constant $c>0$ such
 that 
 \begin{equation}\label{lem:tail-functions-h1}
  K(x) \leq c \exp\bigl(-\tnorm x\bigr)\, , \qquad\qquad x\in \Rd.
 \end{equation}
\end{enumerate}
\end{definition}

Note that kernels of the form $K(x) = k(\snorm x)$ are always symmetric and if the representing function 
$k:[0,\infty)\to [0,\infty)$ is bounded and measurable, so is $K$. Moreover, if $k(r)>0$ for all $r\in [0,\eps)$, 
where $\eps>0$ is some constant, then we further have 
$K(x)>0$ in some neighborhood of $0$. The integrability condition \eqref{int-kern} is standard for kernel density estimators,
and for kernels of the form $K(x) = k(\snorm x)$ it can be translated into 
a condition on $k$.
In particular, for $k=c\eins_{[0,1]}$  we obtain
the ``rectangular window kernel'', which is
 a symmetric kernel with bounded support in the sense of 
Definition \ref{kde} and if $k$ is of the form $k(r) = c\exp(-r^2)$ or $k(r) = c\exp(-r)$, then 
we obtain a symmetric kernel with exponential tail behavior. Examples of the latter are 
Gaussian kernels, while 
the triangular, the Epanechnikov, the quartic, the triweight, and the 
tricube kernels are further examples of symmetric kernels with bounded support.
Finally note that each symmetric kernel with bounded support also has exponential tail behavior,
since we always assume that $K$ is bounded.

Before we proceed with our main goal of establishing \eqref{uq} let us briefly discuss a couple of
simple properties of symmetric kernels $K$ in the sense of Definition \ref{kde}. To this end, we first note that
the properties of the 
 Lebesgue measure $\lbd$ 
ensure
that 
\begin{equation}\label{shift-int-kernel-new}
     \int_{\Rd}K_\d({x-y}  ) \dld(y) = \int_{\Rd}K ({x-y}) \dld(y)  = \int_{\Rd}K ({y-x}) \dld(y)= 1
\end{equation}
for all $x\in \Rd$, $\d>0$, and then by an analogous calculation we obtain 
\begin{equation} \label{inclusion-aux-conversion}
    \int_{\Rd\setminus B(x,\s)} K_\d(x-y) \,\dld(y)
     = \int_{\Rd\setminus B(0,\s / \d)} K(y) \,\dld(y)  
    = \kto{\tfrac\s\d}
    \,.
\end{equation}
In addition, we always have $\kto r\to 0$ for $r\to \infty$ and if $K$ has bounded support, then 
the tail functions with respect to this norm 
satisfy
\begin{equation}\label{kappa-for-bounded-support}
   \kto r = \kti r = 0\, , \qquad \qquad r\geq 1\, .
\end{equation}
The following lemma considers the behavior of the tail functions for kernels with exponential tail behavior.

\begin{lemma}\label{lem:tail-functions}
 Let 
 $K:\R^d\to [0,\infty)$ be a  symmetric, kernel 
 with  exponential tail behavior \eqref{lem:tail-functions-h1}.
Then, for all $r\geq 0$,  the corresponding tail functions satisfy
\begin{align*}
 \kto r & \leq    c d^2 \vold  e^{-r} r^{d-1}           \\
 \kti r & \leq    c  e^{-r}        \, .
\end{align*}
\end{lemma}


Now, let $K$ be a symmetric kernel on $\Rd$ and   $P$ be a 
distribution  on $\Rd$.
For $\d>0$ we then define 
the infinite-sample kernel density estimator
$\hpd:\Rd\to [0,\infty)$ by 
\begin{displaymath}
   \hpd(x):=\delta^{-d}\int_{\R^d}K\Bigl(\frac{x-y}{\delta}\Bigr)\dP(y)\, \qquad \qquad x\in \Rd.
\end{displaymath}
It is easy to see that $\hpd$ is a bounded measurable function with $\inorm\hpd \leq \d^{-d} \inorm K$.
Moreover, a quick application of Tonelli's theorem together with \eqref{shift-int-kernel-new} yields
\begin{displaymath}
   \int_{\Rd} \hpd\dld = \int_\Rd \delta^{-d}\int_{\R^d}K\Bigl(\frac{x-y}{\delta}\Bigr) \dld(x) \dP(y)
= \int_\Rd \eins_{\Rd}(y) \dP(y) = 1\, ,
\end{displaymath}
and hence $\hpd$ is a Lebesgue probability density. Moreover, if $P$ has a Lebesgue density $h$,
then it is well-known, see e.g.~\cite[Theorem 9.1]{DeLu01}, 
that $\snorm{\hpd -h}_{\Lx 1 \lbd} \to 0$ for $\d\to 0$. In addition, if this density is bounded,
then \eqref{shift-int-kernel-new} yields
\begin{eqnarray}\label{bound-hdd}
   \inorm\hpd &=& \sup_{x\in \Rd} \delta^{-d}\int_{\R^d}K\Bigl(\frac{x-y}{\delta}\Bigr) h(y)\dld(y)\nonumber\\
&\leq& \inorm h \sup_{x\in \Rd}  \int_{\Rd}K_\d({x-y}  ) \dld(y) = \inorm h\, .
\end{eqnarray}
Clearly, if $D= (x_1,\dots,x_n)\in X^n$ is a data set, we can consider the corresponding
empirical measure $\frac 1 n \sum_{i=1}^n \d_{x_i}$, where $\d_{x}$ denotes the Dirac measure at   $x$.
In a slight abuse of notation we  also denote this empirical measure  by $D$.
The resulting 
function 
$h_{D,\delta}:\Rd\to \R$,  called kernel density estimator (KDE),
 can then be computed by 
\begin{displaymath}
   h_{D,\delta}(x):=\frac{1}{n\delta^d}\sum^n_{i=1}K\left(\frac{x-x_i}{
\delta}\right)\, , \qquad \qquad x\in \R.
\end{displaymath}
Now, one way to define level set estimates with the help of $\hdd$ is a simple plug-in approach, that is 
\begin{equation}\label{naiv-ls}
   L_{D,\r} := \{\hdd\geq \r\}\, .
\end{equation}
One can show that from a theoretical perspective, this level set estimator is perfectly fine. Unfortunately,
however, it is computationally intractable. For example, if $\hdd$ is a moving window estimator, that is 
$K(x) = c \eins_{[0,1]}(\snorm x)$ for $x\in \Rd$, then the up to $2^n$ different level sets 
\eqref{naiv-ls} are generated by intersection of balls around the samples, and 
the structure of these
intersections may be too complicated to compute  $\t$-connected components  in Algorithm
\ref{cluster-algo-generic}. For this reason, we   consider level set estimates of the form 
\begin{align}\label{Lrho}
   L_{D,\r} := \{x\in D:\hdd(x) \geq \r\}\pds\, ,
\end{align}
where $\s>0$. Note that computing connected components of \eqref{Lrho} is indeed feasible,
since it amounts to computing the connected components of the neighborhood graph, in which 
two vertices $x_i$ and $x_j$ with $i\neq j$ have an edge if $\snorm{x_i-x_j} \leq \s+\t$. In particular,
DBSCAN can be viewed as such a strategy for the moving window kernel. 

With these preparations we can now present our first result that establishes a sort of 
uncertainty control \eqref{uq} for level set estimates of the form \eqref{Lrho}.

\begin{theorem}\label{include-main-thm-new}
Let $\snorm\cdot$ be some norm on $\Rd$, 
$K:\R^d\to [0,\infty)$ be a symmetric kernel, and $\ktof$ and $\ktif$ be its associated tail functions.
Moreover, let $P$ be a distribution for which \assx P is satisfied, 
%
and $D$ be a 
data set such that the corresponding KDE satisfies $\inorm{\hdd-\hpd}<\varepsilon$ for some $\varepsilon>0$ and $\d>0$.
For $\r\geq 0$ and $\s>0$ we define 
\begin{align*}
   L_{D,\r} := \{x\in D:\hdd(x) \geq \r\}\pds
\end{align*}
and $\eps := \max\{\rho\kto{\tfrac\s\d},   \d^{-d} \kti{\tfrac\s\d}  \} $.
Then, for all  $\rho\ge \d^{-d} \kti{\tfrac \s\d}$, we have
\begin{align}\label{connect-main}
   M\mdds_{\r+\e+\eps} \subset L_{D,\r} \subset  M\pdds_{\r-\e-\eps}\, .
\end{align}
Moreover, if $P$ has a bounded density $h$, then \eqref{connect-main} also holds for $\eps = \inorm h \kto{\tfrac\s\d}$.
\end{theorem}

Note that for 
kernels $K$ having bounded support   for the norm considered in Theorem \ref{include-main-thm-new}, 
 Equation \eqref{kappa-for-bounded-support}
shows that \eqref{connect-main} actually holds for $\eps=0$ and all $\r\geq 0$ and all $\s\geq \d$.
Therefore, we have indeed \eqref{uq} for $\d$ replaced by $2\s$. In general, however, we have an 
additional horizontal uncertainty $\eps$ that of course affects the guarantees of Theorem \ref{analysis-main-combined-new}.
To control this influence, our strategy will be to ensure that $\eps\leq \e$, which in view of 
$\eps = \inorm h \kto{\tfrac\s\d}$ means that we need to have an upper bound on $\ktof$ and $\s$.


Theorem \ref{include-main-thm-new} tells us that 
the uncertainty control \eqref{connect-main} is satisfied as soon as we have a data set $D$ with 
$\inorm{\hdd-\hpd}<\varepsilon$. Therefore, our next goal is to establish 
such an estimate
 with high probability. 
Before we begin we like to mention that rates for $\inorm{\hdd-\hpd}\to 0$ have already been proven in 
\cite{GiGu02a}. However, those rates only hold for $n\geq n_0$, where $n_0$, although it 
almost surely exists,
 may actually 
depend on the data set $D$. In addition one is required to choose a sequence $(\d_n)$ of bandwidths 
a-priori. To apply the theory developed in \cite{Steinwart15a} including the 
adaptivity, however, we need  bounds of the 
form $\inorm{\hdd-\hpd}<\varepsilon(\d,n,\vs)$ that hold with probability not smaller than $1-e^{-\vs}$.
For these reasons, the results of 
\cite{GiGu02a} are not suitable for our purposes. 

To establish the bounds described above, we need 
 to recall some notions first.

\begin{definition}
Let $E$ be a Banach space and $A\subset E$ be a bounded subset.
Then, for all $\e>0$,  the covering numbers of $A$ are defined by 
\begin{displaymath}
   \ca N(A,\snorm\cdot_E,\e)
   :=
   \inf\biggl\{
n \geq 1: \exists\, x_{1},\dots,x_{n}\in E \mbox{ such that } A\subset \bigcup_{i=1}^{n} (x_i + \e B_{\snorm\cdot})
\biggr\} \, ,
\end{displaymath}
where $\inf\emptyset := \infty$. Furthermore, we  use the notation $\ca N(A,E,\e) := \ca N(A,\snorm\cdot_E,\e)$.
\end{definition}

We now introduce the kind of covering number bound we will use in our analysis.

\begin{definition}
   Let $(Z,P)$ be a probability space and $\eu{G}$ be a set of bounded  measurable
functions from $Z$ to $\R$ for which there exists a   $B>0$ such that 
$\inorm g\leq B$ for all $g\in \eu G$. Then 
$\eu{G}$
is called a  uniformly bounded VC-class, if there exist
 $A>0$ and $\nu>0$ such that, for every probability measure $P$ on
$Z$ and every $0<\eps\leq B$, the covering numbers satisfy 
\begin{align}\label{VC-ineq}
   \ca N(\eu{G},L_2(P),\eps)\le\left(\frac{AB}{\eps}\right)^\nu.
\end{align}
\end{definition}

Before we proceed, let us briefly look at two important sufficient criteria for 
ensuring that the set of functions
	\begin{align}\label{def-Kd-class}
	   \eu K_\d := \bigl\{  K_\d(x-\cdot) : x\in X\bigl\}
	\end{align}
is a  uniformly bounded VC-class. 
The first result in this direction considers kernels used in moving window estimates.

\begin{lemma}\label{indicator}
   Consider the kernel  $K=c \eins_{B_{\snorm\cdot}}$,  where $\snorm\cdot$ is either the 
   Euclidean- or the supremum norm. Then for all $\d>0$ the set 
	$\eu K_\d$ defined by \eqref{def-Kd-class}
	is a uniformly bounded VC-class with $B:= \d^{-d}\inorm K = \d^{-d} c$ and $A$ and $\nu$ being  independent of $\d$.
\end{lemma}

%
%
%

The next lemma shows that H\"older continuous kernels also induce a
uniformly bounded VC-class $\eu K_\d$, provided that the input space $X$  in \eqref{def-Kd-class} is compact. For 
its formulation we need to recall that for every norm $\snorm\cdot$ on $\Rd$ and 
 every compact subset $X\subset \Rd$ there exists a finite  constant $C_{\snorme}(X)>0$ such that 
for all $0<\e\leq \diam_\snorme (X)$ we have 
\begin{align}\label{cover-x}
   \ca N (X, \snorme, \eps) \leq C_\snorme (X) \eps^{-d}\, .
\end{align}
We can now formulate the announced result for H\"older-continuous kernels.

\begin{lemma}\label{lip-ex}
   Let $K:\Rd\to [0,\infty)$ be a symmetric  kernel  that is   $\a$-H\"older continuous. For some arbitrary but fixed 
	norm $\snorm \cdot$ on $\Rd$ we write 
 $|K|_\a$ for the corresponding  $\a$-H\"older constant. Moreover, 
	let $X\subset \Rd$ be a compact subset and $\eu K_\d$ defined by \eqref{def-Kd-class}.
	Then for all $\d>0$ with $\d\leq \bigl(\frac{|K|_\a}{\inorm K}\bigr)^{1/\a} \diam_\snorme (X)$, 
	all $0<\eps \leq \d^{-d} \inorm K$,
	and all distributions $P$ on $\Rd$ we have 
	\begin{align}\label{lip-ex-h1}
	   \ca N\bigl( \eu K_\d, \Lx 2 P, \eps\bigr) \leq  C_\snorme (X) \biggl(  \frac{|K|_\a}{\d^{\a+d}\eps}   \biggr)^{d/\a}\, .
	\end{align}
In particular, $\eu K_\d$ is a uniformly bounded VC-class with $\nu:= d/\a$, $A:=  (C_\snorme (X))^{\a/d} |K|_\a \snorm K_\infty^{-1}\d^{-\a} $ and
$B:= \d^{-d} \inorm K$.
\end{lemma}

Now that we have collected sufficiently many examples of kernels for which $\eu K_\d$ is a 
uniformly bounded VC-class, we can now present the second main result of this section that 
establishes a finite sample bound  $\inorm{\hdd-\hpd}<\varepsilon(\d,n,\vs)$. 


\begin{theorem}\label{approximate-thm}
Let $X\subset \Rd$ and 
 $P$ be  distribution on 
 $X$ that has a Lebesgue density $h\in \Lx 1 {\Rd} \cap \Lx p{\Rd}$
 for some $p\in (1,\infty]$.
 We write  $\frac 1 p + \frac 1 {p'} = 1$. 
 Moreover, 
 let $K:\Rd\to [0,\infty)$ be a symmetric kernel for which 
 there is a $\d_0\in (0,1]$ such that for all $\d\in (0,\d_0]$ the set 
$\eu K_\d$ 
defined in \eqref{def-Kd-class}
is a uniformly bounded VC-class with constants of the form $B_\d= \d^{-d}\inorm K$,
$A_\d = A_0\d^{-a}$, and $A_0>0, a\geq 0, \nu\geq 1$ being independent of $\d$, that is,
\begin{align}\label{target-cover-bound}
   \ca N\bigl( \eu K_\d, \Lx 2 Q, \eps\bigr) \leq   \biggl( \frac{A_0 \inorm K \d^{-(d+a)} }{\eps}   \biggr)^\nu 
\end{align}
holds for all  $\d\in (0,\d_0]$, all $\eps\in( 0,B_\d]$, and all distributions $Q$ on $\Rd$.
%
%
%
Then, there
exists  a   $C>0$ only depending on
$d$, $p$, and $K$ 
such that, for all
$n\geq 1$, $\d>0$,  and $\vs\geq 1$
satisfying 
\begin{align}\label{extra-d-cond}
 \d  \leq \min\biggl\{ \d_0,  \frac {4^{p'} \inorm K}  {\snorm h_p^{p'}} , \frac{\snorm h_p^{\frac 1 {2a + d/p'}}}  C\biggr\}  
 \qquad \mbox{ and } \qquad   
 \frac{|\!\log \d |}{n\d^{d/p'}}  \leq \frac{\snorm h_p}{C \vs} 
\end{align}
we have 
\begin{align}\label{simplified-conc}
  P^n\bigg(\bigg\{D:\,
  \snorm{\hdd-\hpd}_{\ell_\infty(X)}
  <
C \sqrt{\frac{\snorm h_p \,|\!\log \d| \, \vs }{n\d^{d(1+1/p)}}}
\bigg\}\bigg)\ge
1-e^{-\vs}.
\end{align}
\end{theorem}

For bounded densities,
Theorem \ref{approximate-thm} recovers the same rates as \cite{GiGu02a}. However, 
\cite{GiGu02a}  established the rates in an 
almost sure asymptotic form, whereas Theorem \ref{approximate-thm} provides a 
finite sample bound. Moreover, unlike \cite{GiGu02a}, Theorem \ref{approximate-thm}
also yields rates for unbounded densities.

\section{Statistical Analysis of KDE-based Clustering}\label{sec:algo}

In this section we combine the generic results of Section \ref{sec:smg}
with the uncertainty control for level set estimates obtained from kernel density estimates
we obtained in Section \ref{sec:kde}. As a result we will present  finite sample guarantees, consistency results, and 
rates for estimating split levels $\rs$ and the corresponding    clusters.
In this respect recall the  discussion following Theorem \ref{result:acting-in-the-gray-zone},
which showed that for deriving guarantees for estimating the split tree with the help of Algorithm \ref{split-tree-algo-generic}
it actually suffices to analyze the behavior of Algorithm \ref{cluster-algo-generic}
for distributions satisfying \assx S and \assx M. Following this insight, we will focus on such guarantees for 
Algorithm \ref{cluster-algo-generic} in this section.

Our first result presents   finite sample bounds for estimating both $\rs$ and the single or multiple 
 clusters with the help of 
Algorithm \ref{cluster-algo-generic}.
To treat kernels with bounded and unbounded 
support simultaneously, we restrict ourselves to the case of bounded densities, but at least for 
kernels with bounded support an adaption to $p$-integrable densities is straightforward as we discuss below.

\begin{theorem}\label{analysis-main2-new}
Let $P$ be a distribution for which \assx P is satisfied and 
whose Lebesgue density is bounded.
Moreover, consider a symmetric kernel $K:\R^d\to [0,\infty)$
with exponential tail behavior,
for which the assumptions of Theorem \ref{approximate-thm} hold.
For fixed  $\d\in (0,\eul^{-1}]$ and $\t>0$, we choose a $\s>0$ with 
\begin{equation}\label{def-sigma}
 \s \geq 
 \begin{cases}
  \d & \mbox{ if } \supp K\subset B_{\snorm\cdot}\, ,\\
  \d \, |\log \d|^2 & \mbox{ otherwise.}
 \end{cases}
\end{equation}
and assume this $\s$ further satisfies both  $\s\leq \dthick/2$ and $\t > \psi(2\s)$. Moreover, for fixed 
$\vs\geq 1$,   $n\geq 1$
satisfying the assumptions \eqref{extra-d-cond}, we pick an $\e>0$ satisfying the bound
\begin{equation}\label{analysis-main2-h0}
  \e \geq \frac C2\sqrt{\frac{\inorm h \,|\!\log \d| \, \vs }{n\d^{d}}}\, ,
  \end{equation}
and if $K$ does not have bounded support, 
also
\begin{equation}\label{analysis-main2-h0-add}
 \e \geq \max\bigl\{1, 2 d^2 \vold\} \cdot c   \cdot \d^{|\log \d|-d}\, .
\end{equation} 
Now assume that 
 for each data set $D\in X^n$ sampled from $P^n$,
we feed Algorithm \ref{cluster-algo-generic} with the level set estimators $(L_{D,\r})_{\r\geq 0}$ given by \eqref{Lrho}, 
the parameters  $\t$ and $\e$,
 and a start level $\r_0\geq \e$.
Then the following statements are true:
\begin{enumerate}
 \item If $P$ satisfies \assx S and $\r_0 \geq \rls$, then 
 with probability $P^n$  not less than $1-e^{-\vs}$
 Algorithm \ref{cluster-algo-generic} returns $\r_0$ and $L_0$ and we have 
\begin{equation}\label{single_cluster-symdif}
   \mu\bigl(L_{\r_0} \symdif \hat M_\rls  \bigr) \leq \mu \bigl( M_{\r_0-\e}\pdds \setminus \hat M_\rls\bigr)
 +  \mu \bigl( \hat M_\rls\setminus  M_{\r_0+\e}\mdds \bigr) \, ,
\end{equation}
	where $\hat M_\rls := \bigcup_{\r>\rls}\mr$.
 \item If $P$ satisfies \assx M and we have an 
  \begin{equation}\label{estar}
     \e^* \geq  \e + \inf \bigl\{\e'\in  (0,\rss-\rs] : \t^*(\e') \geq \t   \bigr\}\, .
 \end{equation}
  with $9\e^* \leq \rss-\rs$, then with probability $P^n$  not less than $1-e^{-\vs}$, we have a
  $D\in X^n$ such that 
the following statements are true for Algorithm \ref{cluster-algo-generic}:
\begin{enumerate}
 \item The returned level $\routD$ satisfies both $\routD \in [\rs+2\e , \rs+\e^*+5\e]$
  and 
  \begin{equation*}
     \t-\psi(2\s) < 3\t^*\bigl(\routD-\rs+\e\bigr)\, .
  \end{equation*}
  \item Two sets $B_1(D)$ and $B_2(D)$ are returned and these  
   can be ordered such that for $A^i_{\routD +\e} \in \ca C(M_{\routD+\e})$ 
   ordered in the sense of  $A^i_{\routD+\e}\subset A_i^*$
   we have 
          \begin{equation}\label{cluster-chunk-kde-bound}
   \sum_{i=1}^2 \mu\bigl(B_i(D) \symdif A_i^*\bigr)   
   \leq 
 2\sum_{i=1}^2 \mu \bigl(A_i^* \!\setminus\!  (A^i_{\routD+\e})\mdds\bigr)  
+ \mu\bigl ( M_{\routD-\e}\pdds \!\setminus\! \{ h>\rs \}\bigr)\, .
\end{equation}
\end{enumerate}  
\end{enumerate}
\end{theorem}

For our subsequent asymptotic analysis we note that
the assumptions $\d\in (0,\eul^{-1}]$ and $\vs\geq 1$ of Theorem \ref{analysis-main2-new}
show that \eqref{analysis-main2-h0-add} is satisfied if 
\begin{equation}\label{simplified-add-ass}
   \max\bigl\{1, 2 d^2 \vold\} \cdot c   \cdot \d^{|\log \d|+d/2} 
\leq 
\frac C2\sqrt{\frac{\inorm h  }{n}}\, , 
\end{equation}
and if we choose $\d$ in terms of $n$, i.e.,
$\d=\d_n$, then \eqref{simplified-add-ass} is satisfied for   large $n$ if 
  $\d_n \in O(n^{-a})$ for some 
small $a>0$.  We shall see below, that such rates for $\d_n$ are typical.

If we have a kernel with bounded support, then a variant of Theorem \ref{analysis-main2-new}
also holds for unbounded densities. Indeed, if we have a density  $h\in \Lx 1 {\Rd} \cap \Lx p{\Rd}$
for some $p\in (1,\infty]$, then all conclusions of Theorem \ref{analysis-main2-new} remain valid,
if we replace \eqref{analysis-main2-h0} by
\begin{displaymath}
    \e \geq \frac C2\sqrt{\frac{\snorm h_p \,|\!\log \d| \, \vs }{n\d^{d(1+1/p)}}}\, .
\end{displaymath}
Note that for such kernels the additional assumption \eqref{analysis-main2-h0-add} is not necessary.

While \eqref{def-sigma} only provides a lower bound on possible values for $\s$, Theorem \ref{analysis-main2-new}
actually indicates that $\s$ should not be chosen significantly larger than these lower bounds, either.
Indeed, the choice of $\s$ also implies a minimal value for $\t$ by the condition
$\t>\psi(2\s)$, which in turn influences $\e^*$ by \eqref{estar}. Namely, larger values of $\s$
lead to larger $\t$-values and therefore to larger values for $\e^*$.
As a result, the guarantees in \emph{(a)} become weaker, and in addition, larger values of $\s$
also lead to weaker guarantees in \emph{(b)}. For a similar  reason we do not consider
kernels $K$ with heavier tails than \eqref{lem:tail-functions-h1}. Indeed, if $K$ only has a 
polynomial upper bound for its tail, i.e., there are constants $c$ and $\a>d$ with
\begin{displaymath}
   K(x) \leq c\cdot  \tnorm x^{-\a} \, , \qquad \qquad x\in \Rd\, ,
\end{displaymath}
then $\kto r \preceq r^{-\a+d}$  and $\kti r \preceq r^{-\a}$.
Now, if we picked $\s = \d|\log\d|^b$ for some $b>0$, then we would need to 
replace \eqref{analysis-main2-h0-add} by a bound of the form 
 $  \tilde c \d^{-d} |\log\d|^{-\a b} \leq \e$,
and this would rule out $\e\to 0$ for $\d\to 0$. As a result, no rates would be possible.
Now, one could address this by choosing $\s := \d^{b}$ for some $b\in (0,1)$, which in turn would 
require a bound of the form 
 $  \tilde c \cdot  \d^{\a(1-b)-d} \leq \e$,
instead of  \eqref{analysis-main2-h0-add}. Arguing as around \eqref{simplified-add-ass}
this is guaranteed if 
\begin{displaymath}
   \tilde c \d^{\a(1-b)-d/2}  \leq \frac C2\sqrt{\frac{\inorm h  }{n}}\, ,
\end{displaymath}
and if $\d\to 0$ the latter would actually require $b<1 - \tfrac d {2\a}$. In particular, $b$ would be strictly 
bounded away from $1$. However, such a choice for $\s$ would significantly weaken the 
guarantees given in \emph{(a)} and \emph{(b)} as explained above, and as a consequence, the rates 
obtained below would be worse.
Note that from a high-level perspective this phenomenon is not surprising: indeed, heavier tails 
smooth out the infinite sample density estimator $\hpd$ and as consequence, the uncertainty 
guarantees \eqref{connect-main} become worse in the horizontal direction, that is, we get more blurry
estimates $L_{D,\r}$ of $\mr$. However, for the detection of connected components at 
a level $\r$, less blurry estimates are preferable.

In the remainder of this section, we illustrate how the finite sample guarantee of 
Theorem \ref{analysis-main2-new} can be used to derive both consistency and rates. 
To deal with kernels with unbounded support we restrict our considerations to the case of bounded
densities, but it is straightforward to obtain results for unbounded densities if one restricts considerations 
to kernels with bounded support as already indicated above.

%
%
 
 \begin{corollary}\label{cor:consis}
Let $P$ be a distribution satisfying \assx P and 
whose Lebesgue density is bounded.
Moreover, consider a symmetric kernel $K:\R^d\to [0,\infty)$
with exponential tail behavior,
for which the assumptions of Theorem \ref{approximate-thm} hold.
 Let $(\d_n)$ be a  positive sequence with $\d_n \preceq n^{-a}$ for some $a>0$ and 
pick a sequence $(\s_n)$ converging to zero and  satisfying \eqref{def-sigma} for all
sufficiently large $n$. Moreover, let 
$(\e_n)$ and $(\t_n)$ be   positive sequences converging to zero such that 
%
$\psi(2\s_n) < \t_n$ for all sufficiently large $n$, and 
\begin{equation}\label{cons-assump}
 \lim_{n\to \infty} \frac {  \log \d_n^{-1}} {n \e_n^2 \d_n^{d}}  = 0\, .
\end{equation}
Now assume that 
 for each data set $D\in X^n$ sampled from $P^n$,
we feed Algorithm \ref{cluster-algo-generic} with the level set estimators $(L_{D,\r})_{\r\geq 0}$ given by \eqref{Lrho}, 
the parameters  $\t_n$ and $\e_n$,
 and the start level $\r_0 := \e_n$.
Then the following statements are true:
\begin{enumerate}
   \item If $P$ satisfies \assx S with $\rls=0$, then  for all $\eps>0$ we have 
 \begin{displaymath}
    \lim_{n\to \infty}P^n\Bigl( \bigl\{  D\in X^n: 0< \routD \leq \eps \bigr\}\Bigr) = 1\, ,
 \end{displaymath}
and if $\mu(\overline{\{h>0 \}}\setminus\{h>0\}) = 0$
 we also have
\begin{displaymath}
   \lim_{n\to \infty}P^n\Bigl( \bigl\{  D\in X^n: \mu( L_{D,\routD}\symdif \{h>0\})  \leq \eps \bigr\}\Bigr) = 1\, ,
\end{displaymath}
\item  If $P$ satisfies \assx M, then, for all $\eps>0$, we have 
 \begin{displaymath}
    \lim_{n\to \infty}P^n\Bigl( \bigl\{  D\in X^n: 0< \rds- \rs \leq \eps \bigr\}\Bigr) = 1\, ,
 \end{displaymath}
and, if $\mu(\overline{A_i^* \cup A_2^*}\setminus (A_1^* \cup A_2^*)) = 0$, we also have, for 
$B_1(D)$, $B_2(D)$ as in \eqref{cluster-chunk-kde-bound}:
\begin{displaymath}
   \lim_{n\to \infty}P^n\Bigl( \bigl\{  D\in X^n: \mu(B_1(D) \symdif A_1^*)  + \mu(B_2(D) \symdif A_2^*) \leq \eps \bigr\}\Bigr) = 1\, .
\end{displaymath}
\end{enumerate}
\end{corollary}


 Our next goal is to establish rates of convergence for estimating $\rs$ and the clusters.
 We begin with a result providing a rate of $\rds\to \rs$. To this end we need to recall the following
definition from \cite{Steinwart15a} that describes how well the clusters are separated above $\rs$.
 
\begin{definition}
Let \assx M be satisfied.
 Then the clusters of $P$ have a separation exponent $\k\in (0,\infty]$, if there is a constant
 $\csepl>0$  such that 
\begin{displaymath}
 \t^*(\e) \geq \csepl\, \e^{1/\k}
\end{displaymath}
for all   $\e\in (0,\rss-\rs]$.
Moreover,  the separation exponent $\k$ is exact, if 
there exists another
constant $\csepu>0$ such that,
for all $\e\in (0,\rss-\rs]$, we  have 
\begin{displaymath}
 \t^*(\e) \leq \csepu\, \e^{1/\k}\, .
\end{displaymath}
\end{definition}

The separation exponent  describes how fast the connected components
of the  
$M_\r$ approach each other for $\r\searrow\rs$. Note that 
a distribution having separation exponent $\k$ also has
separation exponent $\k'$ for all $\k'<\k$. 
In particular, the ``best'' separation exponent is $\k=\infty$ and this 
exponent  describes distributions,
for which we have $d(A_1^*, A_2^*)\geq \csepl$, 
i.e.~the clusters $A_1^*$ and $A_2^*$ do not touch each other.

The separation exponent makes it possible to find a good value
for $\e^*$ in Theorem \ref{analysis-main2-new}. Indeed, the proof of 
Theorem \cite[Theorem 4.3]{Steinwart15a} shows that for given $\e$ and $\t$,
the value 
\begin{displaymath}
   \e^* := \e + (\t /\csepl)^\kappa
\end{displaymath}
satisfies \eqref{estar} as soon as we have $9\e^* \leq \rss-\rs$. Consequently, 
the bound in part \emph{ii) (a)} of Theorem \ref{analysis-main2-new} 
becomes
\begin{equation}\label{new-rd-diff}
   2\e \leq \routD-\rs \leq 6\e +\Bigl( \frac{\t}{\csepl}\Bigr)^\k
\end{equation}
if we have a separation exponent $\k\in (0,\infty]$. Moreover, if the separation exponent $\k\in (0,\infty)$  is exact and 
we choose $\t \geq 2\psi(2\s)$, then 
\eqref{new-rd-diff} can be improved to 
\begin{equation}\label{new-rd-diff-impr}
 \e+  \frac 1 4 \Bigl( \frac{\t}{6\csepu}\Bigr)^\k  \leq \routD-\rs \leq 6\e +\Bigl( \frac{\t}{\csepl}\Bigr)^\k
\end{equation}
as the proof of 
Theorem \cite[Theorem 4.3]{Steinwart15a} shows. In order to establish rates, it therefore suffices 
to find null sequences 
 $(\e_n)$, $(\d_n)$, $(\s_n)$,  and $(\t_n)$ that satisfy \eqref{def-sigma} and \eqref{analysis-main2-h0}, and additionally
$\d_n\in \ca O(n^{-a})$ for some $a>0$,  if $K$ does not have bounded support. The following corollary
presents resulting rates of this approach that are, modulo logarithmic terms, the best ones we can obtain from this approach.

\begin{corollary}\label{rates-cor1}
Let $P$ be a distribution for which \assx M is satisfied and 
whose Lebesgue density is bounded.
Moreover, consider a symmetric kernel $K:\R^d\to [0,\infty)$
with exponential tail behavior,
for which the assumptions of Theorem \ref{approximate-thm} hold.
In addition, assume that 
  the clusters of $P$
 have separation exponent $\k\in (0,\infty)$. Furthermore, let
 $(\e_n)$, $(\d_n)$, $(\s_n)$,  and $(\t_n)$ be  sequences with 
   \begin{align*}
      \e_n &\sim  \Bigl( \frac {(\log n)^3\cdot  \log\log n}n  \Bigr)^{\frac {\g \k}{2\g \k +  d}} \, ,
       &\d_n &\sim  \Bigl( \frac {\log n}n  \Bigr)^{\frac {1}{2\g \k +  d}}\, , \\
 \s_n & \sim  \Bigl( \frac {(\log n)^3}n  \Bigr)^{\frac {1}{2\g \k +  d}}\, ,    
       &\t_n &\sim  \Bigl( \frac {(\log n)^3\cdot  \log\log n}n  \Bigr)^{\frac {\g }{2\g \k +  d}} \, , 
   \end{align*}
   and assume that, for $n\geq 1$
 $D\in X^n$ sampled from $P^n$,
we feed Algorithm \ref{cluster-algo-generic} with the level set estimators $(L_{D,\r})_{\r\geq 0}$ given by \eqref{Lrho}, 
the parameters  $\t_n$ and $\e_n$,
 and the start level $\r_0 := \e_n$.
 Then 
  there exists a    $\overline K\geq 1$ such that for all sufficiently large $n$ we have
  \begin{equation}\label{rates-cor1-hx}
     P^n\biggl(\biggl\{D\in X^n:  \rds-\rs 
     \leq \overline K 
     \e_n
     \biggr\}\biggr) \geq 1 - \frac 1n\, .
  \end{equation}
  Moreover,  if the separation exponent $\k$ is exact, there exists another constant $\underbar K\geq 1$ 
  such that for all sufficiently large $n$ we have
  \begin{equation}\label{rates-cor1-hxx}
     P^n\biggl(\in X^n:
     \underbar K 
     \e_n
     \leq \rds-\rs \leq 
     \overline K 
     \e_n
     \biggr) \geq 1 - \frac 1n\, .
  \end{equation}
  Finally, if $\k=\infty$ and $\supp K\subset B_{\snorm \cdot}$, then   \eqref{rates-cor1-hxx} holds
  for all sufficiently large $n$, if $\s_n=\d_n$ and
     \begin{displaymath}
      \e_n \sim  \Bigl( \frac {\log n \cdot \log \log n}n  \Bigr)^{\frac 1 2} \, , \,\,
       \d_n \sim  \bigl(  \log \log n\bigr)^{-\frac {1}{2d}} \, , \,\,
        \mbox{and} \,\, 
       \t_n \sim \bigl(  \log \log n\bigr)^{-\frac {\g}{3d}}\, .
   \end{displaymath}
\end{corollary}

Note that the rates obtained in Corollary \ref{rates-cor1} only differ by the factor $(\log n)^2$
from the rates in \cite[Corollary 4.4]{Steinwart15a}. Moreover, if $K$ has a bounded support, then
an easy modification of the above corollary yields exactly the same rates as in \cite[Corollary 4.4]{Steinwart15a}.

Our next goal is to establish rates for $\mu(B_i(D)\symdif A_i^*)\to 0$. Since this is a modified level set estimation problem, let us recall  some 
assumptions on $P$, which have been used in this context. 
 The first assumption in this direction is one-sided variant of a well-known condition
 introduced by  Polonik
\cite{Polonik95a}.

\begin{definition}
   Let $\mu$ be a finite measure on $X$ and $P$ be a distribution on $X$ that has a $\mu$-density $h$. 
   For a given level $\r\geq 0$, we say that 
    $P$ has flatness exponent  $\vt\in (0,\infty]$, if there exists a constant $\cflat>0$ such that 
    \begin{equation}\label{polon}
       \mu\bigl(\{ 0< h-\r <s    \}  \bigr) \leq (\cflat s)^\vt\qspace s>0.
    \end{equation}
\end{definition}

The larger $\vt$ is, 
 the steeper $h$ must approach  $\r$ from above. In particular, for $\vt=\infty$, the 
density $h$ is allowed to take the value $\r$ but is otherwise 
bounded away from   $\r$. 

Next, we describe 
the roughness of the boundary of the clusters.

\begin{definition}
  Let \assx M be satisfied. Given some $\a\in (0,1]$, 
the clusters have 
an $\a$-smooth boundary, if
there exists a constant $\cbound>0$ such that, for all $\r\in (\rs,\rss]$, $\d\in (0,\dthick]$, and $i=1,2$, we have  
\begin{equation}\label{box-dim}
  \mu\bigl( (A^i_\r)\pde  \setminus  ( A^i_\r)\mde  \bigr) \leq \cbound \d^\a\, ,
\end{equation}
where $A^1_\r$ and $A^2_\r$ denote the two connected components of the level set $M_\r$.
\end{definition}

In $\R^d$, considering $\a>1$ does not make sense, and for an $A\subset \Rd$
with rectifiable boundary  we always have $\a=1$, see \cite[Lemma A.10.4]{SteinwartXXb1}.

\begin{assumption}{R}
 Assumption M is satisfied and $P$ has a bounded Lebesgue density $h$. Moreover, 
 $P$ has flatness exponent $\vt\in (0,\infty]$ at level $\rs$, 
 its clusters have an $\a$-smooth boundary for some $\a\in (0,1]$,
 and its clusters  have separation exponent $\k\in (0,\infty]$.
\end{assumption}

\begin{corollary}\label{rates-cor2}
   Let \assx R be  satisfied 
	and $K$ be as in Corollary \ref{rates-cor1}.
and write $\vr := \min\{\a,  \vt \g \k \}$. 
   Furthermore, let 
    $(\e_n)$, $(\d_n)$, and $(\t_n)$ 
    be sequences with 
     \begin{align*}
      \e_n &\sim  \Bigl( \frac {\log n}{n}  \Bigr)^{\frac {\vr}{2\vr +  \vt d}} (\log\log n)^{-\frac{\vt d}{8\vr+4\vt d}}  \, , &
       \d_n &\sim  \Bigl( \frac {\log n \cdot \log\log n}n  \Bigr)^{\frac {\vt}{2\vr + \vt d}} \, ,\\  
				\s_n &\sim  \Bigl( \frac {(\log n)^3 \cdot \log\log n}n  \Bigr)^{\frac {\vt}{2\vr + \vt d}} \, , &
       \t_n &\sim  \Bigl( \frac {(\log n)^3 \cdot (\log\log n)^2}n  \Bigr)^{\frac {\vt\g}{2\vr + \vt d}}   .
   \end{align*}  
   Assume that, for $n\geq 1$, we feed Algorithm \ref{cluster-algo-generic} as 
in Corollary \ref{rates-cor1}.  Then 
  there is a  constant  $\overline K\geq 1$ such that, for all    $n\geq 1$ and the ordering as in  \eqref{cluster-chunk-kde-bound}, 
  we have
  \begin{displaymath}
     P^n\biggl(\!D: \sum_{i=1}^2\mu\bigl(B_i(D) \!\symdif\! A_i^*\bigr)  
     \leq K \Bigl( \frac {(\log n)^3 \cdot (\log\log n)^2}n  \Bigr)^{\frac {\vt\vr}{2\vr + \vt d}}
    \biggr) \geq  1-\frac 1 n \, .
  \end{displaymath}
\end{corollary}

Again, the rates obtained in Corollary \ref{rates-cor2} only differ by the factor $(\log n)^2$
from the rates in \cite[Corollary 4.8]{Steinwart15a}. Moreover, if $K$ has a bounded support, then
an easy modification of Corollary \ref{rates-cor2} again yields exactly the same rates as in 
\cite[Corollary 4.8]{Steinwart15a}.

Our final goal is to modify the adaptive parameter selection strategy for the histogram-based 
clustering algorithm of \cite{Steinwart15a}
to our KDE-based clustering algorithm. To this end, 
let  $\D\subset (0,1]$ be   finite    and  $n\geq 1$, $\vs\geq 1$.
For $\d\in \D$, we fix   $\s_{\d,n}>0$ and $\t_{\d,n}>0$
such that \eqref{def-sigma} and $\t_{\d,n}\geq 2\psi(2\s_{\d,n})$ are satisfied. 
In addition, we define 
\begin{align}\label{eps-adap}
 \e_{\d,n}
:= C_u\sqrt{\frac{ |\log \d| ( \vs + \log |\D|)\log\log n}{\d^{d}n}}
+ \max\bigl\{1, 2 d^2 \vold\} \cdot c   \cdot \d^{|\log \d|-d}
\end{align}
where $C_u\geq 1$ is some  user-specified constant
and the second term can actually be omitted if the used kernel $K$ has bounded support.
Now assume that, for each $\d\in \D$,
we run Algorithm \ref{cluster-algo-generic} with the parameters  $\e_{\d,n}$ and $\t_{\d,n}$, with
the start level $\r_0 := \e_{\d,n}$, and  
 with the level set estimators $(L_{D,\r})_{\r\geq 0}$ given by \eqref{Lrho}.
Let us consider 
a width $\d_{D,\D}^*\in \D$ that achieves the smallest returned level, i.e.
\begin{equation}\label{def-rdds}
 \d_{D,\D}^* \in \arg\min_{\d\in \D} \routDd\,.
\end{equation}
Note that in general, this width may  not be uniquely determined, so that in the following
we need to additionally assume that we have a well-defined choice, e.g.~the smallest $\d\in \D$
satisfying \eqref{def-rdds}.
Moreover,  we write 
\begin{equation}\label{def-rsdd}
 \r^*_{D,\D} :=   \min_{\d\in \D} \routDd
\end{equation}
for the  smallest returned level.  
Note that unlike the width $\d_{D,\D}^*$, the level $\r^*_{D,\D}$ is always unique.
Finally, we define $\e_{D,\D}:=\e_{\d^*_{D,\D},n}$ and $\t_{D,\D}:=\t_{\d^*_{D,\D},n}$.
With these preparation we can now present the following finite sample bound for 
$\r^*_{D,\D}$.

\begin{theorem}\label{adaptive-level}
Let $P$ be a distribution for which \assx M is satisfied and 
whose Lebesgue density is bounded.
Moreover, consider a symmetric kernel $K:\R^d\to [0,\infty)$
with exponential tail behavior,
for which the assumptions of Theorem \ref{approximate-thm} hold.
In addition, assume that 
 the two clusters of $P$ have separation exponent $\k\in (0,\infty]$. 
 For a fixed finite  $\D\subset (0,\eul^{-1}]$, and $n\geq 1$,  $\vs\geq 1$, and $C_u\geq 1$, we define 
 $\e_{\d,n}$ by \eqref{eps-adap} and 
$\s_{\d,n}>0$ and $\t_{\d,n}>0$
such that \eqref{def-sigma}, $\t_{\d,n}\geq 2\psi(2\s_{\d,n})$, 
and $2\s_{\d,n}\leq \dthick$ are satisfied for all $\d\in \D$.
 Furthermore, assume that 
   $4 C_u^2\log\log n\geq C\inorm h$, where $C$ is the constant in \eqref{simplified-conc}
 and  
    $\e_{\d,n} + (\t_{\d,n}/ \csepl)^\k \leq (\rss-\rs)/9$
  for all $\d\in \D$.  Then we have
 \begin{displaymath}
   P^n\Bigl(\Bigl\{D\in X^n: \e_{D,\D}\,<\, \r^*_{D,\D}- \rs \,\leq\,  \min_{\d\in \D}  \bigl((\t_{\d,n}/ \csepl)^\k  + 6\e_{\d,n} \bigr)
    \Bigr\}\Bigr) \geq 1-e^{-\vs}\, .
 \end{displaymath} 
 Moreover, if the separation exponent $\k$ is exact and $\k<\infty$, then the assumptions above 
 actually guarantee
  \begin{displaymath}
   P^n\Bigl(D:  \min_{\d\in \D} \bigl(c_1\t_{\d,n}^\k + \e_{\d,n} \bigr)\,< \,\r^*_{D,\D}- \rs \,\leq\, \min_{\d\in \D} \bigl(c_2\t_{\d,n}^\k + 6\e_{\d,n} \bigr)
    \Bigr) \geq 1-e^{-\vs}\, ,
 \end{displaymath} 
 where $c_1 := \frac 14( 6\csepu)^{-\k}$ and $c_2 := \csepl^{-\k}$, 
 and similarly
   \begin{displaymath}
   P^n\Bigl(\Bigl\{D\in X^n:  c_1 \t_{D,\D}^\k     +  \e_{D,\D} \,<\, \r^*_{D,\D}- \rs \,\leq\,  c_2 \t_{D,\D}^\k  + 6\e_{D,\D}
    \Bigr\}\Bigr) \geq 1-e^{-\vs}\, .
 \end{displaymath} 
\end{theorem}

To achieve our goal of an adaptive parameter selection strategy, it now suffices in 
view of  Theorem \ref{adaptive-level} to define appropriate $\D$, $\s_{\d,n}$, and $\t_{\d,n}$.
Here we proceed as in \cite[Section 5]{Steinwart15a}. Namely, for 
 $n\geq 16$, we consider the interval
\begin{align}\label{candidate-deltas}
   I_n:= \biggl[  \Bigl( \frac {\log n \cdot (\log\log n)^2}n  \Bigr)^{\frac {1}{d}},\, \Bigl(\frac 1 {\log\log n}\Bigr)^{\frac 1d}    \biggr]
\end{align}
and fix some $n^{-1/d}$-net $\D_n\subset I_n$ of $I_n$ with $|\D_n|\leq n$.
Furthermore, for some fixed $C_u\geq 1$ and $n\geq 16$, we 
define $\s_{\d,n}$ by \eqref{def-sigma}, write 
$\t_{\d,n}:= \s_{\d,n}^\g\log\log\log n$, and define
$\e_{\d,n}$ by \eqref{eps-adap}
for all $\d\in \D_n$ and $\vs = \log n$. Following the ideas of the proofs of \cite[Corollaries 5.2 and 5.3]{Steinwart15a}
we then obtain a constant $\overline K$ such that 
for all sufficiently large $n\geq 16$ we have
  \begin{equation}\label{adaptive-level-cor-hx}
     P^n\biggl(D:  \r^*_{D,\D_n}-\rs 
     \leq \overline K \Bigl( \frac {(\log n)^3 \cdot  (\log\log n)^2}n  \Bigr)^{\frac {\g \k}{2\g \k +  d}}
     \biggr) \geq 1 - \frac 1n\, .
  \end{equation}
Here, \eqref{adaptive-level-cor-hx} holds if $P$ has separation exponent $\k\in (0,\infty)$, and if the kernel 
$K$ has bounded support, it remains true for $\k=\infty$. In addition, the upper bound in \eqref{adaptive-level-cor-hx}
can be matched by a lower bound that  only differs by a double logarithmic factor provided that 
the separation exponent $\k\in (0,\infty)$ is exact.
Finally, if Assumption R is satisfied, we further find 
  \begin{displaymath}
     P^n\biggl(\!D:  \sum_{i=1}^2\mu(B_i(D) \!\symdif\! A_i^*)  
     \leq \hat K \Bigl( \frac {(\log n)^3 \cdot (\log\log n)^2}n  \Bigr)^{\frac {\vt\g\k}{2\g\k + \vt d}}
    \biggr) \geq 1-\frac 1n \, ,
  \end{displaymath}
for all sufficiently large $n\geq 16$, where $\hat K$ is another constant independent of $n$.

Finally, we like to mention that the presented adaptive strategy works optimally as long as all split levels 
have the same exact separation exponent $\k$. Indeed, as described above, our strategy detects, modulo some logarithmic terms, an asymptotically optimal choice 
 $\d_{D,\Delta}^*$ for the first split level, and since for all other split levels this choice is also asymptotically optimal as long as they have the same exact 
separation exponent $\k$, we see that  Algorithm \ref{split-tree-algo-generic} is adaptive when using $\d_{D,\Delta}^*$.


\section{Comparisons}\label{sec:comparison}

In this section we compare our findings to the most closely related papers, namely 
\cite{WaLuRi19a} and \cite{ChDaKpLu14a}. In particular, we discuss the different assumptions as well as 
the obtained statistical guarantees.

 
To begin with, let us have a look at the assumptions on $P$, respectively its density $h$ made in 
\cite{WaLuRi19a}. Here, we first note that $h$ is assumed to be $\a$-smooth for some $\a>0$, that is,
$h$ is $s:=\lfloor \a\rfloor$-times continuously differentiable and all partial derivatives of order $s$ are 
$(\a-s)$-H\"older continuous. In general, $h$ does not need to have compact support, 
instead, \cite{WaLuRi19a} only assumes that $\mhx \r$ is compact for all $\r>\rs$, where $\rs$ is the smallest split level,
see their Assumptions \textbf{C} and \textbf{C'}, respectively. 
In this respect we note that for $P$ having a continuous density their notion of split levels is closely related to ours and that 
their Assumption \textbf{S}$\mathbf(\k)$
equals our separation exponent $\k$.
Finally, \cite{WaLuRi19a} does not impose a thickness assumption, instead
a so-called \emph{inner cone condition} is considered. 
Recall that this cone condition 
assumes that constants $\e_I>$, $c_I>$, and $r_I>$ exist such that for all split levels $\rs$ and all $\r\in (\rs-\e_I,\rs+\e_I)$
we have 
\begin{align*}
 \lbd \bigl(B(x,r) \cap \mhx\r\bigr) \geq c_I r^d\, , \qquad \qquad x\in \mhx \r, r\in (0,r_I]\, .
\end{align*}
In general, it seems unclear how this condition relates to our thickness assumption, but in many 
natural situations
the inner cone condition and the thickness assumption with $\g=1$
are simultaneously satisfied. For details, we refer to the discussion in \cite[Appendix A.5]{SteinwartXXb1}, 
the rather generic examples considered in \cite[Appendix B.2]{SteinwartXXb2}, and 
the discussion on \cite[page 15]{WaLuRi19a}.
In the following comparison we therefore assume that the inner cone condition and the thickness assumption 
with $\g=1$ are simultaneously satisfied. In addition, we assume that $h$ has finitely many split levels.

Like our results, the clustering algorithm of \cite{WaLuRi19a} is also based on 
a kernel density estimator. However, their central Algorithm 2 is using   so-called $\a$-valid kernels, 
whose KDEs enjoy the ideal approximation error behavior $\inorm{h_{P,\d} - h}\preceq \d^\a$ for $\d\to 0$ and 
$\a$-smooth densities $h$. We refer to \cite[page 9]{WaLuRi19a} for details. Finally, 
their split level estimator uses a verification strategy that is similar to Line 3 of our
Algorithm \ref{cluster-algo-generic}, see 
\cite[Definition 6]{WaLuRi19a}.

For a comparison of the split level guarantee provided   in \cite[Proposition 3]{WaLuRi19a}
and our guarantees, we now assume that  $h$ is $\a$-smooth, that all split levels 
have  separation exponent $\a$, and that $h$ 
satisfies the additional assumptions discussed
above. In addition, we assume that their Algorithm 2  uses an $\a$-valid kernel. 
Then \cite[Proposition 3]{WaLuRi19a} shows that, modulo some logarithmic factors,   all split levels can be estimated with rate 
\begin{align*}
 n^{-\frac{\a}{2\a+d}}\, .
 \end{align*}
 Since we can choose $\a=\k$ and $\g=1$, these rate coincide with ours established in Corollary \ref{rates-cor1} if 
 we again ignore logarithmic factors. Given any $\k>0$, however, our results do not 
 require $h$ to be $\k$-smooth, while \cite[Proposition 3]{WaLuRi19a} only holds for $\k$-smooth densities.
Moreover note that there are $\a$-smooth densities with $\k\gg\a$, in fact we may have $\a=1$ and $\k= \infty$, and for such densities,
 our rates are significantly better than those of \cite{WaLuRi19a}.
 In addition, our clustering algorithm does not need to use kernels that are aligned with the smoothness of $h$, 
 and finally, 
 our algorithm can be made adaptive to the unknown $\k$.
 Finally, we like to emphasize that apart from Proposition 3, \cite{WaLuRi19a}
 contains several other interesting results, which are, however, not comparable to ours. 
 

\vspace*{1em}

Unlike the comparison to \cite{WaLuRi19a}, the comparison to \cite{ChDaKpLu14a} turns out to be much more 
involved, as the latter paper uses assumptions that are quire different to ours. Providing a detailed comparison is 
therefore beyond the scope of this paper and we refer the interested reader to 
Appendix \ref{sec:supplement}
for a detailed comparison. 
To summarize the key points of this comparison, we show that, up to logarithmic factors,  the best possible convergence rate achieved by
the central Theorem VII.5 of \cite{ChDaKpLu14a} 
 for estimating the split levels is $n^{-\frac{\theta}{3\theta+d}}$ while our algorithm achieves a rate of $n^{-\frac{\theta}{2\theta+d}}$, which is strictly better for all dimensions, with $\theta$ being the H\"{o}lder-continuity of the underlying density. In particular, our rates can be achieved without knowing $\theta$, while \cite{ChDaKpLu14a} do not offer such adaptivity.
In addition, our results can handle discontinuous densities, e.g.~step functions on rectangles
with mutually positive distances, while their Theorem VII.5 does not provide any guarantee at all for such densities.
In particular, the consistency results for their unpruned algorithm require ``mild uniform continuity conditions'', see the end
of Section III.B in  \cite{ChDaKpLu14a}, and the guarantees for the pruned algorithm stated in their Theorem VII.5 explicitly require control
on the uniform modulus of continuity.
Finally, apart from split level estimation, our results also provide guarantees for corresponding clusters in measure while no such guarantees are in \cite{ChDaKpLu14a}.

\section{Experiments}\label{sec:experiments}

In this section, we illustrate the behavior of our generic KDE-based clustering
algorithm on a few artificial data sets for which the 
ground truth clustering can be computed.

\begin{figure}
\includegraphics[width=0.21\textwidth]{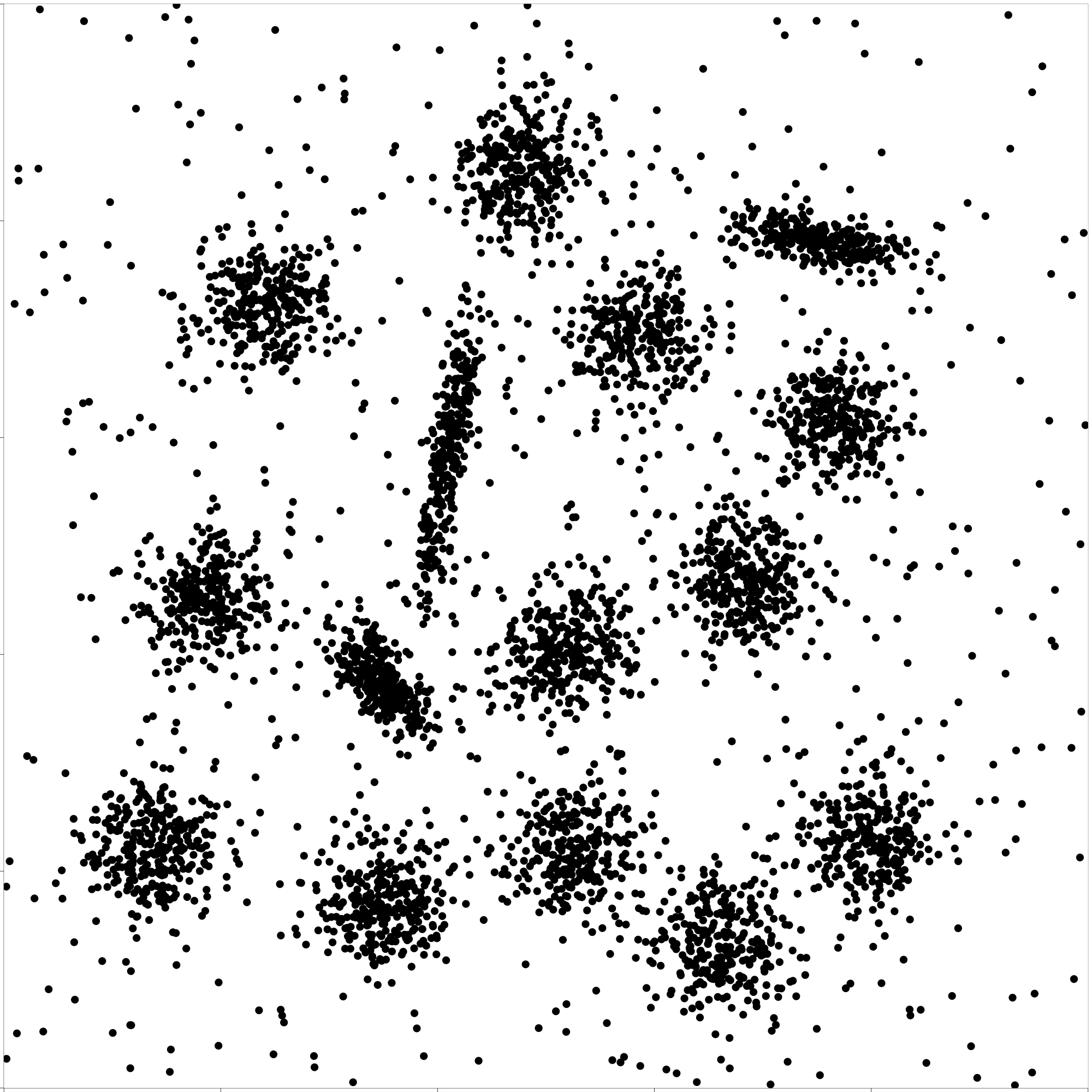}
\hspace*{0.03\textwidth}
\includegraphics[width=0.21\textwidth]{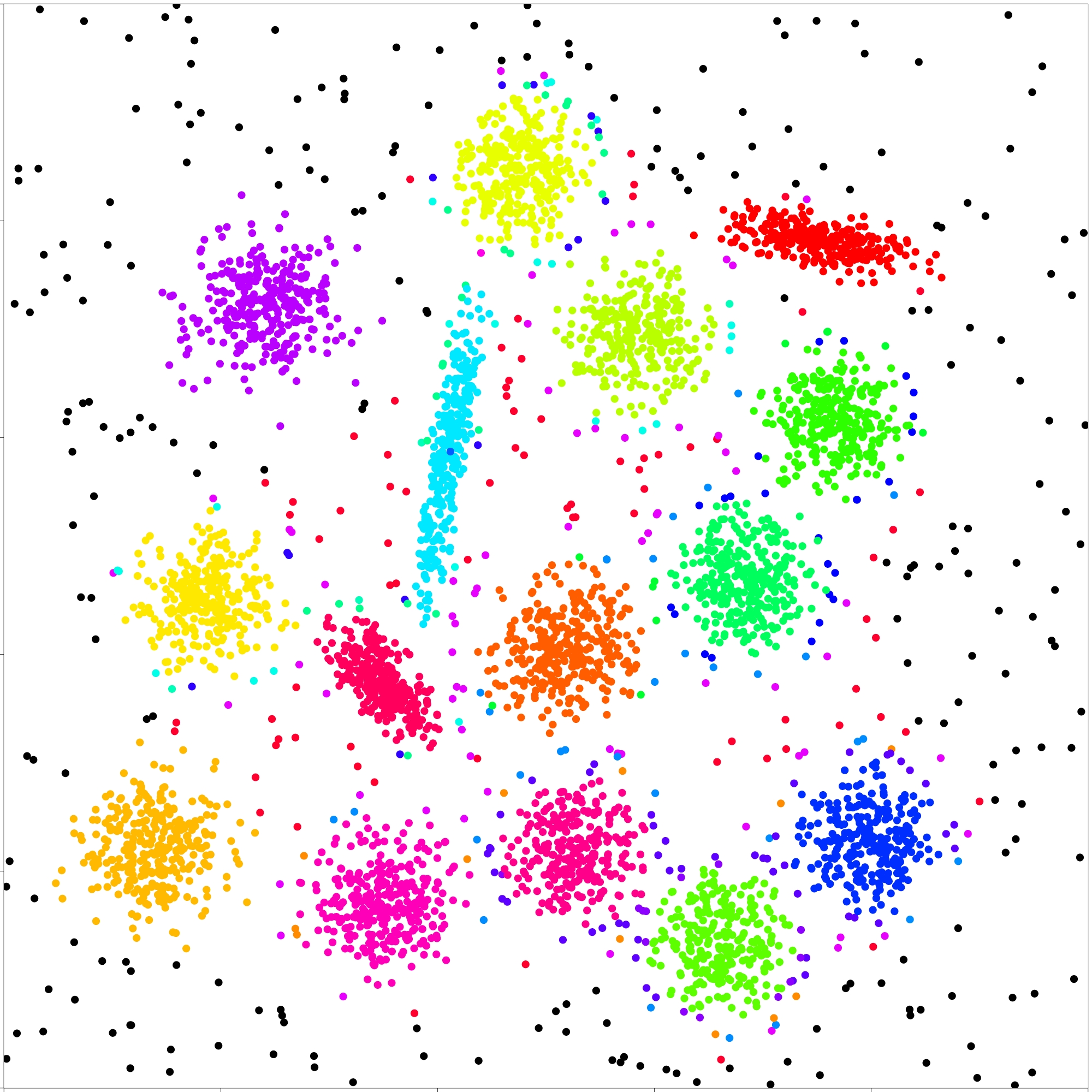}
\hspace*{0.03\textwidth}
\includegraphics[width=0.21\textwidth]{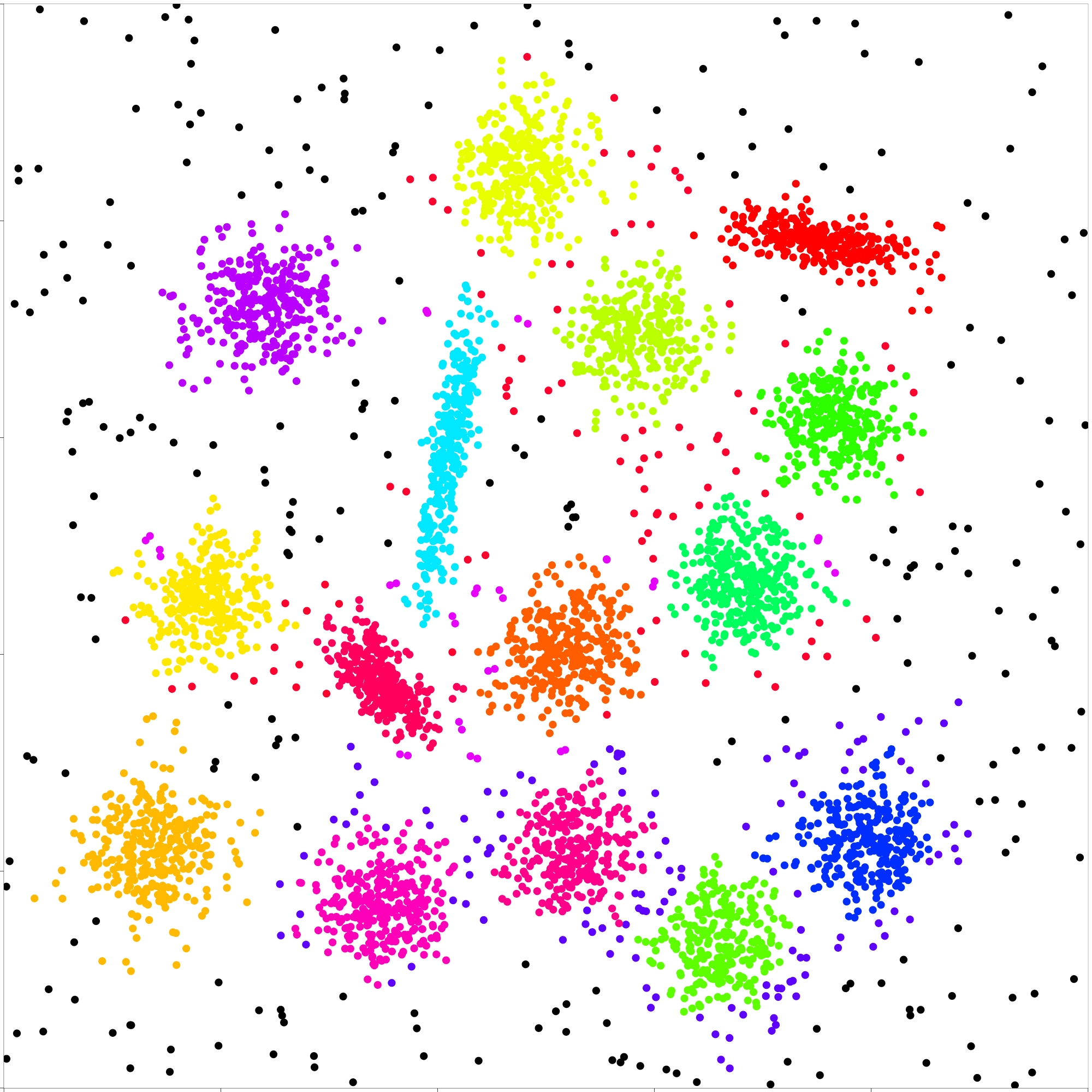}
\hspace*{0.03\textwidth}
\includegraphics[width=0.21\textwidth]{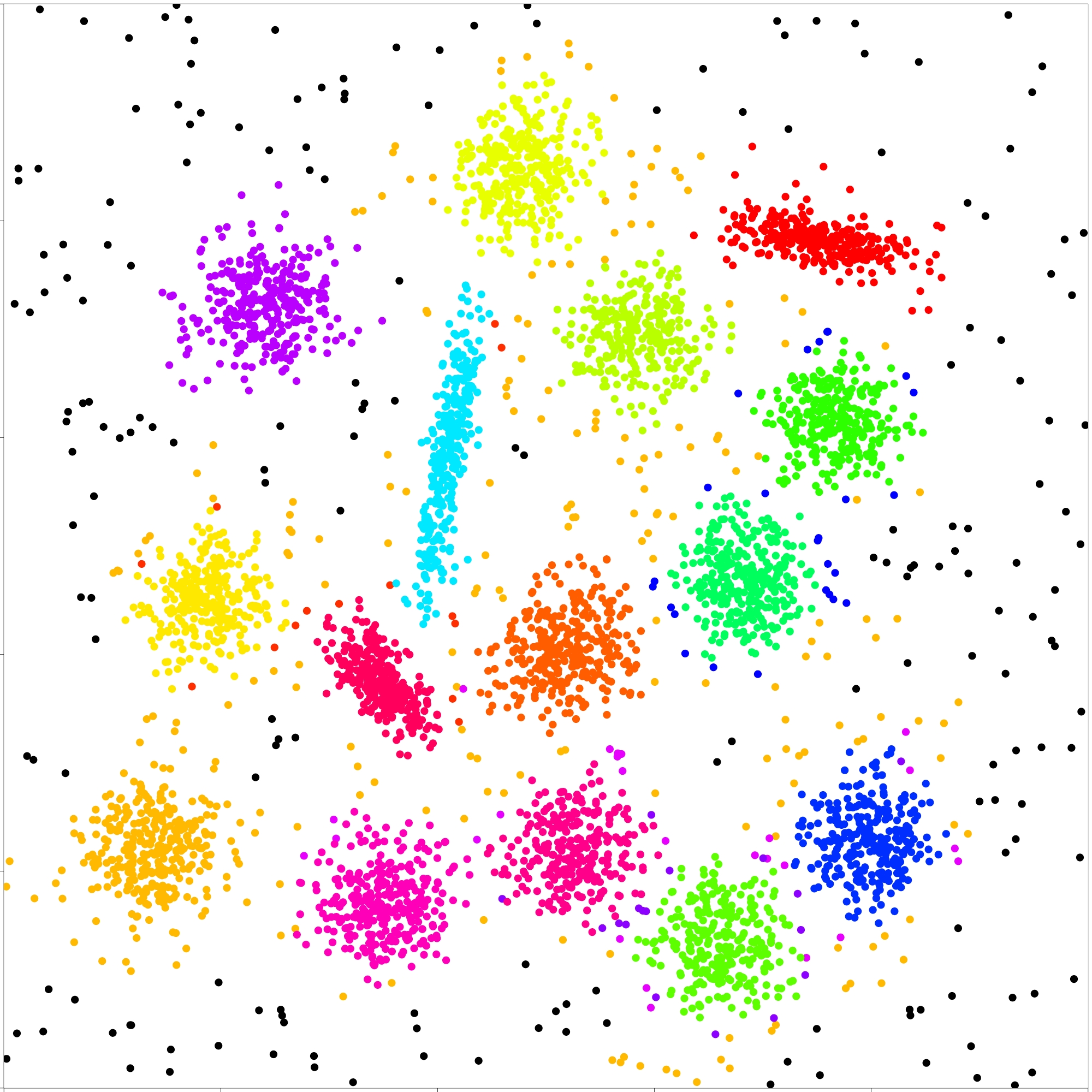}
\caption{\textbf{Left.} 5000 samples from a small Gaussian background noise and  15   well separated Gaussians with 12 of them having the same covariance. 
\textbf{Middle-Left.} Ground truth clustering on the data set obtained with the help of the density.
Notice the magenta, sparsely spread background samples in the middle of the picture that belong 
to one of the two clusters emerging at the lowest split level, but that do not belong to any
other cluster further up in this branch. The black points do not belong to any cluster.
\textbf{Middle-Right and Right.} Clusters obtained from the data with the moving window and Epanechnikov kernel, respectively. All clusters and their shapes
are correctly identified, but 
the hierarchical structure of the cluster tree is not correctly estimated as, e.g., the
black points in the middle of the image for the moving window kernel, whose ground truth color is magenta, illustrate.
}
\label{figure:s2}
\end{figure}

\begin{figure}
\includegraphics[width=0.21\textwidth]{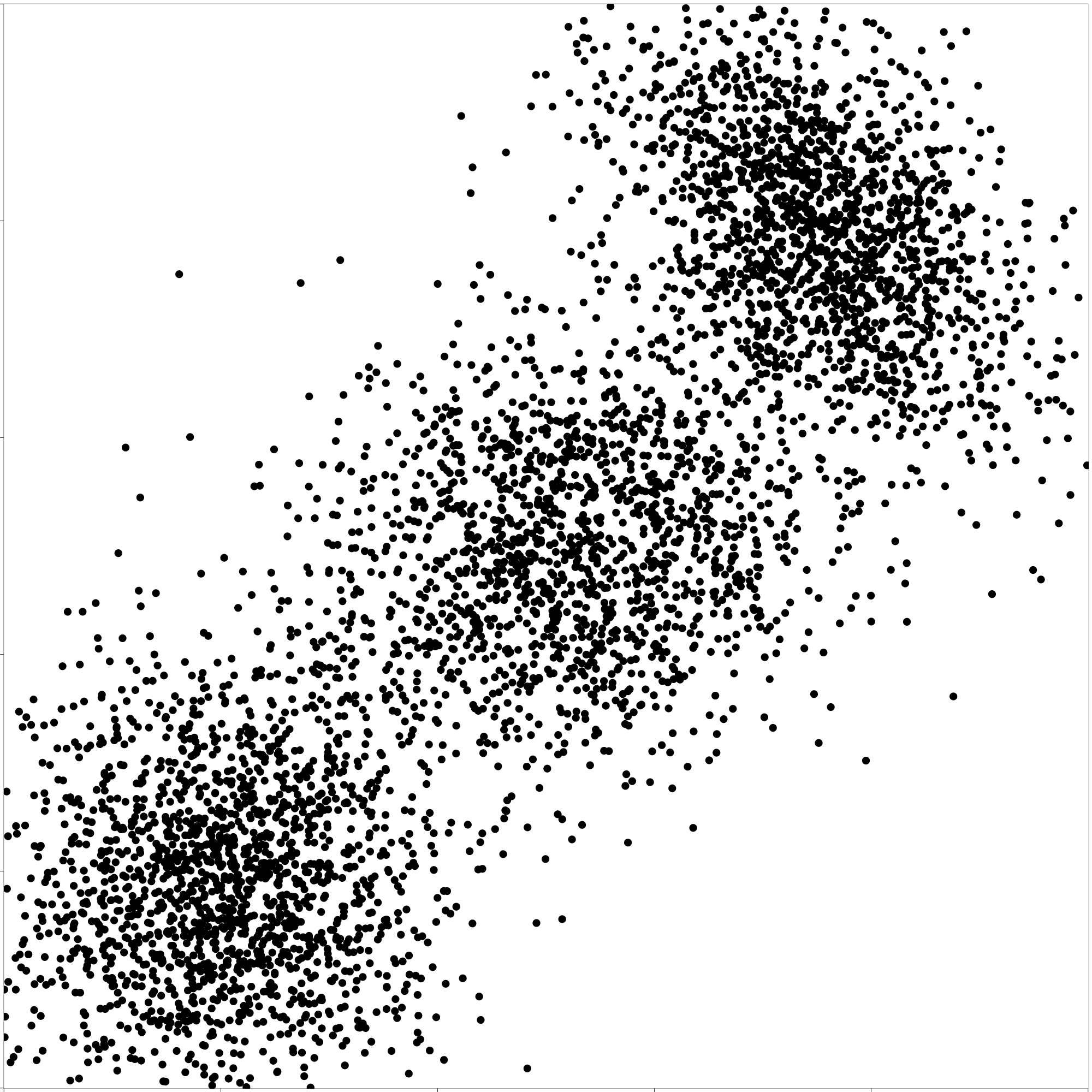}
\hspace*{0.03\textwidth}
\includegraphics[width=0.21\textwidth]{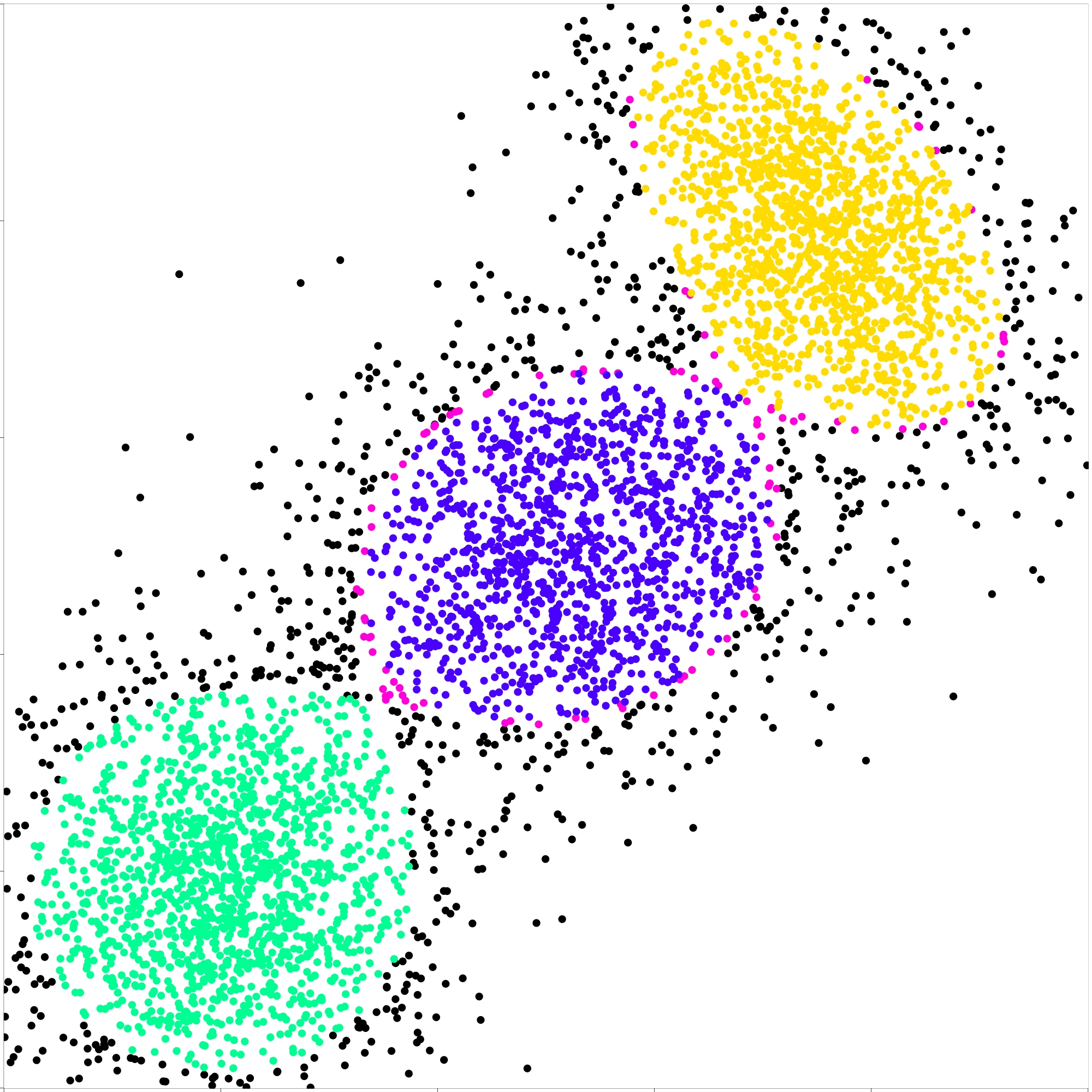}
\hspace*{0.03\textwidth}
\includegraphics[width=0.21\textwidth]{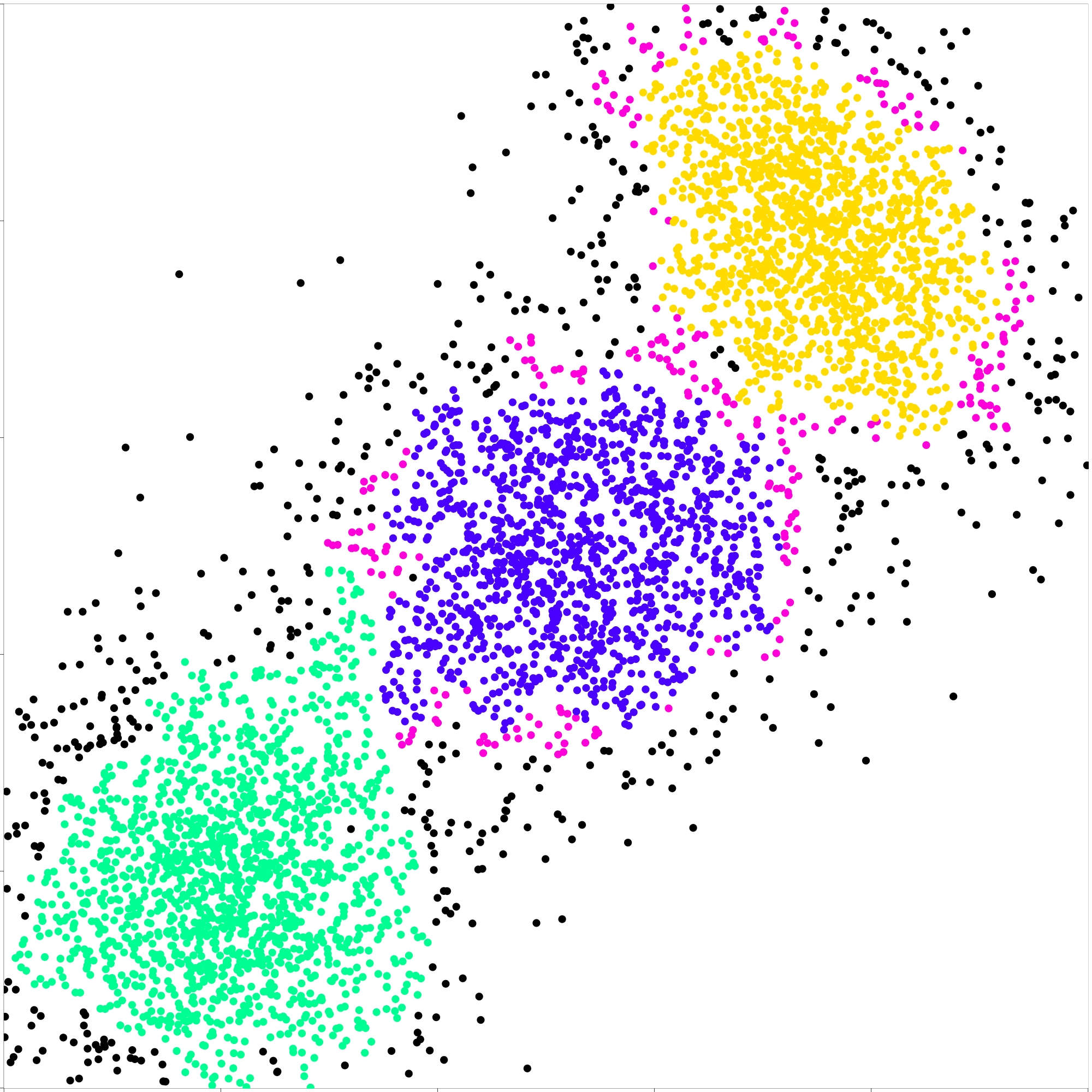}
\hspace*{0.03\textwidth}
\includegraphics[width=0.21\textwidth]{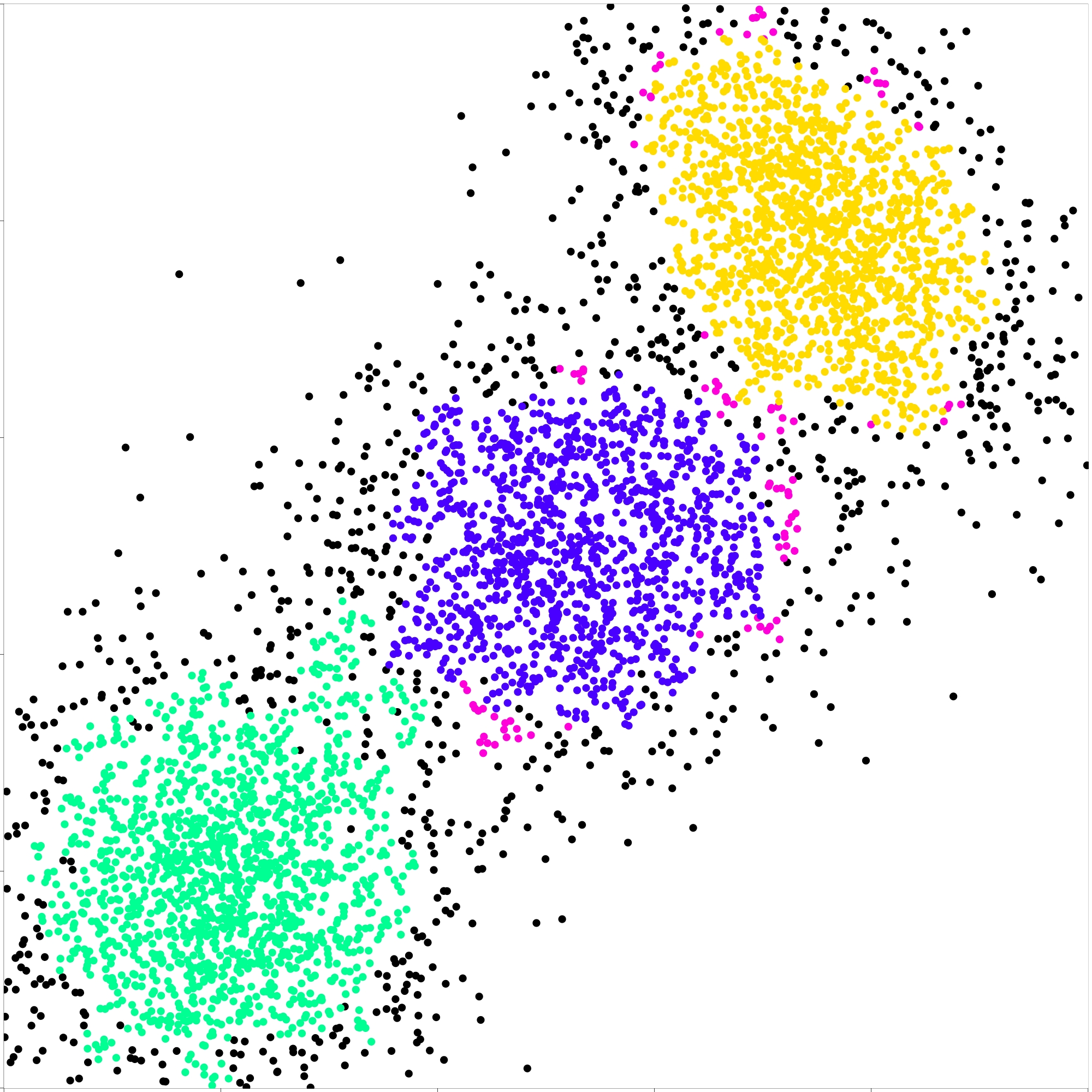}
\caption{\textbf{Left.} 5000 samples from 3 close-by Gaussians with similar covariances.
\textbf{Middle-Left.} Ground truth clustering on the data set obtained with the help of the density.
\textbf{Middle-Right and Right.} Clusters obtained from the data with the moving window and Epanechnikov kernel, respectively.
With both kernels, all 3 clusters and the structure of the cluster tree are correctly estimated. Moreover,  the shapes of the clusters are sufficiently well  estimated. 
}
\label{figure:toy-3g}
\end{figure}

\textbf{Data.}
We consider four different two-dimensional data sets in $[0,1]^2$ with 5000 samples and 
different degrees 
of difficulty:
The first data set, depicted in Figure \ref{figure:s2} is generated from a mixture
of 16 Gaussian distributions, where 15 of them are well separated and the 16th distribution,
which shares its mean with one of the other distributions, creates some background noise.
In addition, 12 out of the 15 well-separated distributions have a covariance of the form $\lb I_2$, where $I_2$ is the identity matrix and $\lb$ is some scaling factor, 
while the remaining 3 Gaussians have a different covariance matrix. 
The second data set, which is shown in Figure \ref{figure:toy-3g}, is generated from 
a mixture of 3 Gaussian distribution with similar but not identical covariances.
Unlike  in the first data set, however, the Gaussians 
are less
separated and less concentrated
making this data set slightly more difficult.
The third data set is a variant of the classical two moons, or bananas data set, which
is often
used to assess the clustering performance on non-centroid-like
clusters.
Unlike  its usual form, however, our variant consists of two very ``fuzzy'' 
bananas, which makes even a visual inspection not immediately obvious.  
The fourth data set displayed in Figure \ref{figure:crosses} is another mixture of Gaussians. 
In this data set, however, each of its three  clusters corresponds to
a mixture of two Gaussian that have the same mean but different variances.
In addition, the clusters are  ``merging'' into each other, and as a result, this data set 
can be viewed as the most challenging one.

Besides the random samples, we also generated a $1000 \times 1000$ grid of 
$[0,1]^2$ on which we computed the densities of the 4 data sets. On these 
 density grids, we then computed the ground truth clustering
 and applied it to the 4 data sets. This way we can visually compare 
how well the clustering algorithm estimates the ground truth.

\textbf{Algorithm.}
We implemented an iterative version of Algorithm \ref{split-tree-algo-generic}, in which
for $k=0,1,\dots$ we first computed 
the $\t$-connected components of $L_{\r+k\e}$  that do not vanish
at level $L_{\r+(k+2)\e}$, see also Line 3 of Algorithm \ref{cluster-algo-generic}.
With the help of these connected components we then generated the corresponding cluster tree 
estimate, where we note that we skipped the Line 7 of Algorithm \ref{cluster-algo-generic}
since this line has only been inserted into Algorithm \ref{cluster-algo-generic}
to simplify the form of the statistical guarantees. We then called the resulting  cluster tree 
estimator for different KDE level set estimators, that is, for 
different values of $\d$.
The first split in the cluster tree was then obtained by
choosing the $\d^*$ that resulted in the smallest first split level.
If all split levels have the same separation exponent $\k$, then it asymptotically 
suffices to 
consider this $\d^*$ for the entire cluster tree. In general, however, splits further up in the 
tree may have either a different $\k$ or at least a different constant $\csepl$.
In these cases, choosing a different $\d$ at different splits may be beneficial.
We adopted this insight by considering, after each detected split level $\r$, those $\d$
whose estimated cluster tree has another split level within one of the clusters 
emerging at $\rho$. Among those $\d$ we then again choose the one resulting in the smallest 
next split level.

\begin{figure}
\includegraphics[width=0.21\textwidth]{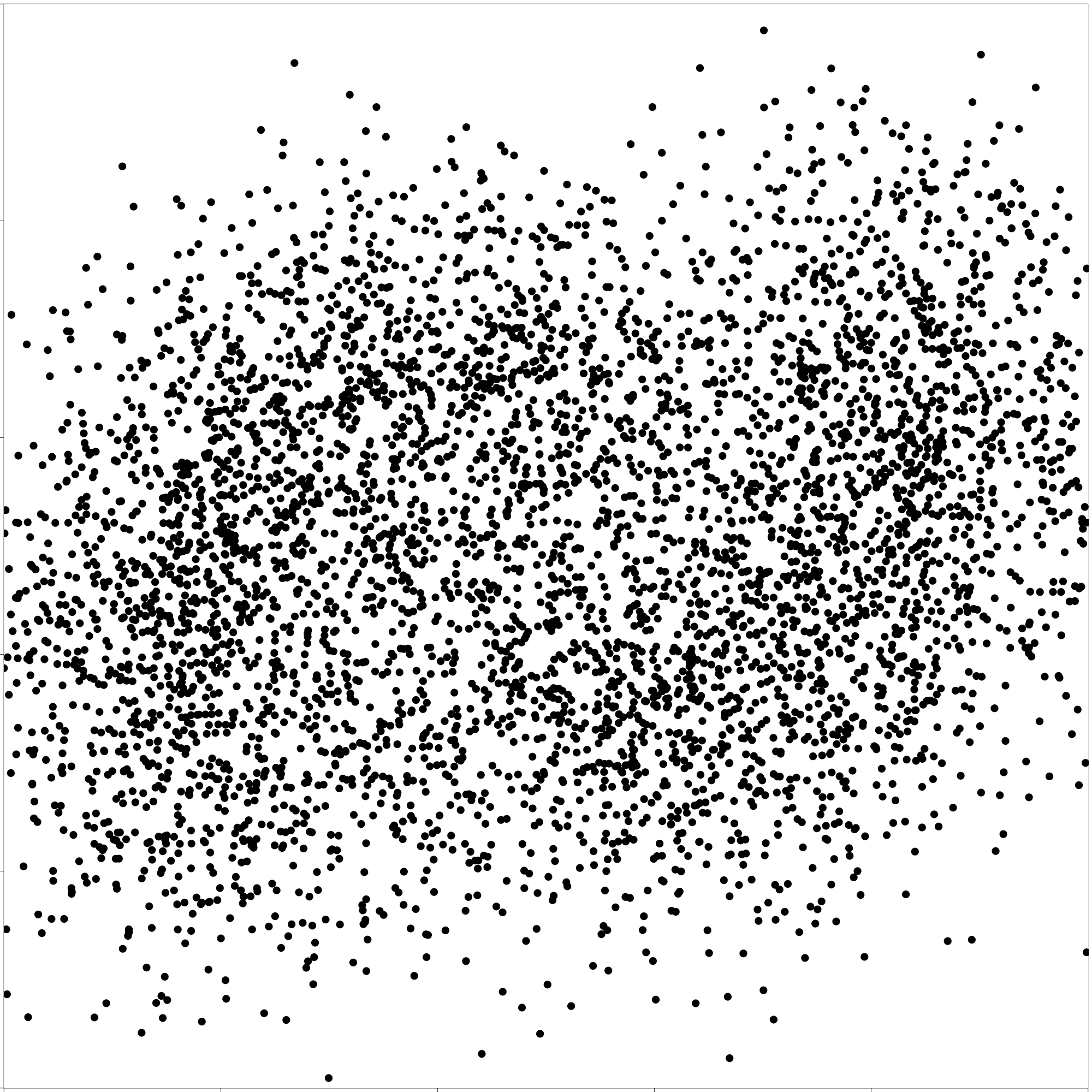}
\hspace*{0.03\textwidth}
\includegraphics[width=0.21\textwidth]{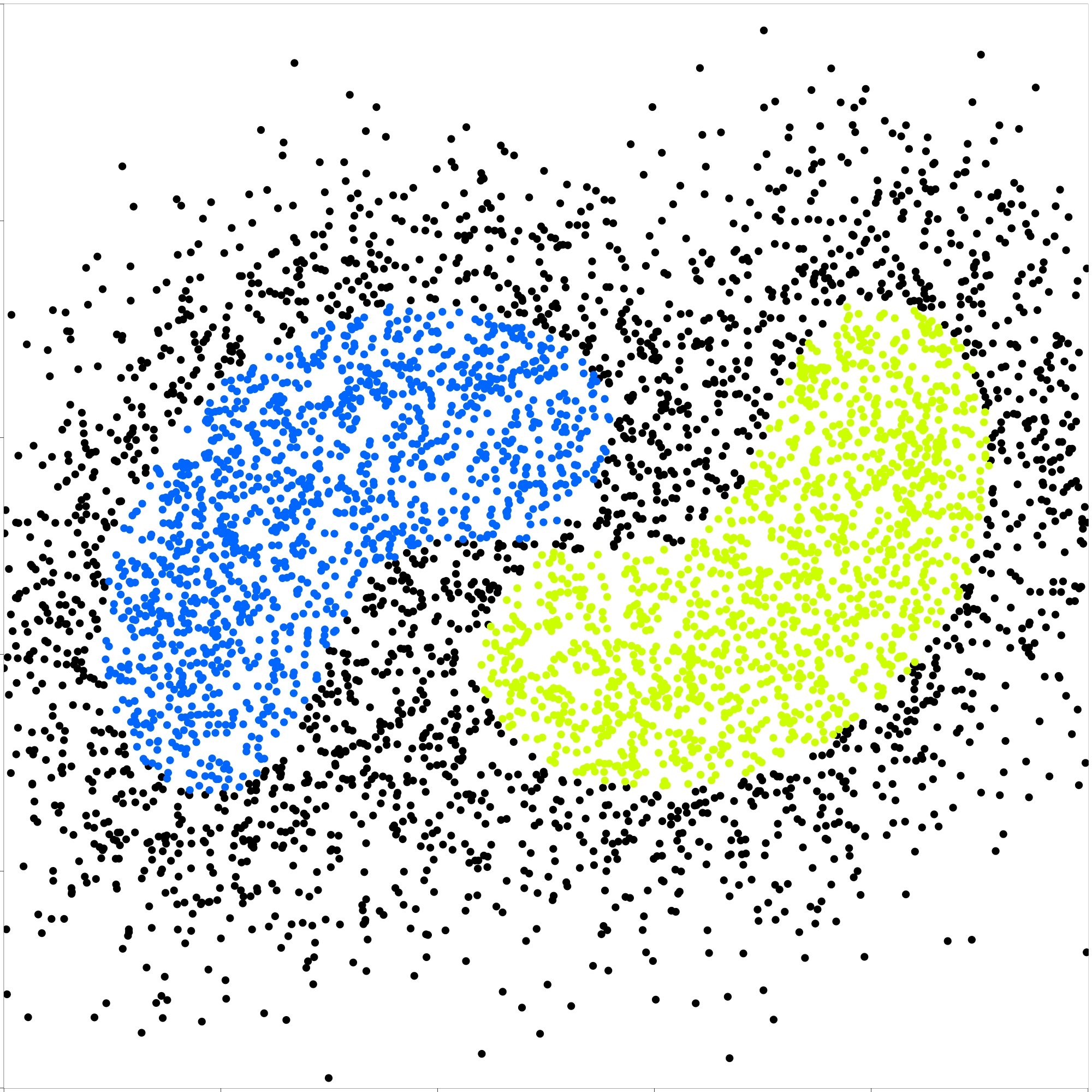}
\hspace*{0.03\textwidth}
\includegraphics[width=0.21\textwidth]{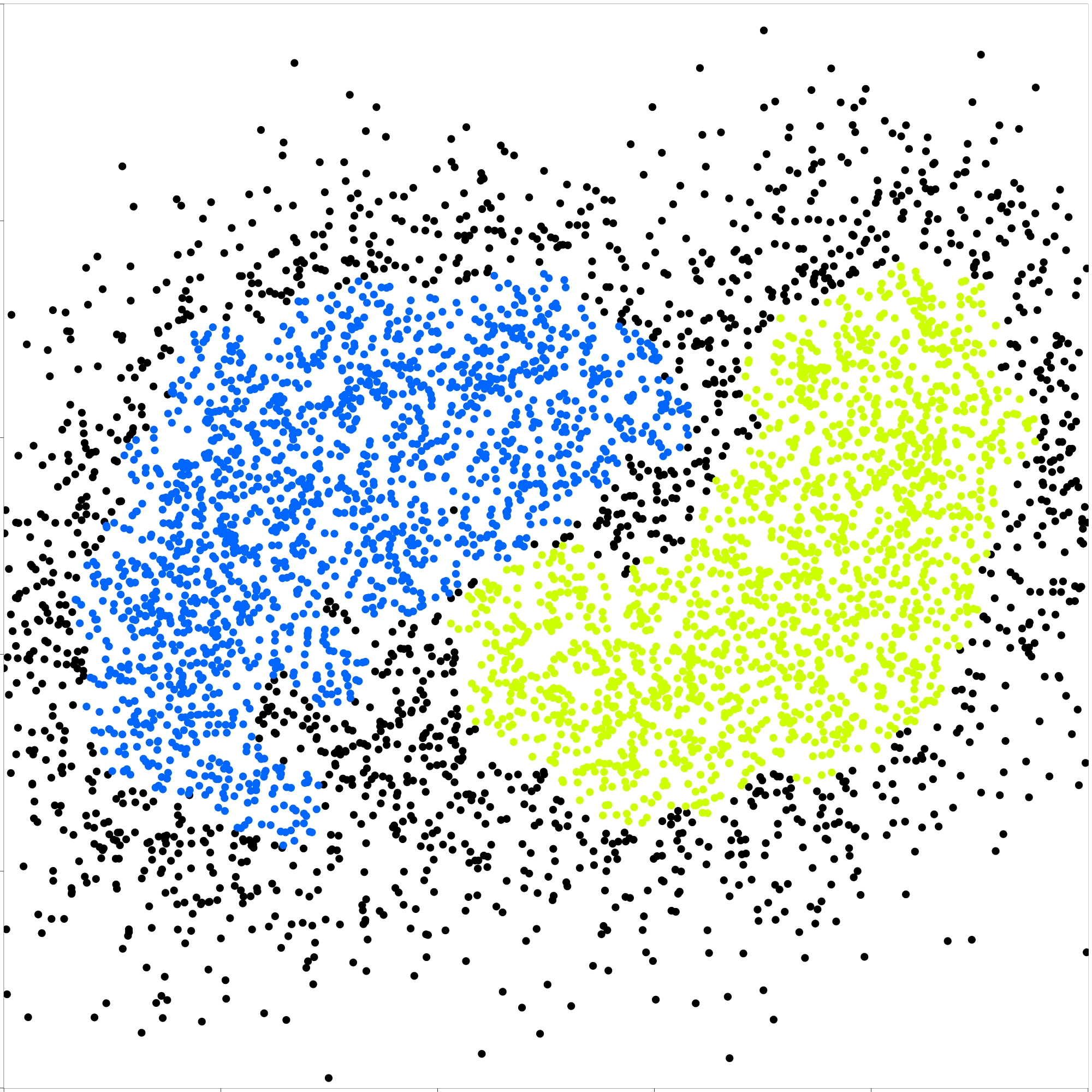}
\hspace*{0.03\textwidth}
\includegraphics[width=0.21\textwidth]{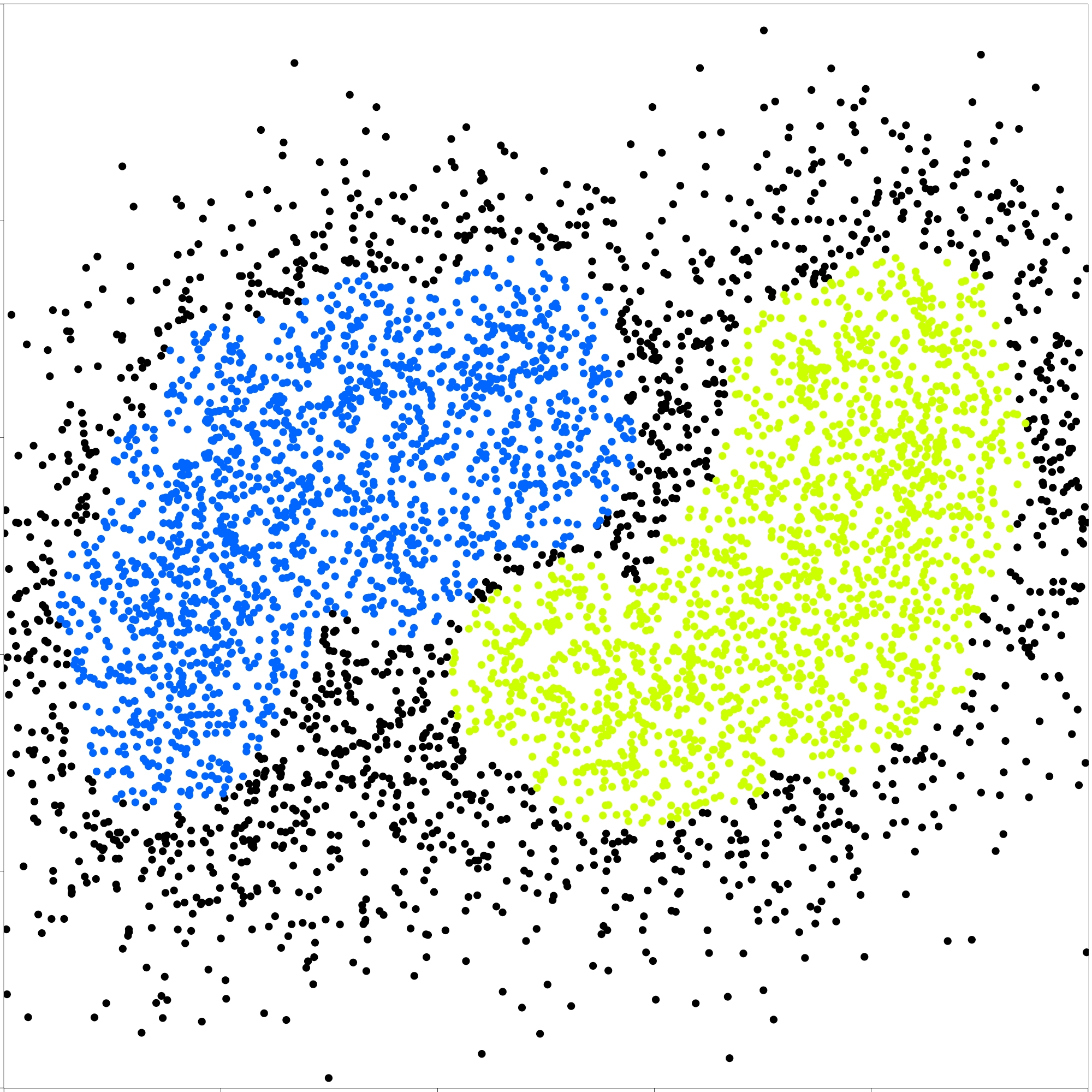}
\caption{\textbf{Left.} 5000 samples from 2 very noisy bananas.
\textbf{Middle-Left.} Ground truth clustering on the data set obtained with the help of the density.
\textbf{Middle-Right and Right.} Clusters obtained from the data with the moving window and Epanechnikov kernel, respectively.
For both kernels, the clusters and the structure of the cluster tree are correctly estimated,
and the shape of the clusters is reasonably well estimated.
}
\label{figure:bananas}
\end{figure}

We considered 500 geometrically spaced 
candidate values of $\d$ between 
$c (\ln(n)/n)^{1/d}$ and $c (\ln n)^{-1/d}$, where in the experiments, the factor $c$ was determined by an estimate of the median mutual distance between the samples of the considered data set.
Notice that modulo this factor $c$ and some (double) logarithmic terms, this setup coincides with the theoretically derived one, see 
\eqref{candidate-deltas}.
Moreover, we considered both a plain moving window kernel and the Epanechnikov kernel,
where in both cases the underlying norm was the Euclidean distance. Since both kernels have bounded support, we simply chose $\s := \d$, see  \eqref{def-sigma}, 
and $\e := 3 \sqrt{\inorm {h_{D,\d}} n^{-1} \d^{-d} }$ for each candidate value $\d$.
Modulo some logarithmic terms, this choice for $\e$ follows our theoretical insights, see  \eqref{eps-adap}. 
Finally, we decided to focus on thickness guarantees with the most natural choice $\g:=1$,
see the detailed discussion in \cite[Appendix A.5]{SteinwartXXb1},
that is, we do not expect the algorithm 
to correctly keep clusters together that have thinner cusps or bridges.
Based on this decision, we  choose
$\t:= (2+\eps)\cdot \d$
with $\eps = 0.00001$, where we 
note that our theoretical findings actually hold true for each value $\t>2\d$, 
if one carefully tracks all constants. In addition, this choice makes it possible that 
the estimated clusters can be as close as $\eps\cdot \d$ to each other.

\textbf{Results.}
The results on the first data set are somewhat mixed. While the algorithms successfully detects
all 15 clusters correctly, it fails to reproduce the cluster tree structure. 
To illustrate this, we note that
the ground truth cluster tree has 14 \emph{different} split levels, that is, even at the 
lowest split level, only two connected components emerge. In contrast, the algorithm using the moving window kernel
identifies 9 different clusters on its lowest split level,
and a similar behavior occurs for the Epanechnikov kernel.
While at first glance, this seems
to be disappointing, we emphasize that also visually it is very difficult, if not impossible,
to infer the correct cluster tree structure from the data, despite the fact that all clusters are
easy to identify. 
On the remaining three data sets, the algorithm  both detects all clusters 
and it 
also 
identifies the cluster tree structure correctly.
In this respect we note that at least the last two data sets
 probably pose a serious challenge to all existing clustering algorithms, in particular 
if they are run in an automated way.

\begin{figure}
\includegraphics[width=0.21\textwidth]{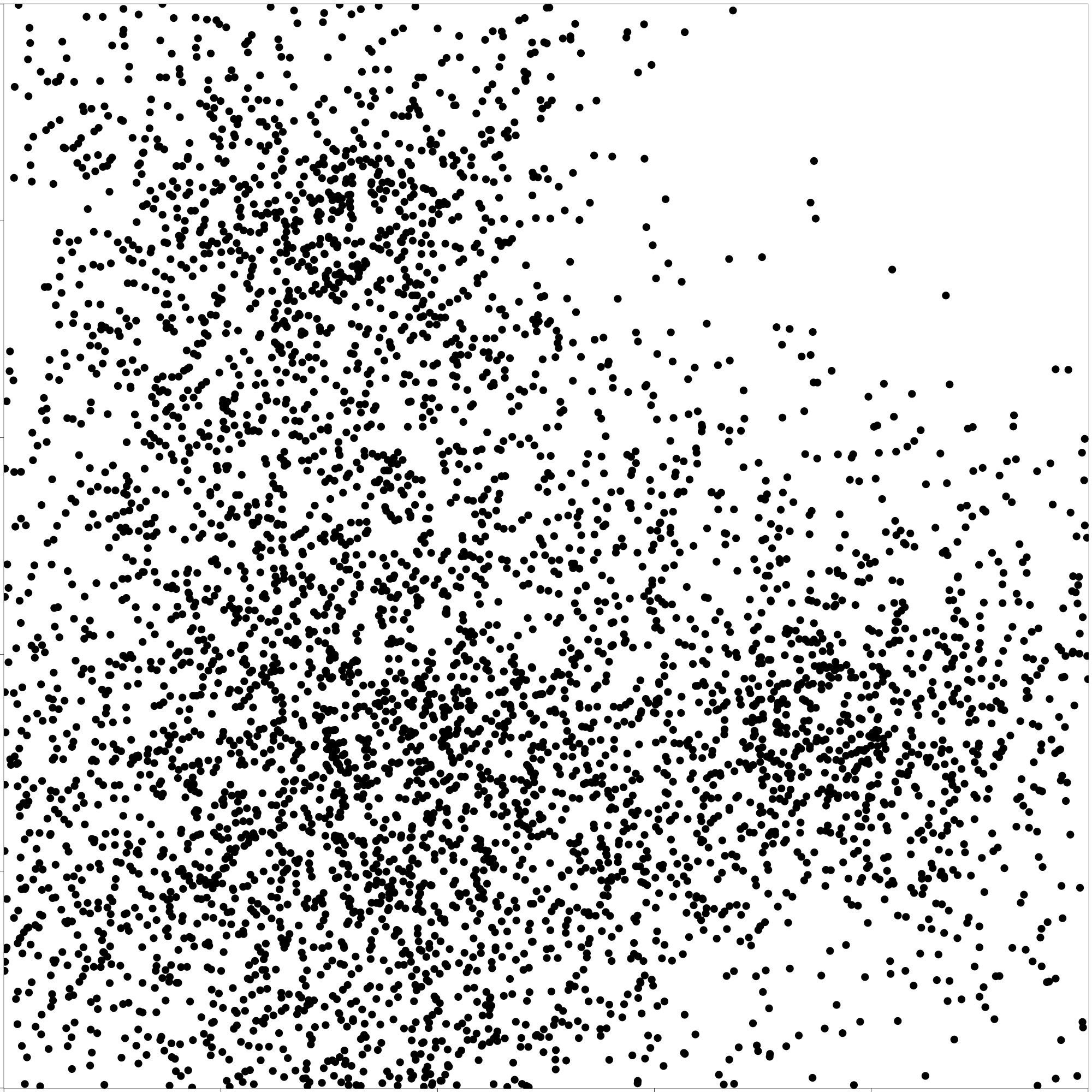}
\hspace*{0.03\textwidth}
\includegraphics[width=0.21\textwidth]{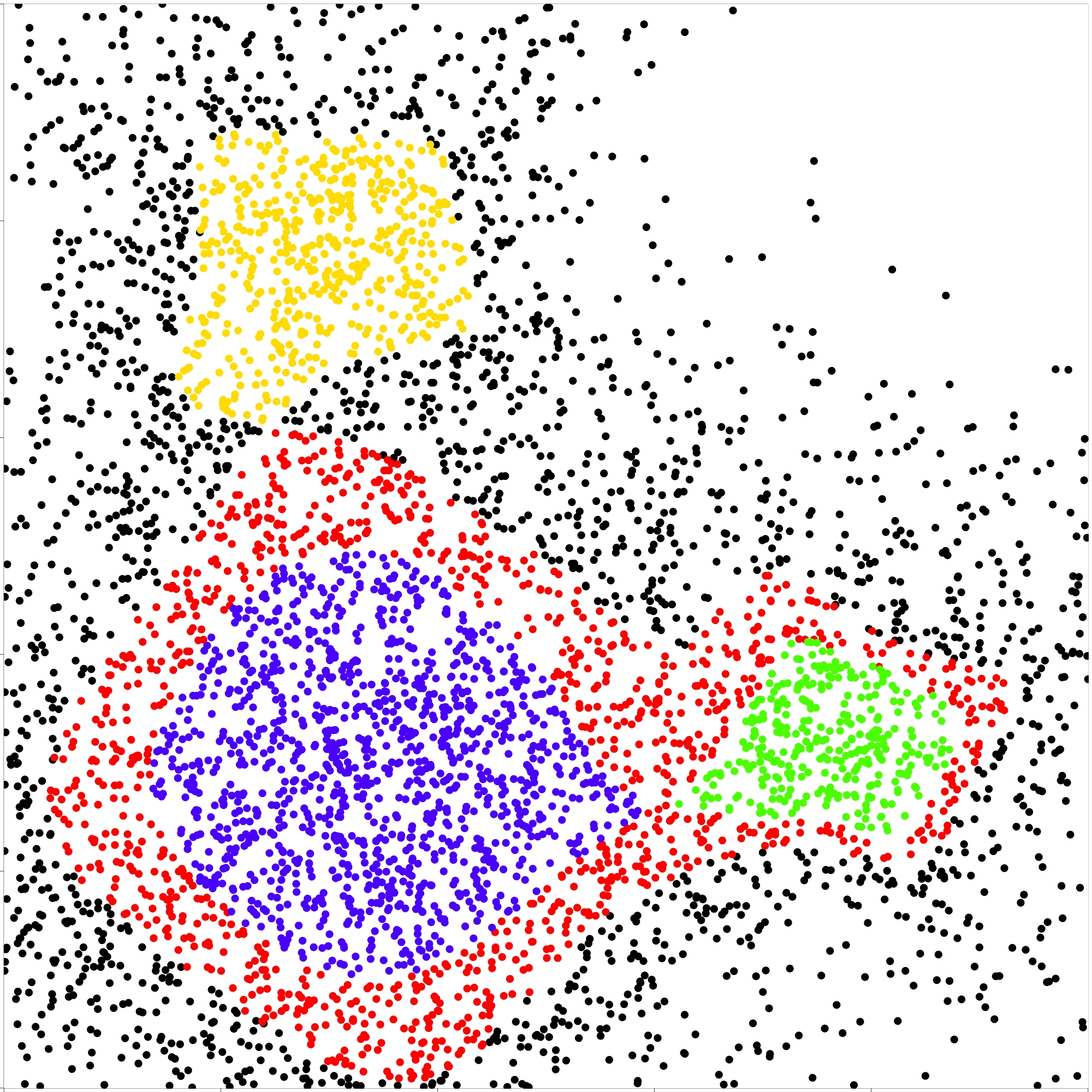}
\hspace*{0.03\textwidth}
\includegraphics[width=0.21\textwidth]{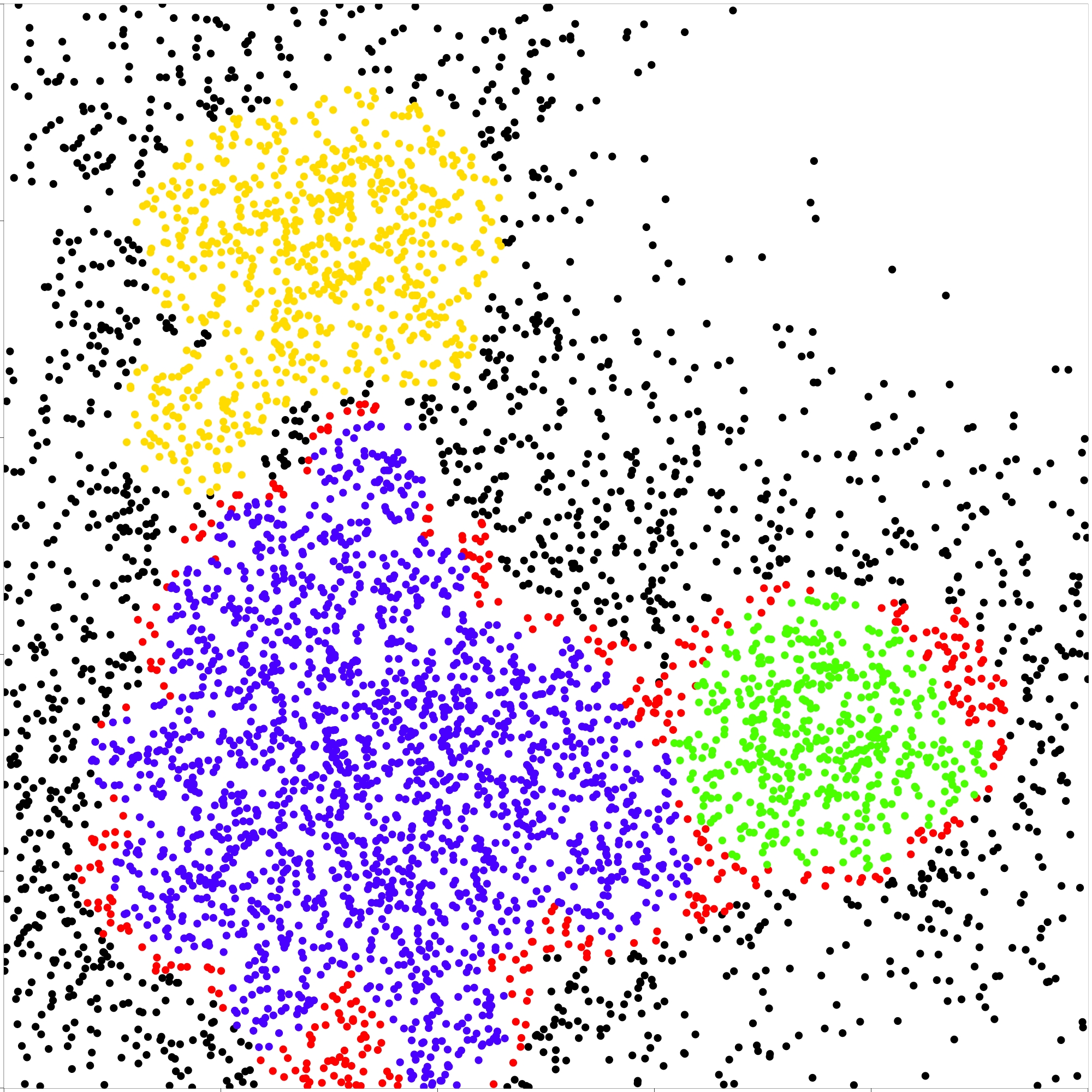}
\hspace*{0.03\textwidth}
\includegraphics[width=0.21\textwidth]{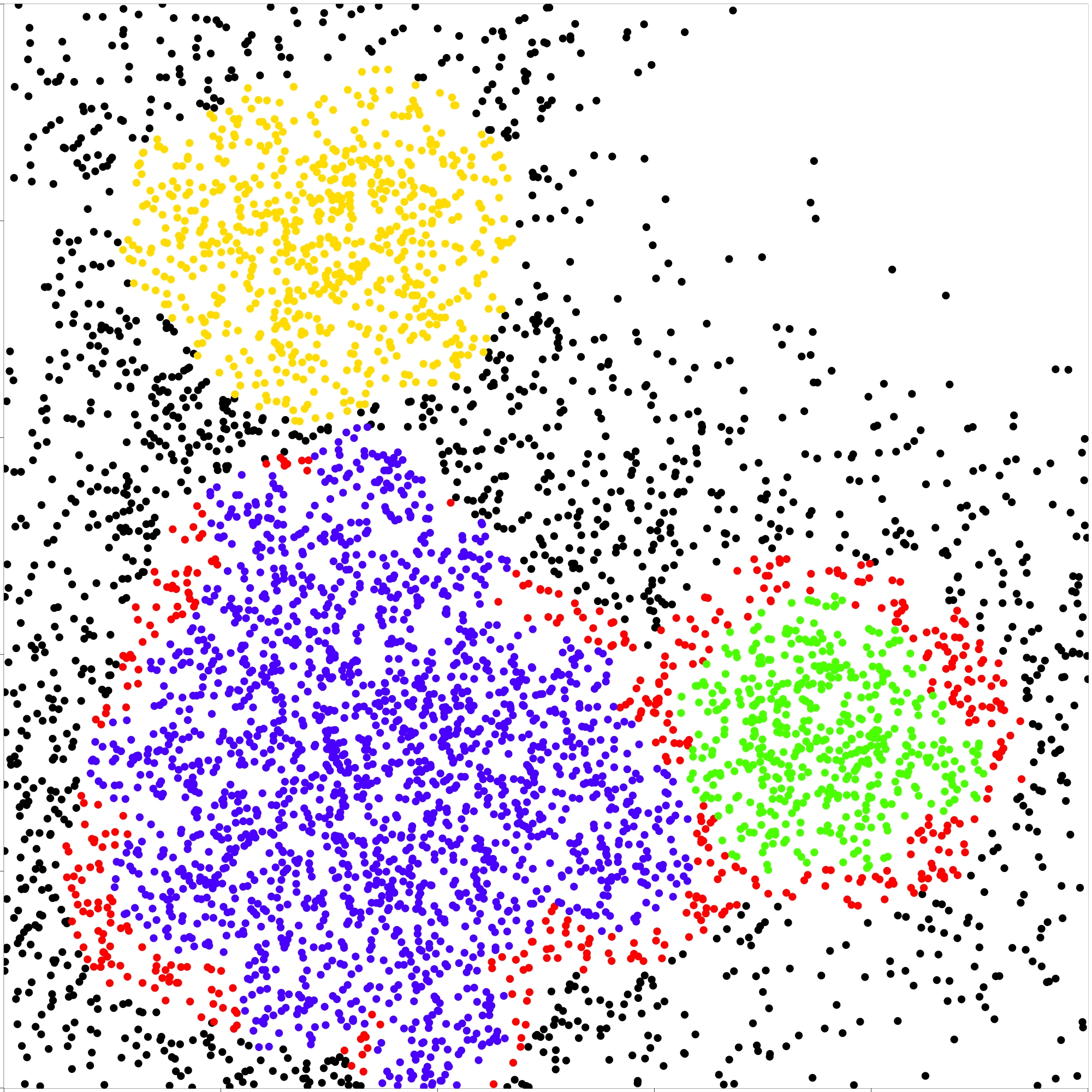}
\caption{\textbf{Left.} 5000 samples from 6 Gaussians with paired common means.
\textbf{Middle-Left.} Ground truth clustering on the data set obtained with the help of the density.
\textbf{Middle-Right and Right.} Clusters obtained from the data with the moving window and Epanechnikov kernel, respectively.
All clusters and the structure of the cluster tree  are correctly estimated. However, for both kernels, the sizes of   clusters are somewhat over-estimated. 
}
\label{figure:crosses}
\end{figure}

\section{Proofs}\label{sec:proofs}

\subsection{Proofs for the Generic Algorithm in Section \ref{sec:smg}}

\begin{lemma}\label{lemma:inrad}
 For   all bounded $A\subset \Rd$, and all $\d>0$ we have $A_\Rd\mde\neq \emptyset$ if and only if $\d<\inrad A$. 
\end{lemma}

\begin{proofof}{Lemma \ref{lemma:inrad}}
 Let us first assume that $A\mde \neq \emptyset$. Then there exists an $x\in A\mde$ and since 
 $A\mde$ is open there also exists an $\e>0$ with $B(x,\e)\subset A\mde$. 
 Our first intermediate goal is to 
 show
 \begin{align}\label{lemma:inrad-h1}
  B(y,\d) \subset A \, , \qquad\qquad \mbox{ for all } y\in B(x,\e)\, .
 \end{align}
To this end, we note that $y\in B(x,\e)\subset A\mde$ implies $y\not \in (\Rd\setminus A)\pde$.
For $z\in \Rd\setminus A$ we thus find 
$d(y,z) \geq d(y,\Rd\setminus A) >\d$, which shows \eqref{lemma:inrad-h1}.

Now observe that for the proof of  $\d<\inrad A$ it clearly suffices to  establish 
 \begin{align}\label{lemma:inrad-h2}
  B(x,\d+\e) \subset A\, .
 \end{align}
Let us therefore fix a $z\in B(x,\d+\e)$. By considering $y:= x$ in \eqref{lemma:inrad-h1}
we first note that in the case $\snorm{x-z}\leq \d$ we have $z\in A$. Hence it suffices to consider the case $\d<\snorm{x-z}\leq \d+\e$. 
For $t:= \d/\snorm{x-z}\in (0,1)$ and $y:= (1-t)z + tx$ a quick calculation then shows both 
$\snorm{y-z} = \d$ and $\snorm{x-y} = \snorm{x-z}-\d \leq \e$. Applying \eqref{lemma:inrad-h1}
then yields \eqref{lemma:inrad-h2}.

Let us now assume that $\d<\inrad A$. Then there exists an $\e>0$ with $\d+\e<\inrad A$ and hence we find 
 an $x\in A$ with $B(x,\d+\e) \subset A$. Clearly, it suffice to show that $x\in A\mde$.  To this end, we assume that $x\not \in A\mde$,
 that is $x\in (\Rd\setminus A)\pde$. Since this means $d(x,\Rd\setminus A)\leq \d$, we then find a $y\in \Rd\setminus A$
 with $d(x,y) \leq \d+ \e$. This contradicts $B(x,\d+\e) \subset A$.
\end{proofof}

\begin{lemma}\label{lemma:diam}
 For  all bounded $A\subset \Rd$, and all $\d>0$ we have $\diam A_\Rd\pde = 2\d + \diam A$.
\end{lemma}

\begin{proofof}{Lemma \ref{lemma:diam}}
 To establish $\diam A\pde \leq 2\d + \diam A$ we first some $x,y\in A\pde$. For $n\geq 1$ we then find 
 $x_n,y_n\in A$ with
 $d(x,x_n)\leq \d+1/n$ and $d(y,y_n)\leq \d+1/n$. This implies 
 \begin{align*}
  d(x,y) \leq d(x_n,y_n) + 2\d + 2/n \leq \diam A + 2\d + 2/n\, ,
 \end{align*}
and letting $n\to \infty$ then gives the assertion.

To establish the converse inequality, we pick, for all $n\geq 1$, some $x_n,y_n\in A$ with $d(x_n,y_n)\geq \diam A - 1/n$.
We define $t_n := 1 + \d /\snorm{x_n-y_n}$ and both $x_n^* := (1-t_n)y_n + t_nx_n$  and $y_n^* := (1-t_n)x_n + t_n y_n$.
These choices immediately give $\snorm{x_n-x_n^*} = \d = \snorm{y_n-y_n^*}$ and hence we have $x_n^*,y_n^*\in A\pde$.
Moreover, our construction ensures 
\begin {align*}
 \snorm{x_n^*-y_n^*} 
 = |1 - 2t_n| \cdot \snorm{x_n-y_n} 
 = \snorm{x_n-y_n} + 2\d
 \geq \diam A - 1/n + 2\d\, .
\end {align*}
Letting $n\to \infty$ then yields  $\diam A\pde \geq 2\d + \diam A$.
\end{proofof}

\begin{lemma}\label{lemma:inradius-decreases}
 Let \assx  S be satisfied. Then, for each $\d\in(0,\dthick]$, there exists a $\r\geq \rls$ with $\mrmde=\emptyset$.
\end{lemma}

\begin{proofof}{Lemma \ref{lemma:inradius-decreases}}
 Let us assume that the assertion is false, that is, there exists a $\d\in (0,\dthick]$ with $\mrmde \neq \emptyset$
 for all $\r\geq \rs$.  Lemma \ref{lemma:inrad} then gives $\d< \inrad \mr$ for all $\r\geq \rs$. 
 By the definition of the inradius, we thus find, for all $\r\geq \rs$, an $x_\r\in \mr$ with 
 $B(x_\r,\d) \subset \mr$. Let us fix a $\lbd$-density $h$ of $P$.
 Then we have 
 $\lbd(B(x_\r,\d) \setminus \mhx \r) \leq \lbd(\mr \setminus \mhx \r) = 0$, and consequently
 we find 
 \begin{align*}
  P(\mr) 
  \geq 
  P(B(x_\r,\d) ) 
  = 
  \int_{B(x_\r,\d) \cap \mhx \r} h\, d\lbd
  \geq 
  \r \, \lbd(B(x_\r,\d) \cap \mhx \r)
  =
  \r \cdot \d^d \cdot \lbd(B_{\snorm\cdot})\, .
 \end{align*}
Letting $\r\to \infty$ then leads to a contradiction.
\end{proofof}

\begin{lemma}\label{lemma:tau-connect-subsets}
 Let $B\subset \Rd$ be non-empty, compact subset and $\t>0$. Then we have $|\cct A|=1$ for all non-empty $A\subset B$ if and only if $\t>\diam B$.
\end{lemma}

\begin{proofof}{Lemma \ref{lemma:tau-connect-subsets}}
 Let us first assume that $|\cct A|=1$ for all non-empty $A\subset B$. If $|B| = 1$, then $\diam B=0$, and hence there is nothing to prove.
 Since $\snorm{\mycdot - \mycdot}:B\times B\to \R$ is continuous, there then exists $x,y\in B$ with $\snorm{x-y} = \diam B>0$.
 Now, our assumption ensures that  $A:= \{x,y\}$ is $\t$-connected, and this implies $\snorm{x-y}< \t$.
 
 Let us now consider a $\t>\diam B$. For $A\subset B$ and $x,y\in A$ we then find $\snorm{x-y} \leq \diam B<\t$, and hence $A$ is $\t$-connected.
\end{proofof}

\begin{lemma}\label{psi-bound}
   Let $(X,d)$ be a   connected metric  space  and $A\subset X$ be an open or closed subset
        with $\emptyset\neq A\neq X$.
        Then, for all $\d>0$ we have $\psis_A(\d)  \geq \d$.
\end{lemma}

\begin{proofof}{Lemma \ref{psi-bound}}
In view of \cite[(A.5.1)]{SteinwartXXb1} it suffices to 
    show  $d(A,X\setminus A) = 0$. Let us assume the converse, that is 
        $\e:=d(A,X\setminus A) > 0$. Then, if $A$ is closed we know that $X\setminus A$ is open, and for 
        $x\in A$  and $y\in X$ with $d(x,y) < \e$ our assumption yields $y\in A$. In other words,
        the open ball with center $x$ and radius $\e$ is contained in $A$, and therefore $A$ is open, too.
However, this gives a partition of $X$ into two open and non-empty sets, which contradicts 
        the assumption that $X$ is connected. Clearly, by interchanging the roles of $A$ and $X\setminus A$, 
we analogously find $d(A,X\setminus A) = 0$ for $A$ open.
\end{proofof}


\begin{theorem}\label{estim-control-low}
   Let $\rls\geq 0$ and
   Assumption P be satisfied with $|\cc \mr|\leq 1$ for all $\r\geq \rls$.
Moreover, let
 $(\lr)_{\r\geq 0}$ be a decreasing family of sets $\lr\subset X$ such that 
\begin{equation}\label{generic-inclus}
  \mrs\subset \lr \subset  \mrl
\end{equation}
for some fixed $\e,\d>0$ and all $\r\geq \rls$.
For all $\r\geq \rls$ we then have:
\begin{enumerate}
   \item  If $\mrmdex{+3\e} \neq \emptyset$, then for 
 all 
$\t>2 \psis_{\mrx{+\e}}(\d)$
we have $|\cct \mrs|=1$ and 
 the corresponding  CRM 
$\z:\cct \mrs  \to  \cct\lr$ satisfies
\begin{equation}\label{main-thick-hfr-result}
    \cct\lr = \z\bigl(\cct\mrs\bigr) 
\cup \bigl\{B'\in \cct\lr:  B'\cap \lrp = \emptyset \bigr\}\, .
\end{equation}
\item If $\mrmdex{+\e} = \emptyset$, Assumption S is satisfied, and $\d\in (0,\dthick]$, 
then  we have 
\begin{equation}\label{main-thick-hfr-result-upper}
   \Bigl|\bigl\{  B\in \cct\lr: B\cap \lrp \neq\emptyset \bigr\}  \Bigr| \leq 1 \, , \qquad \qquad \t>2\cthick\d^\g\, .
\end{equation}
\end{enumerate}
\end{theorem}

\begin{proofof}{Theorem \ref{estim-control-low}}
\ada i
We first note that $\mrmdex{+3\e} \neq \emptyset$  implies $\mrs\neq \emptyset$.
By \eqref{tmde} we thus find 
$|\cct\mrs|\leq |\cc {\mrx{+\e}}|\leq  1$, and by the already observed  $\mrx{+\e}\neq\emptyset$ we conclude that 
$|\cct\mrs|= 1$.
To establish  \eqref{main-thick-hfr-result}
we now
write $A:= \mrs$ and $B:= \z(A)$. Our first intermediate goal is to 
 establish the following \emph{disjoint} union: 
\begin{align} \nonumber
 \cct\lr = \{B\} 
 &\cup \bigl\{B'\in \cct\lr\setminus \{B\}:  B'\cap \lrp \neq \emptyset \bigr\} \\ \label{ghost-motiv-2}
& \cup \bigl\{B'\in \cct\lr:  B'\cap \lrp = \emptyset \bigr\}\, .
\end{align}
To this end,   note that 
$\mrmdex{+3\e} \neq \emptyset$ 
and $\mrmdex{+3\e} \subset A$
together with   $A\subset \z(A) = B$ implies
\begin{displaymath}
   \emptyset \neq \mrmdex{+3\e} = A \cap \mrmdex{+3\e} \subset B\cap \lrp\, .
\end{displaymath}
This yields 
$ \{B'\in \cct\lr\setminus \{B\}:  B'\cap \lrp = \emptyset \} 
= \{B'\in \cct\lr :  B'\cap \lrp = \emptyset \}$,
which in turn   implies \eqref{ghost-motiv-2}.

Let us now show \eqref{main-thick-hfr-result}.
To this end, we first observe that $|\cct \mrs|=1$ implies 
$\cct\mrs =\{A\}$ and hence $\z\bigl(\cct\mrs\bigr) = \{B\}$.
In view of  \eqref{ghost-motiv-2} 
it thus remains to show 
\begin{displaymath}
 B' \cap \lrp  = \emptyset\, ,
\end{displaymath}
for all $B'\in \cct\lr$ with $B'\neq B$. 
Let us  assume  the converse, that is, there is a
$B'\in \cct\lr$ with $B'\neq B$ and 
$B'\cap \lrp  \neq \emptyset$. 
Since  $L_{\r+2\e}  \subset     \mrpdex{+\e}$, there then exists
an $x\in B' \cap \mrpdex{+\e}$. 
By part \emph{i)} of  \cite[Lemma A.3.1]{SteinwartXXb1}
this gives an $x'\in\mrx{+\e}$ with $d(x,x')\leq \d$, and hence
we obtain 
\begin{displaymath}
 d(x', \mrs ) \leq \psis_{M_{\r+\e}}(\d) 
< \frac \t2\, .
\end{displaymath}
From this inequality we conclude that there is an $x'' \in \mrs$ satisfying 
$d(x',x'') < \t/2$. 
The CRM property then yields
$x''\in\mrs = A\subset B$ and therefore Lemma \ref{psi-bound} yields
%
\begin{displaymath}
 d(B', B) \leq d(x, x'') 
\leq d(x,x') + d(x',x'') 
< \d + \t/2
\leq  \psis_{M_{\r+\e}}(\d) + \t/2
< \t
\end{displaymath}
On the other hand,  $B'\neq B$ implies $d(B', B'') \geq \t$ by \cite[Lemma A.2.4]{SteinwartXXb1},
and hence we have found a contradiction.

\ada {ii} Clearly, if $\lrp=\emptyset$ there is nothing to prove, and hence we may assume that 
$\lrp \neq \emptyset$.
Now assume that \eqref{main-thick-hfr-result-upper} is false. Then there exist 
$B_1,B_2\in \cct\lr$ with $B_1\neq B_2$ and $B_i\cap \lrp\neq\emptyset$ for $i=1,2$.
For $i=1,2$ we 
consequently find $x_i \in B_i\cap \lrp$, and for these there exist
$A_i\in \cct\lrp$ with $x_i\in A_i$. Now recall from 
\cite[Lemma A.2.7]{SteinwartXXb1}
that $\lrp\subset \lr$ implies
$\cct\lrp\comparable\cct\lr$, and therefore we have a CRM $\z:\cct\lrp\to\cct\lr$.
Our construction then gives
\begin{displaymath}
   x_i\in A_i\cap B_i \subset \z(A_i)\cap B_i\, , 
\end{displaymath}
and therefore we have $\z(A_i)\cap B_i \neq \emptyset$ for $i=1,2$. However, $\z(A_i)$ and $B_i$ are both 
elements of the partition $\cct\lr$ and hence we conclude $\z(A_i) = B_i$ for $i=1,2$. Moreover, $\z$ is a map,
and therefore $B_1\neq B_2$ implies $A_1\neq A_2$.
Let us write $A:= A_1\cup A_2$. Since we know from \cite[Lemma A.2.4]{SteinwartXXb1}
that $d(A_1,A_2) \geq \t$, we conclude by \cite[Lemma A.2.8]{SteinwartXXb1} that $\cct A = \{A_1,A_2\}$,
and thus $|\cct A|=2$. However, we also have 
\begin{displaymath}
   A \subset \lrp \subset \mrpdex{+\e}\, ,
\end{displaymath}
and since  $\mrs =\emptyset$ holds, Assumption S together with 
$\d\in (0,\dthick]$ and  $\t>2\cthick\d^\g$ ensures $|\cct A|= 1$.
Since this contradicts    $|\cct A|=2$ we have proven \eqref{main-thick-hfr-result-upper}.
\end{proofof}

\begin{proofof}{Theorem \ref{main-generic-single}}
   For $i\geq 0$ we write $\r_i := \r_0 +  i \e$ for the sequence of potential levels Algorithm \ref{cluster-algo-generic} visits.
Moreover, let $i^* := \max\{i\geq 0: M_{\r_i+3\e}\mde\neq\emptyset\}$, where we note that 
this maximum is finite by Lemma \ref{lemma:inradius-decreases}.
For $i=0,\dots,i^*$, part \emph{i)} of Theorem \ref{estim-control-low} then shows that 
Algorithm \ref{cluster-algo-generic} identifies exactly one component in its Line 3, and therefore
it only identifies more than one component in   Line 3, if $i\geq i^*+1$. 
If it finishes the loop at   Line 5, we thus know that $\r\geq \r_{i^*+2}$, and therefore 
the level $\r$ considered in   Line 7 satisfies $\r\geq \r_{i^*+4}$. Now the definition of 
$i^*$ yields $M_{\r_{i^*+1}+3\e}\mde=\emptyset$, and 
since $\r_{i^*+1}+3\e =  (i^*+1)\e + 3\e = (i^*+4)\e = \r_{i^*+4}$, we find 
$M_{\r}\mde = \emptyset$ for the $\r$ considered in   Line 7. This implies $\mrs=\emptyset$, and hence 
part \emph{ii)} of Theorem \ref{estim-control-low} shows that Algorithm \ref{cluster-algo-generic}
identifies at most one component in  Line 7.
\end{proofof}

\begin{lemma}\label{result:cutting-tubes-from-intersections}
 Let $(X,d)$ be a metric space, $A,B\subset X$ be two subsets, and $\d>0$. Then we have 
 \begin{displaymath}
  \bigl(A\cap B)\mde = A\mde \cap B\mde\, .
 \end{displaymath}
\end{lemma}

\begin{proofof}{Lemma \ref{result:cutting-tubes-from-intersections}}
 From part \emph{iv)} of  \cite[Lemma A.3.1]{SteinwartXXb1} we know that 
 $(\tilde A\cup \tilde B)\pde = \tilde A\pde \cup \tilde B\pde$ for all $\tilde A, \tilde B\subset X$ and $\d>0$.
 This gives
 \begin{align*}
  \bigl(A\cap B)\mde 
  = X \setminus \bigl( X\setminus (A\cap B)\bigr)\pde
  &= X \setminus \bigl( (X\setminus A) \cup (X\setminus B)\bigr)\pde \\
  & = X \setminus \bigl( (X\setminus A)\pde \cup (X\setminus B)\pde \bigr) \\
  & = \bigl( X \setminus (X\setminus A)\pde\bigr) \cap \bigl( X \setminus (X\setminus B)\pde \bigr)\\
  & = A\mde \cap B\mde\, ,
 \end{align*}
and hence we have show the assertion.
\end{proofof}

\begin{lemma}\label{result:connected-components}
 Let \assx  M  be satisfied, $\r \in (\rs, \rss]$, $\e := \r-\rs$, and 
 $\ccr 1\r$, and $\ccr 2\r$ be the two connected components of $\mr$, i.e.~$\cc\mr = \{\ccr 1\r,\ccr 2\r\}$.
 Then the following statements hold:
\begin{enumerate}
 \item For all $0<\t \leq 3\ts(\e)$ we have $\cct\mr = \cc\mr$.
 \item For all $0<\d<\t \leq  \ts(\e)$ we have $\cct\mr \persist \cct\mrpde = \{\ccr 1\r\pde  , \ccr 2\r\pde\}$.
 \item For all $\d\in (0,  \dthick]$ and  $\psi(\d) <\t \leq  \ts(\e)$ we have $|\cct \mrmde |=2$ with  $\cct \mrmde = \{\ccr 1\r\mde,\ccr 2\r\mde\}$.
%
\end{enumerate}
\end{lemma}

\begin{proofof}{Lemma \ref{result:connected-components}}
To adapt to the notation of \cite{Steinwart15a,SteinwartXXb1} we write $\t_{\mr}^* := d(\ccr 1\r,\ccr 2\r)$.
Note that this definition gives $\t_{\mr}^* = 3\ts(\r-\rs) = 3\ts(\e)$.

 \ada i  The assertion directly  follows part \emph{ii)} from \cite[Proposition A.2.10]{SteinwartXXb1}.
 
 \ada {ii} The assertion has been shown in  part \emph{iii)} of   \cite[Lemma  A.4.1]{SteinwartXXb1}.
 
 \ada {iii} We first note that using  part \emph{ii)} of \cite[Theorem 2.7]{Steinwart15a}  
 with $\e^*:= \e$ and $\r = \r^* + \e^*$ we find $|\cct \mrmde|= |\cc \mr|= 2$. Moreover, we have 
 \begin{align}\label{result:connected-components-h1}
  d(\ccr 1\r,\ccr 2\r) = \t_{\mr}^* = 3\ts(\e) \geq \t > \psi(\d) = 3\cthick \d^\g > \psis_\mr(\d) \geq \d\, ,
 \end{align}
where the last inequality follows from Lemma \ref{psi-bound} since Assumption M implies Assumption P, and hence $X$
is connected.
 Consequently, 
  part \emph{v)} of   \cite[Lemma  A.3.1]{SteinwartXXb1} yields
  \begin{align}\label{result:connected-components-h2}
   \mrmde = \bigl(\ccr 1\r \cup \ccr 2\r  \bigr)\mde =  \ccr 1\r\mde \cup \ccr 2\r\mde\, ,
  \end{align}
 and we additionally note that \eqref{result:connected-components-h1} implies 
\begin{align}\label{result:connected-components-h3}
 d(\ccr 1\r\mde,\ccr 2\r\mde) \geq d(\ccr 1\r,\ccr 2\r)  \geq \t\, .
\end{align}
Now let $A_1$ and $A_2$ be the two $\t$-connected components of $\cct \mrmde$. Let us assume that 
$A_1\neq \ccr 1\r\mde$ and $A_1 \neq \ccr 2\r\mde$. Then 
\eqref{result:connected-components-h2} shows that there exist $x' \in A_1 \cap \ccr 1\r\mde$
and $x'' \in A_1 \cap \ccr 2\r\mde$. Since $A_1$ is $\t$-connected, there further exist $x_1,\dots,x_n\in A_1$
with $x_1 = x'$, $x_n = x''$ and $d(x_i, x_{i+1}) < \t$ for all $i=1,\dots,n-1$.
By $x_1\in \ccr 1\r\mde$, $x_n\in \ccr 2\r\mde$, and \eqref{result:connected-components-h2} we 
conclude that there is an $i=\{1,\dots,n-1\}$ with $x_i\in \ccr 1\r\mde$ and  $x_{i+1}\in \ccr 2\r\mde$.
This gives 
\begin{displaymath}
 d(\ccr 1\r\mde,\ccr 2\r\mde) \leq  d(x_i, x_{i+1}) <  \t  \, , 
\end{displaymath}
which clearly contradicts \eqref{result:connected-components-h3}.
\end{proofof}

\begin{lemma}\label{result:level-sets-of-childs}
 Let \assx  M be satisfied, and $P_1$ and $P_2$ be defined by \eqref{child-measures}
 for some fixed $\rda \in  (\rs, \rss]$. Then for $i=1,2$ and $\r\geq \rda$ we have 
  \begin{align}\label{result:level-set-computation-h1}
  \mri i = \mr \cap \ccr i\rda\,   .
 \end{align}
\end{lemma}

\begin{proofof}{Lemma \ref{result:level-sets-of-childs}}
 We first note that since $P$ is normal at all levels $\r>0$, we have 
 $\mu(\mr \symdif \{h\geq \r\}) = 0$ for all $\mu$-densities $h$ of $P$ and all $\r> 0$.
 For a  fixed  $\r\geq \rda>0$.
  we can thus find a $\mu$-density $h$ of $P$ such that $\mr = \{h\geq \r\}$ and
 $\mx \rda = \{h\geq \rda\}$. Let us define $h_i := \eins_{\ccr i\rda}   h$. Then $h_i$ is 
 a $\mu$-density of $P_i$ and we have 
 \begin{align}\label{result:level-sets-of-childs-h1}
  \{h_i\geq \r\}  = \mr \cap \ccr i\rda\, .
 \end{align}
    Moreover, by our definitions we find 
    \begin{align*}
     \mri i = \supp \mu\bigl( \,\cdot\, \cap \{h_i \geq \r\}  \bigr) \, , 
    \end{align*}
    and hence it suffices to show that $\mri i = \{h_i\geq \r\}$. 
    
    For the proof of the inclusion ``$\subset$'' we fix 
    an $x\in \mri i$ and an open $U\subset X$ with $x\in U$. The definition of the support of a measure then yields 
    \begin{align*}
     \mu (  U \cap \mr) = \mu\bigl(U\cap \{h\geq \r\}\bigr) \geq \mu\bigl(U\cap \{h_i\geq \r\}\bigr) > 0\, ,
    \end{align*}
    which in turn implies $x\in \mr$. This shows $\mri i \subset \mr$. Moreover, $\ccr i\rda$
    is closed by definition and we further  have 
    \begin{displaymath}
     \mu\bigl( \ccr i\rda \cap \{h_i\geq \r\}\bigr) = \mu(\{h_i\geq \r\}) = \mu\bigl(X \cap \{h_i\geq \r\}\bigr) \, .
    \end{displaymath}
    Since the support of a finite measure is also the smallest closed subset having full measure, we conclude that 
    $\mri i \subset \ccr i\rda$. Combining the two found inclusions $\mri i \subset \mr$ and 
    $\mri i \subset \ccr i\rda$ with \eqref{result:level-sets-of-childs-h1} we have thus found the desired 
    $\mri i \subset \{h_i\geq \r\}$.
    
    For the proof of the converse inclusion we fix an $x\in \{h_i\geq \r\} = \mr \cap \ccr i\rda$. Moreover, we fix an 
    open $U\subset X$ with $x\in U$, so that it suffices  to show $\mu(U\cap  \{h_i\geq \r\}) >0$. To this end, we
    may assume 
    without loss of generality  that $i=1$. Moreover, since $d(\ccr 1\rda, \ccr 2\rda)>0$ and $x\in \ccr 1 \rda$
    we may additionally assume that $U\cap \ccr 2\rda = \emptyset$. Now, $x\in \mr$ implies $\mu(U\cap \mr) >0$.
    Let us write $A_k := \mr \cap \ccr k\rda = \{h_k\geq \r\}$. This yields $\mr = A_1 \cup A_2$, $A_1\cap A_2 = \emptyset$, and
    \begin{align*}
     \mu(U \cap A_2 ) \leq \mu(U\cap \ccr 2\rda) = 0\, .
    \end{align*}
        Using the disjoint union $U\cap \mr = (U\cap A_1) \cup (U\cap A_2)$, we conclude that 
        \begin{align*}
         \mu(U\cap  \{h_1\geq \r\})  = \mu (U\cap A_1) = \mu(U\cap \mr) >0\, .
        \end{align*}
    As mentioned above this shows $x\in \mri 1$.
\end{proofof}

\begin{lemma}\label{result:level-set-computation}
 Let \assx  M be satisfied, and $P_1$ and $P_2$ be defined by \eqref{child-measures}
 for some fixed $\rda \in  (\rs, \rss]$. 
 Then, for $i=1,2$, the following statements are true:
 \begin{enumerate}
  \item For $\eda := \rda - \rs$  and   all   $0<\d<  \ts(\eda)$ and $\r\geq \rda$ we have $\mripde i = \mrpde \cap \ccr i\rda\pde$.
  \item For all $\d > 0$ and $\r\geq \rda$ we have $\mrimde i = \mrmde \cap \ccr i\rda\mde$.
 \end{enumerate}
\end{lemma}

\begin{proofof}{Lemma \ref{result:level-set-computation}}
%
\ada i 
Let $\xi: \cc\mr \to \cc {\mx\rda}$ be the CRM and  
$B_1,\dots,B_n$ be the connected components of $\cc\mr$.  
Without loss of generality we may assume there is an $m\in \{0,\dots, n\}$ such 
that $\xi(B_j) \subset \ccr 1\rda$ for all $j=1,\dots, m$ and 
 $\xi(B_j) \subset  \ccr 2\rda$ for all $j=m+1 , \dots, n$.
 We define $A_1 := B_1 \cup \dots \cup B_m$ and $A_2 := B_{m+1}\cup\dots\cup B_n$.
 Clearly, this construction ensures 
 \begin{align}\label{result:level-set-computation-h4}
  A_k \subset \xi(A_k) \subset \ccr k\rda\,  , \qquad \qquad k=1,2.
 \end{align}
 Moreover, we have $\mr = A_1 \cup A_2$, and hence we find 
 \begin{align*}
  \mrpde = A_1\pde \cup A_2\pde
 \end{align*}
by part \emph{iv)} of \cite[Lemma A.3.1]{SteinwartXXb1}. In view of \eqref{result:level-set-computation-h1}, we
 consequently  need to prove that 
 \begin{align}\label{result:level-set-computation-h2}
  \bigl((A_1\cup A_2) \cap \ccr i\rda\bigr)\pde  =  \bigl(A_1\pde \cup A_2\pde\bigr) \cap \ccr i\rda\pde\, .
 \end{align}
 Be begin by observing that 
 \begin{align}\label{result:level-set-computation-h9}
  (A_1\cup A_2) \cap \ccr i\rda
  = ( A_1\  \cap \ccr i\rda) \cup ( A_2\  \cap \ccr i\rda)
  = A_i\, ,
 \end{align}
where we used both \eqref{result:level-set-computation-h4} and 
$\ccr 1\rda \cap \ccr 2\rda = \emptyset$.
 Similarly, the right-hand side of \eqref{result:level-set-computation-h2} can be written as 
 \begin{align}\label{result:level-set-computation-h3}
   \bigl(A_1\pde \cup A_2\pde\bigr) \cap \ccr i\rda\pde
   = 
   \bigl(A_1\pde \cap \ccr i\rda\pde\bigr) \cup \bigl(A_2\pde \cap \ccr i\rda\pde\bigr)\, .
 \end{align}
 In addition, \eqref{result:level-set-computation-h4} ensures $A_i\pde \subset \ccr i\rda\pde$,
 and by continuing \eqref{result:level-set-computation-h3}
 we thus find for $k\in \{1,2\}$ with $k\neq i$ that 
 \begin{align}\label{result:level-set-computation-h5}
   \bigl(A_1\pde \cup A_2\pde\bigr) \cap \ccr i\rda\pde
   = 
    A_i\pde \cup \bigl(A_k\pde \cap \ccr i\rda\pde\bigr)\, .
 \end{align} 
Moreover, by part  
\emph{ii)} of
Lemma \ref{result:connected-components} we know that $\ccr 1\rda\pde$ and 
 $\ccr 2\rda\pde$ are the two $\t$-connected components of 
 $\mxpde\rda$ for any $\t$ with  $\d  <\t< \ts(\eda)$. Consequently, we have 
 $\ccr 1\rda\pde \cap \ccr 2\rda\pde = \emptyset$, and since 
\eqref{result:level-set-computation-h4} ensures $A_k\pde \subset \ccr k\rda\pde$
 we conclude that $A_k\pde \cap \ccr i\rda\pde = \emptyset$.
 Inserting the latter into \eqref{result:level-set-computation-h5}
 gives
 \begin{align}\label{result:level-set-computation-h6}
  \bigl(A_1\pde \cup A_2\pde\bigr) \cap \ccr i\rda\pde = A_i\pde\, .
 \end{align}
Now, \eqref{result:level-set-computation-h2} follows from combining 
\eqref{result:level-set-computation-h9} with \eqref{result:level-set-computation-h6}.

 \ada {ii} This directly follows from combining \eqref{result:level-set-computation-h1} with Lemma 
 \ref{result:cutting-tubes-from-intersections}.
\end{proofof}

\begin{proofof}{Theorem \ref{result:acting-in-the-gray-zone}}
 \ada i We first note that $\e^* \leq \e^{**} := \rss-\rs$ implies
 $\ts(\e^*) \leq \ts(\e^{**})$. Consequently, part \emph{iii)} of Lemma \ref{result:connected-components}
 applied for $\rho:= \rss$ gives the assertion.
 
 \ada {ii} We begin by showing that the CRMs 
 $\xi_{\r+\e}: \cct {\mxmde\rss} \to \cct{\mxmde{\r+\e}}$
 and 
 $\xi:  \cct{\mxmde{\r+\e}}\to \cct{\mxmde{\rout+\e}}$
 are bijective. To this end we consider the following commutative diagram of CRMs:
 \begin{displaymath}
 \tridia {\cct {\mxmde\rss}} {\cct {\mxmde{\rout+\e}}} {\cct{\mxmde{\r+\e}}} {\xi_{\rout+\e}}  {\xi_{\r+\e}}\xi
 \hspace*{20ex}
 \end{displaymath}
 Now, part \emph{iv)} of \cite[Theorem A.6.2]{SteinwartXXb1} shows that $\xi_{\rout+\e}$ is bijective, and consequently, 
 $\xi_{\r+\e}$ is injective. 
 Moreover, part \emph{i)} of \cite[Theorem 2.7]{Steinwart15a}
 shows that $1\leq |\cct {\mxmde{\r+\e}}|\leq 2$, and since we already know that $|\cct {\mxmde\rss}| = 2$
 and that 
 $\xi_{\r+\e}$ is injective, we conclude that $|\cct {\mxmde{\r+\e}}|= 2$ and that  
    $\xi_{\r+\e}$ is bijective. 
  Using the diagram we then see that the CRM $\xi$ is also bijective.
 
 Our next goal is to show that  the CRM $\widehat \xi_\r: \cct{\mxmde{\r+\e}} \to \ccth{L_\r}$ is well-defined and bijective.
 To this end, we first recall that our assumption 
 $\r\leq \rss-3\e$ together with 
 \cite[Theorem 2.8]{Steinwart15a}   gives  the following  \emph{disjoint} union:
 \begin{align*}
  \cct{L_\r} 
  &= \widehat \xi_\r\bigl( \cct{\mxmde{\r+\e}}   \bigr)  \cup \bigl\{ B'\in \cct{L_\r}: B'\cap L_{\r+2\e} = \emptyset \bigr\} \, .
 \end{align*}
 Consequently, we have $\widehat \xi_\r\bigl( \cct{\mxmde{\r+\e}}   \bigr) = \ccth{L_\r}$, 
 that is, we can view $\widehat \xi_\r$ as a \emph{surjective  CRM} $\widehat \xi_\r: \cct{\mxmde{\r+\e}} \to \ccth{L_\r}$.
Similarly, part \emph{i)} of Theorem \ref{analysis-main-combined-new}
 ensures 
\begin{displaymath}
   \rout  \leq \rs+\e^*+5\e \leq \rs + 6\e^*\leq  \rss - 3\e\, ,
\end{displaymath}
  and repeating the reasoning above we see 
 that
 the CRM $\widehat \xi_\rout: \cct{\mxmde{\rout+\e}} \to \cct{L_\rout}$ can be viewed as a \emph{surjective CRM}
 $\widehat \xi_\rout: \cct{\mxmde{\rout+\e}} \to \ccth{L_\rout}$.
 Finally,  consider the CRM $\breve \xi: \cct{L_\r}\to \cct{L_\rout}$. For $B\in \ccth{L_\r}$ we then have 
\begin{displaymath}
 \emptyset \neq B\cap L_{\r+2\e}
 \subset \breve\xi(B) \cap L_{\r+2\e} 
 \subset \breve\xi(B) \cap L_{\rout+2\e} \, , 
\end{displaymath}
i.e.~we have shown $\breve \xi(B) \in \ccth{L_\rout}$. Consequently, the restriction 
\begin{displaymath}
 \breve \xi_{| \ccth{L_\r}}: \ccth{L_\r} \to \ccth{L_\rout}
\end{displaymath}
is well-defined, and obviously also a CRM. 
Combining these considerations we obtain the following commutative diagram of CRMs
\begin{displaymath}
\quadiann {\cct{\mxmde{\rout+\e}}}{\ccth{L_\rout}}{\cct{\mxmde{\r+\e}}}{\ccth{L_\r}} {\widehat \xi_\rout}\xi{\breve \xi_{| \ccth{L_\r}}}{\widehat \xi_\r} \hspace*{20ex}
\end{displaymath}
Now, we have already seen that $\xi$ is bijective, and in addition, 
part \emph{ii)} of \cite[Theorem A.6.2]{SteinwartXXb1} shows that 
$\widehat \xi_\rout$ is injective. Moreover, our considerations above showed that 
 $\widehat \xi_\rout: \cct{\mxmde{\rout+\e}} \to \ccth{L_\rout}$ is surjective, and hence 
 the latter CRM is bijective. Using the diagram we conclude that 
 $\widehat \xi_\r: \cct{\mxmde{\r+\e}} \to \ccth{L_\r}$ 
 is   injective. Since we have already seen that it is surjective, we conclude that it is indeed bijective.

With the help of these preparations, the first assertion now easily follows from \emph{i)} and  the bijectivity of  $\widehat\xi_\r$ and $\xi_{\r+\e}$, namely 
\begin{displaymath}
 |\ccth{L_\r}| = \bigl|\widehat\xi_\r \circ \xi_{\r+\e} \bigl((\cct{\mxmde\rss}\bigr) \bigr|
 = \bigl|  \cct{\mxmde\rss} \bigr| = 2\, .
\end{displaymath}
To show the second assertion, we write $B_i^\rho := \widehat\xi_\r \circ \xi_{\r+\e}(V_i)$.
This immediately gives $V_i \subset B_i^\rho$ for $i=1,2$.
Moreover, using the diagram we find
\begin{align*}
 B_i^\r 
 \subset \breve \xi_{| \ccth{L_\r}}(B_i^\r) 
 = \breve \xi_{| \ccth{L_\r}}\circ \widehat\xi_\r \circ \xi_{\r+\e}(V_i) 
  = \widehat\xi_\rout \circ \xi \circ \xi_{\r+\e}(V_i) 
  = B_i\, ,
\end{align*}
where the latter identity follows from part \emph{iii)} of \cite[Theorem A.6.2]{SteinwartXXb1}.

\ada {iii} We first observe that $\eda := \rda - \rs$ satisfies $\eda \geq \e^*$ and by 
Lemma \ref{psi-bound} we
hence  find 
$\d \leq \psis(\d) < \ts(\e^*) \leq \ts(\eda)$.  Lemma \ref{result:level-set-computation}
then shows 
\begin{align*}
 \mrimde i = \mrmde \cap \ccr i\rda\mde \qquad \qquad \mbox{ and } \qquad \qquad \mripde i = \mrpde \cap \ccr i\rda\pde\, .
\end{align*}
By the definition of $L_{i,\r}$ we thus have to show the following two inclusions
\begin{align}\label{result:acting-in-the-gray-zone-uq-h1}
 \mrs \cap \ccr i\rda\mde & \subset L_\r \cap B_i \\ \label{result:acting-in-the-gray-zone-uq-h2}
   L_\r \cap B_i & \subset \mrl \cap \ccr i\rda\pde \, .
\end{align}
We begin by proving \eqref{result:acting-in-the-gray-zone-uq-h1}. To this end, we first observe that 
\eqref{uq} ensures $\mrs \subset  L_\r$ and hence it suffices to establish $\ccr i\rda\mde \subset B_i$.
Now, we have already observed that 
$\t \leq \ts(\e^*) \leq \ts (\eda)$, and consequently
part \emph{iii)} of Lemma \ref{result:connected-components} shows that 
$\ccr 1\rda\mde$ and $\ccr 2\rda\mde$ are the two $\t$-connected components of $\mxmde \rda$.
Moreover, part \emph{i)} of Theorem \ref{analysis-main-combined-new}
shows $\rout \leq \rs + \e^* + 5\e \leq \rda-\e$, and hence we have 
$\rda - \e \in [\rout, \rss-3\e]$.
Applying 
 \cite[Theorem 2.8]{Steinwart15a} and the already established 
part \emph{ii)}  to the level $\rda - \e$
we then obtain 
\begin{align*}
 \ccr i\rda\mde \subset B_i^{\rda - \e} \subset B_i\, , \qquad \qquad i=1,2\, .
\end{align*}

Let us now establish $L_{i,\r} \subset B_i^{\rda + 2\e}$.
Without loss of generality we may assume $i=1$.
Now, 
consider the CRM
$\xi: \cct{L_{\r}\cap B_1} \to \cct{L_{\rda+2\e}\cap B_1}$, which is possible since $\r\geq \rda + 2\e$.
Let us assume that there was a $B'\in \cct{L_{\r}\cap B_1}$ with 
\begin{displaymath}
   \xi (B') \not\subset  B_1^{\rda + 2\e}\, .
\end{displaymath}
Since $ B_1^{\rda + 2\e}$ is a $\t$-connected component of $L_{\rda+2\e}\cap B_1$
by part \emph{ii)} applied to the level $\rda+2\e \in [\rout, \rss-3\e]$ and $\xi (B')$ is another such 
$\t$-connected component we conclude that  $\xi (B')\cap   B_1^{\rda + 2\e} = \emptyset$.
Moreover, our construction and part \emph{ii)} give
\begin{displaymath}
\xi (B')\cap   B_2^{\rda + 2\e} \subset B_1 \cap   B_2^{\rda + 2\e} 
\subset B_1\cap B_2 =\emptyset \, ,
\end{displaymath}
and therefore part \emph{ii)} shows $\xi(B') \not\in \ccth{L_\rda+2\e}$. 
Together with $\r\geq \rda+4\e$ the latter implies  
\begin{displaymath}
 B'\cap L_\r \subset B' \cap L_{\rda + 4\e} \subset \xi (B')\cap    L_{\rda + 4\e} = \emptyset\, .
\end{displaymath}
Consequently, we have found a contradiction, and therefore we have 
$\xi (B')  \subset  B_1^{\rda + 2\e}$ for all $\t$-connected components of $L_{1,\r}  = L_{\r}\cap B_1$.
Since $B'\subset \xi(B')$ we have thus found  $L_{1,\r} \subset B_1^{\rda + 2\e}$.

Let us now show \eqref{result:acting-in-the-gray-zone-uq-h2}. 
To this end, we note that \eqref{uq}
ensures $L_\r \subset \mrl$, and hence it suffices to prove 
$L_\r \cap B_i   \subset   \ccr i\rda\pde$.
Moreover, we have already shown that  $L_\r \cap B_i  \subset B_i^{\rda + 2\e}$, and therefore, it suffices to 
establish
\begin{displaymath}
   B_i^{\rda + 2\e} \subset  \ccr i\rda\pde\, .
\end{displaymath}
To this end, recall that we have 
 already observed $\t\leq \ts(\e^*) \leq \ts(\eda)$. 
Part \emph{ii)} of Lemma \ref{result:connected-components} thus shows that 
$\ccr 1\rda\pde$ and $\ccr 2\rda\pde$ are the two $\t$-connected components of $\mxpde\rda$.
Now consider the CRM $\xi: \cct{L_{\rda+2\e}} \to \cct {\mxpde\rda}$.
Then the $\t$-connected component   $B_1^{\rda+2\e}$ of $L_{\rda+2\e}$ satisfies $B_1^{\rda+2\e} \subset \xi(B_1^{\rda+2\e})$,
and therefore, exactly one of the following two conditions is satisfied
\begin{align}\label{result:acting-in-the-gray-zone-final-choice-1}
 B_1^{\rda+2\e} & \subset \ccr 1\rda\pde \, , \\ \label{result:acting-in-the-gray-zone-final-choice-2}
 B_1^{\rda+2\e} & \subset \ccr 2\rda\pde \, .
\end{align}
However, our construction ensures $V_1 \subset \ccr 1\rda\pde$, and part \emph{ii)}
gives $V_1 \subset  B_1^{\rda+2\e}$. This gives $\emptyset\neq V_1 \subset B_1^{\rda+2\e} \cap \ccr 1\rda\pde$,
and therefore we can 
exclude \eqref{result:acting-in-the-gray-zone-final-choice-2}. Consequently
 \eqref{result:acting-in-the-gray-zone-final-choice-1} is true.
 The inclusion $ B_2^{\rda+2\e}   \subset \ccr 2\rda\pde$ can be shown analogously.
%
%
%
%
%
%
%
\end{proofof}

\subsection{Proofs for Section \ref{sec:kde}}


\begin{proofof}{Lemma \ref{lem:tail-functions}}
 For the tail function $\ktof$ the estimate follows from
 \begin{align*}
  \kto r 
  &= \int_{\Rd\setminus B(0,r)} K(x) \,\dld(x) 
   \leq c \int_{\Rd\setminus B(0,r)} \exp\bigl(-\tnorm x\bigr) \,\dld(x)\\
  & = c d \vold \int_r^\infty e^{-s} s^{d-1} ds
   \leq c d^2 \vold  e^{-r} r^{d-1} \, ,
 \end{align*}
where the last estimate for the incomplete gamma function is taken from \cite[Lemma A.1.1]{StCh08}.
The second inequality   follows from the monotonicity of the function $r\mapsto e^{-r}$.
\end{proofof}

\begin{lemma}\label{inclusion-aux}
   Let $K:\R^d\to [0,\infty)$ be a symmetric kernel with  tail function $\ktof$.
Moreover, let
$P$
be a $\lbd$-absolutely continuous distribution on $\R^d$ that is normal at some level $\r\geq 0$.
Then for all $x\in \Rd$ and $\s>0$ with $B(x,\s)\subset \mr$ and all $\d>0$ we have 
\begin{equation}\label{inclusion-aux-h1}
   \hpd(x) \geq \r - \r\kto{\tfrac \s\d} 
\end{equation}
while for all $x\in \Rd$ and $\s>0$ with 
 $B(x,\s)\subset X\setminus\mr$ and all $\d>0$ we have 
\begin{equation}\label{inclusion-aux-h2}
   \hpd(x) < \r  + \d^{-d}\kti {\tfrac \s\d}\, .
\end{equation}
Finally, if $P$ has a bounded density $h$, 
%
%
then the inequality \eqref{inclusion-aux-h1} 
can be replaced by 
\begin{equation}\label{inclusion-aux-h1-bounded}
   \hpd(x) \geq \r - \kto{\tfrac \s\d}\cdot \inorm{h}
\end{equation}
whenever $0\leq \r\leq \inorm h$  and \eqref{inclusion-aux-h2} can be replaced, for all $\r\geq 0$,  by 
\begin{equation}\label{inclusion-aux-h2-bounded}
   \hpd(x) < \r + \kto{\tfrac \s\d}\cdot \inorm{h}\, .
\end{equation}
\end{lemma}

\begin{proofof}{Lemma \ref{inclusion-aux}}
Let $h$ be a $\lbd$-density of $P$.
We begin by proving \eqref{inclusion-aux-h1}. To this end, we first observe that 
 $  \lbd (B(x,\s)\setminus \{h\ge \rho\} )\leq
\lbd (M_{\rho}\setminus\{h\ge \rho\} )=0$,
since $P$ is
normal at   level $\rho$.
Therefore, we  obtain 
\begin{align}\nonumber
\int_{B(x,\s)}K_\d(x-y)  \,  h(y)\,\dld(y)
&= \int_{B(x,\s)\cap \{h\ge \rho\}}K_\d(x-y)  \,  h(y)\,\dld(y)\\ \nonumber 
& \geq \rho  \int_{B(x,\s)\cap \{h\ge \rho\}}K_\d(x-y) \,\dld(y)\\ \label{inclusion-aux-h1-h}
& = \rho  \int_{B(x,\s)}K_\d(x-y) \,\dld(y)\, ,
\end{align}
and this leads to 
\begin{align*}
 \hpd(x)
&=  \int_{\Rd}K_\d(x-y)  \,  h(y)\,\dld(y) \\
&\geq \rho  \int_{B(x,\s)}K_\d(x-y) \,\dld(y)
 + \int_{\Rd\setminus B(x,\s)}K_\d(x-y)  \,  h(y)\,\dld(y)\\
&= \rho  \int_{B(x,\s)}K_\d(x-y) \,\dld(y)
 +  \rho  \int_{\Rd\setminus B(x,\s)}K_\d(x-y) \,\dld(y)
 \\&\qquad-  \rho  \int_{\Rd\setminus B(x,\s)}K_\d(x-y) \,\dld(y)
  + \int_{\Rd\setminus B(x,\s)}K_\d(x-y)  \,  h(y)\,\dld(y)\\
&\geq \rho 
 -  \rho  \int_{\Rd\setminus B(x,\s)}K_\d(x-y) \,\dld(y)\, ,
\end{align*}
where in the last step we used \eqref{shift-int-kernel-new}.
In the case of a general density $h$ the 
 assertion now follows from \eqref{inclusion-aux-conversion},
and for a bounded density $h$ and $\r\leq \inorm h$ the inequality
\eqref{inclusion-aux-h1-bounded} is a direct consequence of \eqref{inclusion-aux-h1}.

%
%
%

To show \eqref{inclusion-aux-h2} we first note that \eqref{reg-half}
yields
\begin{displaymath}
 \lbd\bigl(B(x,\s)\setminus\{h<\rho\}\bigr)\le
\lbd\bigl((\Rd\setminus \mr)\setminus\{h<\rho\}\bigl)
=\lbd\bigl(\{h\geq\rho\}\setminus\mr \bigr)
=0  \, .
\end{displaymath}
Analogously to  \eqref{inclusion-aux-h1-h} we then obtain 
\begin{align*}
  \int_{B(x,\s)}K_\d(x-y)  \,  h(y)\,\dld(y)
&= \int_{B(x,\s)\cap \{h< \rho\}}K_\d(x-y)  \,  h(y)\,\dld(y)\\
& < \rho  \int_{B(x,\s)\cap \{h< \rho\}}K_\d(x-y) \,\dld(y)\\ 
& = \rho  \int_{B(x,\s)}K_\d(x-y) \,\dld(y)\, ,
\end{align*}
where for the strict inequality we used our assumption that $K$ is strictly positive in a neighborhood of $0$.
Adapting the last estimate of the proof of \eqref{inclusion-aux-h1} we then find
\begin{align*}
 \hpd(x)
&<  \r \int_{B(x,\s)}K_\d(x-y)\,\dld(y) + \int_{\Rd\setminus B(x,\s)}K_\d(x-y)  \,  h(y)\,\dld(y)
\\&=  \r \int_{B(x,\s)}K_\d(x-y)\,\dld(y) + \r \int_{\Rd\setminus B(x,\s)}K_\d(x-y)\,\dld(y)
\\&\qquad - \r \int_{\Rd\setminus B(x,\s)}K_\d(x-y)\,\dld(y)
+ \int_{\Rd\setminus B(x,\s)}K_\d(x-y)  \,  h(y)\,\dld(y)
\\&\leq \r + \int_{\Rd\setminus B(x,\s)}K_\d(x-y)  \,  h(y)\,\dld(y)\, .
\end{align*}
Now, in the case of a bounded density $h$
the inequality \eqref{inclusion-aux-h2-bounded} follows from 
\eqref{inclusion-aux-conversion}, while in the general case
the estimate 
\begin{align*}
   \int_{\Rd\setminus B(x,\s)}K_\d(x-y)  \,  h(y)\,\dld(y)
\leq \sup_{y \in \Rd\setminus B(x,\s)}K_\d(x-y) 
= \d^{-d} \kti{\tfrac\s\d}
\end{align*}
leads to \eqref{inclusion-aux-h2}.
\end{proofof}

\begin{proofof}{Theorem \ref{include-main-thm-new}}
We begin by proving the first inclusion. To this end, we 
 fix an  $x\in  M\mdds_{\rho+\varepsilon+\eps}=\Rd\setminus
(\Rd\setminus M_{\r+\e+\eps})\pdds$. This means $x\notin
(\Rd\setminus M_{\r+\e+\eps})\pdds$, i.e., for all $x^\prime\in
\Rd\setminus M_{\r+\e+\e}$ we have $\snorm{x-x'}>2\s$. In other words, for all $x'\in \Rd$ with
$\snorm{x-x'}\le 2\s$, we have  $x^\prime\in M_{\rho+\varepsilon+\eps}$, i.e., we have found
\begin{equation}\label{include-main-thm-new-h1-key}
   B(x,2\s)\subset M_{\rho+\varepsilon+\eps}\, .
\end{equation}
Let us now suppose that there exists a sample $x_i\in D$ such that $\hdd(x_i)<\r$ and
$\snorm{x-x_i}\le\s$. By $\inorm{\hdd-\hpd}<\varepsilon$
we then find 
\begin{equation}\label{include-main-thm-new-h1}
\hpd(x_i) < \rho+\varepsilon\, .
\end{equation}
On the other hand, $\snorm{x-x_i}\le\s$ together with the already shown \eqref{include-main-thm-new-h1-key}
implies 
$B(x_i,\s)\subset M_{\rho+\varepsilon+\eps}$ by a simple application of the triangle inequality.
Consequently, \eqref{inclusion-aux-h1} 
together with $\eps \geq  \rho\kto{\tfrac\s\d}$
gives $\hpd(x_i) \geq \r+\varepsilon$, which contradicts
\eqref{include-main-thm-new-h1}.
%
%
 For all samples $x_i\in D$, we thus have 
$\hdd(x_i)\geq \r$ or $\snorm{x-x_i}>\s$.
Let us assume that we have $\snorm{x-x_i}>\s$ for all $x_i\in D$.
Then we find 
\begin{align*}
   \hdd(x) 
=\frac{1}{n}\sum^n_{i=1}  \d^{-d}K\Bigl(\frac{x-x_i}\d\Bigr) 
\leq \frac{1}{n}\sum^n_{i=1}  \d^{-d} \kti{\tfrac \s\d}
= \d^{-d} \kti{\tfrac \s\d} \leq \r
\, .
\end{align*}
%
%
%
%
On the other hand,  we have
$B(x,\s)\subset B(x,2\s)\subset M_{\rho+\varepsilon+\eps}$ and 
therefore  \eqref{inclusion-aux-h1} 
together with $\eps \geq  \rho\kto{\tfrac\s\d}$
gives $\hpd(x) \geq \r+\varepsilon$.
By $\inorm{\hdd-\hpd}<\varepsilon$ we conclude that 
%
%
%
$\hdd(x)>\rho$, and hence we have found a contradiction. Therefore
there does exist a sample $x_i\in D$ with $\snorm{x-x_i}\leq \s$.
Using the inclusion \eqref{include-main-thm-new-h1-key}  
together with the triangle inequality
we then again find $B(x_i,\delta)\subset M_{\rho+\varepsilon+\eps}$, and hence 
\eqref{inclusion-aux-h1} yields $\hpd(x_i) \geq \r+\varepsilon$. This leads to 
$\hdd(x_i)\geq\r$, and hence we 
finally obtain 
\begin{displaymath}
   x\in  \{x'\in D:\hdd(x') \geq \r\}\pds = L_{D,\r}\, .
\end{displaymath}
Finally, if $h$ has a bounded density, then we have $\mr=\emptyset$ for $\r> \inorm h$ 
and therefore $ M\mdds_{\r+\e+\eps} \subset L_{D,\r}$ is trivially satisfied. 
Moreover, to show the assertion for 
$\r\leq\inorm h$, we simply need to replace
\eqref{inclusion-aux-h1} with \eqref{inclusion-aux-h1-bounded} in the proof above.

Let us now prove the second inclusion. To this end, we pick an $x\in L_{D,\r}$. By the definition of $L_{D,\r}$, 
there then 
exists an $x_i\in D$ such that $\snorm{x-x_i}\le\s$ and 
$\hdd(x_i)\ge\rho$. 
The latter implies $\hpd(x_i)>\rho-\varepsilon$.

Our first goal is to show that $M_{\rho-\varepsilon-\eps}\cap B(x_i,\s)\ne\emptyset$.
To this end, let us assume the converse, that is 
$B(x_i,\s)\subset
\Rd\setminus M_{\r-\e-\eps}$. 
By \eqref{inclusion-aux-h2} 
and $\eps \geq \d^{-d} \kti{\tfrac\s\d}$
we then find $\hpd(x_i)<\rho-\varepsilon$, which 
%
%
%
contradicts the earlier established $\hpd(x_i) > \r-\e$. Consequently, there 
exists an  
$\tilde{x}\in M_{\rho-\varepsilon-\eps}\cap
B(x_i,\s)$, which in turn leads to 
\begin{displaymath}
   d(x,M_{\rho-\varepsilon-\eps})\le \snorm{x -\tilde{x}}\le
\snorm{x-x_i}+\snorm{x_i-\tilde{x}}\le 2\s\, .
\end{displaymath}
This shows the desired
$x\in M^{+2\s}_{\rho-\varepsilon-\eps}$.
Finally, to show the assertion for bounded densities, we simply need to replace
\eqref{inclusion-aux-h2} with \eqref{inclusion-aux-h2-bounded} in the proof above.
\end{proofof}



For the proof of Lemma \ref{indicator} we need to 
recall the following 
classical result, which is a reformulation of \cite[Theorem
2.6.4]{VaWe96}. 

\begin{theorem}\label{thm:finite-vc}
 Let $\eu A$ be a set of subsets of $Z$ that has finite VC-dimension $V$. Then the  
set of indicator functions $\eu G:= \{\eins_A: A\in \eu A \}$ is a uniformly bounded VC-class 
for which we have $B=1$ and the   constants $A$ and $\nu$ in \eqref{VC-ineq} only depend on $V$.
\end{theorem}

We also need the next result, which investigates the effect of scaling in the input space. 

\begin{lemma}\label{scal-inv-cov}
 Let $\eu G$ be set of measurable functions $g:\R^d\to \R$ such that 
there exists a constant $B\geq 0$ with  $\Vert g\Vert_\infty\le B$
for all $g\in\eu{G}$. For $\d>0$, we define 
$g_\d:\Rd\to \R$ by $g_\d(x) := g(x/\d)$, $x\in \Rd$
Furthermore, we write
$\eu G_\d := \{g_\d: g \in \eu G  \}$. Then, for all $\eps\in (0,B]$
and all $\d>0$, we have 
$$
\sup_P \Cal{N}(\eu{G},L_2(P),\eps) = \sup_P \Cal{N}(\eu{G_\d},L_2(P),\eps)\, ,
$$
where the suprema are taken over all  probability measures $P$ on $\Rd$.
\end{lemma}

\begin{proofof}{Lemma \ref{scal-inv-cov}}
Because of symmetry we only prove ``$\leq$''.
Let us fix $\eps,\d>0$ and a distribution $P$ on $\Rd$. We  define a new distribution $P'$ on $\Rd$ by
$P'(A) := P(\frac 1 \d A)$ for all measurable $A\subset \Rd$. 
Furthermore, let $\eu C'$ be an $\eps$-net of $\eu G_\d$ with respect to $L_2(P')$.
For  $\eu C:= \eu C'_{1/\d}$, we then have $|\eu C|= |\eu C'|$, and hence
it suffices to show that $\eu C$ is an $\eps$-net of $\eu G$ with respect to $L_2(P)$.
To this end, we fix a $g\in \eu G$. Then $g_\d\in \eu G_\d$, and hence there exists an $h'\in \eu C'$
with $\snorm {g_\d-h'}_{L_2(P')}\leq \eps$. Moreover, we have  $h:= h'_{1/\d}\in \eu C$, and since the definition
of $P'$ ensures $\E_{P'} f_\d = \E_{P} f$ for all measurable $f:\Rd\to [0,\infty)$, we obtain
\begin{align*}
 \snorm {g-h}_{L_2(P)} = \snorm {g_\d-h_\d}_{L_2(P')}  = \snorm {g_\d-h'}_{L_2(P')} \leq \eps,
\end{align*}
i.e.~$\eu C$ is an $\eps$-net of $\eu G$ with respect to $L_2(P)$.
\end{proofof}

\begin{proofof}{Lemma \ref{indicator}}
   The  collection $\eu A:= \{x + B_{\snorm\cdot}: x\in \Rd \}$ of 
        closed balls with radius 1 has finite VC-dimension by
 \cite[Corollary 4.2]{DeLu01} or \cite[Lemma 4.1]{DeLu01}, respectively. In both cases, Theorem \ref{thm:finite-vc}
        thus shows that 
   \begin{align*}
      \eu G : = \bigl\{K(x - \cdot): x\in \Rd\bigr\}  
   \end{align*}
     there are constants $A$ and $\nu$ only depending on the VC-dimension of $\eu A$ such that 
         \begin{displaymath}
        \ca N \bigl(\eu G, \Lx 2 P,  \snorm K_\infty \eps\bigr)  
        =\ca N \bigl( \snorm K_\infty^{-1} \eu G, \Lx 2 P, \eps\bigr) 
        \leq \biggl( \frac A\eps  \biggr)^\nu
         \end{displaymath}      
      for all $\eps\in(0, 1]$ and all distributions $P$ on $\Rd$.
      Our next step is to apply
        Lemma \ref{scal-inv-cov}. To this end, we first observe that 
                \begin{displaymath}
                   \eu G_\d  
                        = \bigl\{ K(x - \d^{-1}\cdot\,) : x\in \Rd \bigr\} 
                        = \biggl\{ K\Bigl(\frac{x' - \cdot}\d\Bigr) : x'\in \Rd \biggr\}
                        = \d^d \eu K_\d\, .
                \end{displaymath}
                Consequently, Lemma \ref{scal-inv-cov} leads to 
                \begin{align*}
                 \sup_P \Cal{N}( \eu K_\d,L_2(P),\d^{-d} \eps)
                 &=  \sup_P \Cal{N}(\d^d \eu K_\d,L_2(P),\eps)\\
                &= \sup_P \Cal{N}(\eu{G},L_2(P),\eps)
                \leq \biggl( \frac {  A  \inorm K}\eps  \biggr)^\nu
                \end{align*}
              for all $\eps\in (0,  \inorm K]$.
      A simple variable transformation then yields the assertion.
\end{proofof}

\begin{proofof}{Lemma \ref{lip-ex}}
   We first observe that $A\subset E$ is a compact subset of some Banach space $E$ and 
  $T:A\to F$ is a $\a$-H\"older continuous map into another  Banach space $F$ with Lipschitz norm 
 then, for all $\eps>0$,  we have 
   \begin{displaymath}
      \ca N\bigl(T(A) , \snorm \cdot _F, |T|_\a\eps^\a\bigr) \leq \ca N\bigl(A,\snorm \cdot _E,  \eps\bigr)\, ,
   \end{displaymath}
        where $|T|_\a$ denotes the $\a$-H\"older constant of $T$. In the following we fix a 
        $\d>0$ with 
$\d\leq \bigl(\frac{|K|_\a}{\inorm K}\bigr)^{1/\a} \diam_\snorme (X)$ and 
        a probability measure $P$ on $\Rd$.
        For $x\in X$ we now consider the map $\kxd:\Rd\to [0,\infty]$ defined by 
        \begin{displaymath}
           \kxd (y) := K_\d(x-y) = \d^{-d} K\Bigl( \frac{x-y}\d \Bigr) \, , \qquad \qquad y\in \Rd.
        \end{displaymath}
        Since $K$ is bounded and measurable, so is $\kxd$, and hence we obtain a map 
$T:X\to L_\infty(P)$  defined by $T(x) := \kxd$.
 Our next goal is to show that $T$ is $\a$-H\"older continuous.
     To this end, we pick $x,x'\in X$. A simple estimate then yields
     \begin{align*}
        \inorm{T(x) - T(x')} 
        &= 
        \sup_{y\in \Rd} \biggl|  \d^{-d} K\Bigl( \frac{x-y}\d \Bigr) -  \d^{-d} K\Bigl( \frac{x'-y}\d \Bigr) \biggr|\\
        &\leq 
         \d^{-(\a+d)}   |K|_\a \, \snorm{x-x'}^\a\, ,
     \end{align*}
    i.e.~$T$ is indeed $\a$-H\"older  continuous with $|T|_\a \leq \d^{-(\a+d)}   |K|_\a $. By our initial observation
        and \eqref{cover-x}
 we then conclude that 
    \begin{align*}
       \ca N\bigl(\eu K_\d, \snorm \cdot_{\Lx 2 P} ,|T|_\a\eps^\a\bigr)
       &= 
       \ca N\bigl(T(X), \snorm \cdot_{\Lx 2 P} ,|T|_\a\eps^\a\bigr)\\
       &\leq \ca N\bigl(X, \inorm \cdot ,\eps \bigr)
                         \leq C_\snorme (X) \eps^{-d}
    \end{align*}
                        for all $0<\eps\leq \diam_\snorme (X)$. A simple variable transformation together with our bound on $|T|_\a$ thus 
                        yields
                        \begin{displaymath}
                           \ca N\bigl(\eu K_\d, \snorm \cdot_{\Lx 2 P} ,\eps\bigr) 
                        \leq C_\snorme (X) \biggl( \frac{|T|_\a}{\eps}  \biggr)^{d/\a}
                        \leq  C_\snorme (X) \biggl(  \frac{|K|_\a}{\d^{\a+d}\eps}   \biggr)^{d/\a}
                        \end{displaymath}
                        for all $0<\eps \leq  \d^{-(\a+d)}   |K|_\a \diam_\snorme (X)$.
                Since the assumed $$\d\leq \bigl(\frac{|K|_\a}{\inorm K}\bigr)^{1/\a} \diam_\snorme (X)$$ implies 
                \begin{displaymath}
                   \d^{-d} \inorm K \leq \bigl(\diam_\snorme (X)\bigr)^\a \d^{-(\a+d)} |K|_\a
                \end{displaymath}
                we then see that \eqref{lip-ex-h1} does hold for all $0<\eps \leq \d^{-d} \inorm K$.
\end{proofof}

For the proof of Theorem \ref{approximate-thm} we quote a version of Talagrand's inequality due to
\cite{Bousquet02} from \cite[Theorem 7.5]{StCh08}.

\begin{theorem}\label{Thm:talagrand}
Let $(Z,P)$ be a probability space and $\eu{G}$ be a set of measurable
functions from $Z$ to $\R$. Furthermore, let $B\ge 0$ and $\sigma\ge 0$ be
constants such that $\E_Pg=0$, $\E_P g^2\le \sigma^2$, and $\Vert
g\Vert_\infty\le B$ for all $g\in\eu{G}$. For $n\ge 1$, define
$G:Z^n\rightarrow\R$ by
$$G(z):=\sup_{g\in\eu{G}}\left|\frac{1}{n}\sum^n_{j=1}
g(z_j)\right|,\,\,\, z=(z_1,\ldots,z_n)\in Z^n.$$
Then, for all $\vs>0$, we have
\begin{displaymath}
P^n\Big(\!\Big\{z\!\in\! Z^n\,:\,G(z)\ge
4\E_{P^n}G+\sqrt{\frac{2\vs\sigma^2}{n}} +\frac{\vs
B}{n}\Big\}\!\Big) \le e^{-\vs}.
\end{displaymath}
\end{theorem}

For the proof of Theorem~\ref{approximate-thm} we also need 
\cite[Proposition 2.1]{GiGu01a}, which
 bounds   the
expected suprema of empirical processes indexed by uniformly bounded VC-classes. The following theorem provides a 
slightly simplified version of that proposition.

\begin{theorem}\label{Thm:gigui}
Let $(Z,P)$ be a probability space and $\eu{G}$ be a 
uniformly bounded VC-class on $Z$ with constants $A$, $B$, and $\nu$. Furthermore, let 
$\s>0$ be a constant such that $\sigma\leq B$ and 
$\E_P g^2\le \sigma^2$ for all $g\in \eu G$.
Then there exists a universal constant $C$ 
such that $G$ defined as in Theorem \ref{Thm:talagrand} satisfies 
\begin{equation}\label{Eq:expectation}
\E_{P^n}G\le
C\left(\frac{\nu
B}{n}\log\frac{AB}{\sigma}+\sqrt{\frac{\nu\sigma^2}{n}\log\frac{ AB}{\sigma}}
\right).\end{equation}
\end{theorem}

We are now able to establish the following 
generalization of Theorem~\ref{approximate-thm}.

\begin{proposition}\label{approximate-prop} 
 Let $X\subset \Rd$ and 
 $P$ be a probability measure on 
 $X$ that has a Lebesgue density $h\in \Lx 1 {\Rd} \cap \Lx p{\Rd}$ 
 for some $p\in (1,\infty]$. Moreover, let $\frac 1 p + \frac 1 {p'} = 1$ and $q:= \frac 1 {2p'} = \frac 1 2 - \frac 1 {2p}$
 and  $K:\Rd\to [0,\infty)$ be a symmetric kernel.
Suppose further that the set $\eu K_\d$  defined in \eqref{def-Kd-class}
satisfies 
\eqref{target-cover-bound} for all $\d\in (0,\d_0]$, where $\d_0\in(0,1]$.
Then, there
exists  a positive constant $C$ only depending on
$d$, $p$, and $K$  
such that, for all
$n\geq 1$, all $\d\in(0,\d_0]$ satisfying $\d \snorm h_p^{p'}\leq  4^{p'}{\inorm K}{} $, and all $\vs\geq 1$ we have
\begin{displaymath}
   P^n\bigg(\bigg\{D\,:\, \snorm{\hdd-\hpd}_{\ell_\infty(X)}
< \frac {C \vs}{n\d^d} \log\frac{C }{\d^{a + dq} \snorm h_p^{1/2}}  
  +\sqrt{\frac{ C \snorm h_p  \vs}{\d^{d(1+1/p)}n}\log\frac{C}{\d^{a + dq} \snorm h_p^{1/2}}   }   
\bigg\}
\bigg)\ge 1-e^{-\vs}.
\end{displaymath}
\end{proposition}

\begin{proofof}{Proposition \ref{approximate-prop}}
We define $\theta := 1 - \frac 1{2p'} = \frac 12 + \frac 1{2p}$.
Then $K\in \Lx 1 \Rd \cap \Lx \infty \Rd$ leads to
\begin{displaymath}
 \snorm K_{2p'} \leq \snorm K_1^{1-\theta} \snorm K_\infty^\theta = \snorm K_\infty^\theta \, .
\end{displaymath}
%
%
We further define 
\begin{displaymath}
   \kxd:=\d^{-d}K\left(\frac{x-\cdot}{\d}\right)
\end{displaymath}
and
$\fxd:=\kxd-\E_P \kxd$.
Then it
is easy to check that $\E_P \fxd=0$ and $\Vert \fxd\Vert_\infty\le 2\inorm K \d^{-d}$
for all $x\in X$ and $\d>0$.
Moreover, we have $\E_Pf^2_{x,\d} {}\leq \E_P
k^2_{x,\d}$ and thus 
\begin{align*}
\E_Pf^2_{x,\d} 
&=
\d^{-2d}\int_{\R^d}K^2\left(\frac{x-y}{\d}\right)h(y)\dld(y)\\
&\leq \d^{-2d} \snorm h_p \biggl(\int_{\R^d}K^{2p'}\left(\frac{x-y}{\d}\right) \dld(y)  \biggr)^{1/p'}\\
&= \d^{-2d} \snorm h_p \biggl(  \d^d \int_{\R^d}K^{2p'} ( {x-y}) \dld(y)  \biggr)^{1/p'}\\
& \leq \d^{-d(1+1/p)} \snorm h_p \inorm K^{2\theta}\\
& =:  \s_\d^2
\end{align*}
for all $x\in X$ and $\d>0$.
In addition, for all $D\in X^n$ we have 
\begin{align}\label{approximate-prop-h1}
 \E_D \fxd =  \frac{1}{n}\sum^n_{i=1}\fxd(x_i)=\hdd(x)-\hpd(x)\, . 
\end{align}
Applying Theorem~\ref{Thm:talagrand} to $\eu{G}:=\{\fxd: x\in X\}$, we hence obtain, 
for all $\d>0$, $\vs>0$, and $n\geq 1$, that
\begin{align}\label{Eq:bound}
 \snorm{\hdd-\hpd}_{\ell_\infty(X)}  < 4\E_{D'\sim P^n} \snorm{\hdpd-\hpd}_{\ell_\infty(X)}  +\frac{2\vs}{n\d^d}
   + \sqrt{\frac{2\vs \snorm h_p \inorm K^{2\theta}}{n\d^{d(1+1/p)}}}
\end{align}
holds with probability $P^n$ not smaller than $1-e^{-\vs}$. It thus remains to bound the term 
\begin{displaymath}
  \E_{D'\sim P^n} \snorm{\hdpd-\hpd}_{\ell_\infty(X)}  =  \E_{D'\sim P^n} \sup_{x\in X}\bigl|\E_D \fxd\bigr| \, .
\end{displaymath}
To this end, we first note that 
$|\E_P \kxd| \leq \inorm\kxd = \d^{-d}\inorm K =: B_\d$. Consequently, we have 
\begin{displaymath}
 \eu F_\d := \{ \fxd: x\in X\} \subset  \bigl\{\kxd-b:\kxd\in\eu{K}_\d, |b| \leq   B_\d\bigr\}\, ,
\end{displaymath}
and since $\ca N([-B_\d,B_\d], |\cdot|,\eps) \leq 2B_\d\eps^{-1}$ we conclude that for $\tilde A := \max\{1,A_0\}$ we have 
\begin{align*}
 \sup_{Q}\Cal{N}\bigl({\eu{F}_\d},L_2(Q),\eps\bigr)
 &\le
 2 \biggl( \frac{A_0 \inorm K \d^{-(d+a)} }{\eps}   \biggr)^\nu  \cdot \frac{\inorm K \d^{-d}}\eps\\     
 &\leq \biggl( \frac{2\tilde A \inorm K \d^{-(d+a)} }{\eps}   \biggr)^{\nu+1}
\end{align*}
for all $\d\in (0,\d_0]$ and all $\eps\in( 0,B_\d]$, where the supremum runs over all
distributions $Q$ on $X$.
Now, our very first estimates showed $\inorm \fxd \leq 2B_\d$ and 
$\E_P\fxd^2 \leq \s_\d^2$ and since 
$\s_\d \leq 2B_\d$ is equivalent to 
\begin{displaymath}
 \d \leq 4^{p'}\frac{\inorm K^{p'(2-2\theta)}}{\snorm h_p^{p'}} =  4^{p'}\frac{\inorm K}{\snorm h_p^{p'}} 
\end{displaymath}
Theorem \ref{Thm:gigui} together with $2\theta = 1+1/p$ thus yields
\begin{align*}
 \E_{D'\sim P^n} \sup_{x\in X}\bigl|\E_D \fxd\bigr|
\le 
 C\Bigg(\frac{2(\nu\!+\!1)\inorm K }{n\d^{d}}\log\frac{2\tilde A \inorm K }{\s_\d \d^{d+a}}  
+\sqrt{\frac{(\nu\!+\!1) \snorm h_p \inorm K^{1+1/p}}{2\d^{d(1+1/p)}n}\log\frac{2\tilde A \inorm K}{\s_\d \d^{d+a}} }\,\Bigg).\nonumber
\end{align*}
for such $\d$. Moreover, we have 
\begin{displaymath}
 \log\frac{2\tilde A \inorm K }{\s_\d \d^{d+a}}  
 = 
 \log\frac{2\tilde A \inorm K }{\d^{-d(1+1/p)/2} \snorm h_p^{1/2} \inorm K^{\theta} \d^{d+a}}  
 = 
 \log\frac{2\tilde A \inorm K^{q} }{\d^{a + dq} \snorm h_p^{1/2}}  \, ,
\end{displaymath}
and hence the previous estimate can be simplified to 
\begin{align*}
 \E_{D'\sim P^n} \sup_{x\in X}\bigl|\E_D \fxd\bigr|
\le 
 C\Bigg(\frac{4\nu\inorm K }{n\d^{d}}\log\frac{2\tilde A \inorm K^{q} }{\d^{a + dq} \snorm h_p^{1/2}}  
+\sqrt{\frac{\nu \snorm h_p \inorm K^{1+1/p}}{\d^{d(1+1/p)}n}\log\frac{2\tilde A \inorm K^{q} }{\d^{a + dq} \snorm h_p^{1/2}}   }\,\Bigg).\nonumber
\end{align*}
Combining this with \eqref{Eq:bound} gives 
\begin{align*}
  \snorm{\hdd-\hpd}_{\ell_\infty(X)}  
  &< 4\E_{D'\sim P^n} \snorm{\hdpd-\hpd}_{\ell_\infty(X)}  +\frac{2\vs}{n\d^d}
   + \sqrt{\frac{2\vs \snorm h_p \inorm K^{1 + 1/p}}{n\d^{d(1+1/p)}}} \\
   & \leq 
  4 C\Bigg(\frac{4\nu\inorm K }{n\d^{d}}\log\frac{2\tilde A \inorm K^{q} }{\d^{a + dq} \snorm h_p^{1/2}}  
+\sqrt{\frac{\nu \snorm h_p \inorm K^{1+1/p}}{\d^{d(1+1/p)}n}\log\frac{2\tilde A \inorm K^{q} }{\d^{a + dq} \snorm h_p^{1/2}}   }\,\Bigg) \\
  &\qquad +\frac{2\vs}{n\d^d} + \sqrt{\frac{2\vs \snorm h_p \inorm K^{1 + 1/p}}{n\d^{d(1+1/p)}}}  \\
  & \leq \frac {\tilde C \vs}{n\d^d} \log\frac{\tilde C }{\d^{a + dq} \snorm h_p^{1/2}}  
  +\sqrt{\frac{ \tilde C \snorm h_p  \vs}{\d^{d(1+1/p)}n}\log\frac{\tilde C}{\d^{a + dq} \snorm h_p^{1/2}}   }
\end{align*}
with probability $P^n$ not smaller than $1-e^{-\vs}$.
\end{proofof}

\begin{proofof}{Theorem \ref{approximate-thm}}
By Proposition \ref{approximate-prop}, it suffices to find a constant $C'$ such that 
\begin{align}\label{approximate-thm-h1}
 \frac {C \vs}{n\d^d} \log\frac{C }{\d^{a + dq} \snorm h_p^{1/2}}  
  +\sqrt{\frac{ C \snorm h_p  \vs}{\d^{d(1+1/p)}n}\log\frac{C}{\d^{a + dq} \snorm h_p^{1/2}}   }   
   \leq C' \sqrt{\frac{\snorm h_p \,|\!\log \d| \, \vs }{n\d^{d(1+1/p)}}}\, .
\end{align}
To this end, we first observe that $ \d^{a + dq}\leq  \frac{\snorm h_p^{1/2}}  C$ implies 
\begin{displaymath}
   \frac C {\d^{a + dq} \snorm h_p^{1/2}} \leq \d^{-2a - 2dq}\, ,
\end{displaymath}
and thus we obtain $\log\frac C {\d^{a + dq} \snorm h_p^{1/2}} \leq (2a+2dq) \log \d^{-1}$. For $C'':= (2a+2dq)C$ we therefore find
\begin{eqnarray*}
 \frac {C \vs}{n\d^d} \log\frac{C }{\d^{a + dq} \snorm h_p^{1/2}}  
  +\sqrt{\frac{ C \snorm h_p  \vs}{\d^{d(1+1/p)}n}\log\frac{C}{\d^{a + dq} \snorm h_p^{1/2}}   }   
\Leq
   \frac{C'' \vs }{n\d^d} \log \d^{-1}
+\sqrt{\frac{C''  \snorm h_p\vs}{n\d^{d(1+1/p)}}\log \d^{-1}}\, .
\end{eqnarray*}
Moreover, it is easy to check that the assumption 
$$
\frac{|\!\log \d |}{n\d^{d/p'}}  \leq \frac{\snorm h_p}{C'' \vs}
$$
ensures that 
$$
\frac{C'' \vs }{n\d^d} \log \d^{-1} \leq \sqrt{\frac{C''  \snorm h_p\vs}{n\d^{d(1+1/p)}}\log \d^{-1}}\, ,
$$
and from the latter we conclude that \eqref{approximate-thm-h1} holds for $C' := 2 \sqrt{C''} $. 
The assertion now follows for the constant $C''' := \max\{C, C', C''\}$.
\end{proofof}

\subsection{Proofs for the KDE-Based Clustering in  Section \ref{sec:algo}}

\begin{lemma}\label{delta-lemma}
   For all $\d \in (0,\eul^{-1}]$ and $d\geq 1$ we have 
$    \d^{|\log \d|} |\log \d|^{2d-2} \leq  \d^{|\log \d|-d}$.
\end{lemma}

\begin{proofof}{Lemma \ref{delta-lemma}}
   The derivative of the function $h(\d) := \d^{-1/2} + \log \d$ is given by 
\begin{displaymath}
   h'(\d) = \frac 1 2 \cdot \frac{2\sqrt\d-1}{\d^{3/2}}
\end{displaymath}
and from this we conclude that $h$ has a global minimum at $\d=1/4$. Since
$h(1/4) = 2 - 2\log 2>0$, we thus find 
$     |\log \d| = -\log \d< \d^{-1/2}$ for all $\d\in (0,1]$. The latter yields
$|\log \d|^{2d} < \d^{-d}$, and since $|\log \d| \geq 1$ for $\d \in (0,\eul^{-1}]$, 
we then obtain the assertion.
\end{proofof}

\begin{proofof}{Theorem \ref{analysis-main2-new}}
Let us fix a $D\in X^n$   with  $\inorm {\hdd-\hpd  } < \e/2$.
By \eqref{simplified-conc} we see that the probability $P^n$ of such a $D$
is not smaller than $1-e^{-\vs}$. 
We define $\eps: = \inorm h \kto{\tfrac\s\d}$. In the case of $\supp K\subset B_{\snorm\cdot}$
this leads to $\eps = 0$ 
$\d^{-d} \kti{\tfrac \s\d} = 0 \leq \r_0$
as noted after Theorem \ref{include-main-thm-new}.
Furthermore, 
in the case of \eqref{lem:tail-functions-h1}, Lemma \ref{lem:tail-functions} shows
\begin{align*}
 \eps 
 = \inorm h \kto{\tfrac\s\d}
 \leq  \inorm h \kto{|\log \d|^{2}} 
 &\leq c d^2 \vold  e^{-|\log \d|^{2}} |\log \d|^{2d-2}\\
 & \leq c d^2 \vold   \d^{|\log \d|-d} \leq \e/2\, ,
\end{align*}
where in the third to last step we used $0<\d\leq 1$ and in the second to last step 
we used Lemma \ref{delta-lemma}.
In addition, we have 
\begin{displaymath}
 \d^{-d} \kti{\tfrac \s\d}  
 \leq 
  c \d^{-d}  e^{-|\log \d|^{2}}
  = 
  c \d^{|\log \d|-d}
  \leq \e\leq \r_0\, .
\end{displaymath}
Consequently, Theorem \ref{include-main-thm-new} shows, for all  $\r\geq \r_0$, that 
\begin{equation}\label{key-uq}
   M_{\r+\e}\mdds\subset L_{D,\r} \subset  M_{\r-\e}\pdds\, .
\end{equation}

\ada i
The assertion follows from Theorem \ref{main-generic-single} applied in the case $\tilde d := 2\s$.
Indeed, we have just seen that 
\eqref{generic-inclus} holds for all $\r\geq \r_0$, if we replace $\d$ by $\tilde \d$, and our 
assumptions guarantee 
$\tilde \d\in (0,  \dthick]$, $\r_0\geq \rls$, 
and 
\begin{displaymath}
\t> \psi(\tilde \d) = 3 \cthick \tilde\d^\g > 2 \cthick \tilde \d^\g \, .
\end{displaymath}
Moreover, \eqref{single_cluster-symdif} is a simple consequence of \eqref{key-uq}.

\ada {ii}
Let us check that the remaining assumptions of Theorem \ref{analysis-main-combined-new}
 are also satisfied for $\tilde \d:= 2\s$, if $\e^* \leq (\rss-\rs)/9$.
Clearly, we have $\tilde \d\in (0,  \dthick]$, $\e\in (0 , \e^*]$, and $\psi(\tilde \d)< \t$. To show  
$\t\leq \ts(\e^*)$ we write 
$$
E:= \{\e'\in  (0,\rss-\rs] : \t^*(\e') \geq \t \}.
$$
Since we assumed $\e^*< \infty$, we obtain
$E\neq \emptyset$ by the definition of $\e^*$. 
There thus exists an $\e'\in E$ with $\e' \leq \inf E + \e \leq \e^*$. Using the monotonicity of $\ts$ established in 
\cite[Theorem A.4.2]{Steinwart15a} we then conclude that $\t\leq \ts(\e')\leq \ts(\e^*)$, and hence all
assumptions of \cite[Theorem 2.9]{Steinwart15a}  are indeed satisfied with $\d$ replaced by  $\tilde \d$. The assertions now immediately follow from this theorem.
\end{proofof}

\begin{proofof}{Corollary \ref{cor:consis}}
   Using Theorem \ref{analysis-main2-new} the proof of \emph{ii)}
  is a literal copy of the proof of \cite[Theorem 4.1]{Steinwart15a} and the proof of \emph{i)} is an easy adaptation
of this proof.
\end{proofof}

\begin{proofof}{Corollary \ref{rates-cor1}}
   Using Theorem \ref{analysis-main2-new} the proof is a simple combination and adaptation of the proofs of \cite[Theorem 4.3]{Steinwart15a}
   and \cite[Corollary 4.4]{Steinwart15a}.
\end{proofof}

\begin{proofof}{Corollary \ref{rates-cor2}}
   Using Theorem \ref{analysis-main2-new} the proof is a simple combination and adaptation of the proofs of \cite[Theorem 4.7]{Steinwart15a}
   and \cite[Corollary 4.8]{Steinwart15a}.
\end{proofof}

\begin{proofof}{Theorem \ref{adaptive-level}}
   The definition of $\e_{\d,n}$ in \eqref{eps-adap} together with 
$4 C_u^2\log\log n\geq C\inorm h$
 ensures that \eqref{analysis-main2-h0} and \eqref{analysis-main2-h0-add}
are satisfied for all $\d\in \D$. In addition, the assumptions of Theorem \ref{adaptive-level}
further ensure that the remaining conditions of Theorem \ref{analysis-main2-new}
are also satisfied. Now the assertion follows by some standard union bound arguments, which are analogous 
to those of the proof of \cite[Theorem 5.1]{Steinwart15a}.
\end{proofof}

\subsection*{Acknowledgment}
The work of I.~Steinwart and P.~Thomann for this article was funded by the DFG Grant STE 1074/2-1.

\bibliographystyle{plain}
\bibliography{steinwart-mine,steinwart-books,steinwart-proc,steinwart-article}

\newpage

\renewcommand\thesection{\Alph{section}}
\setcounter{section}{0}
\setcounter{subsection}{0}
\setcounter{theorem}{0}

\section{Appendix: A Detailed Comparison to \cite{ChDaKpLu14a}}\label{sec:supplement}

In this supplement we present a detailed comparison of our results to the paper 
\cite{ChDaKpLu14a}. In particular, we discuss the different assumptions as well as 
the obtained statistical guarantees.

The rest of this supplement is organized as follows: In Section \ref{sec:common_ground}
we provide a common ground for comparing our results with those of 
\cite{ChDaKpLu14a}.
The comparison of these results is then presented in Section \ref{sec:comparison_to_CDKL},
and all required proofs can be found in Section \ref{sec:comparison-proofs}.

\subsection{A Common Ground}\label{sec:common_ground}

Let us begin by establishing a common ground for the comparison. 
Here, we note that \cite{ChDaKpLu14a}  directly works on 
density level sets $\mhx \r$ for a fixed density $h$, whereas we 
consider the sets $\mr$ that are independent on the specific choice of $h$. 
To treat both cases simultaneously, we need to assume that we have fixed a density $h$
such that $\mhx\r = \mr$ for the \emph{considered} $\r>0$. Note that by
\eqref{Mr-diff-closure} this is satisfied if $\mhx \r$ is \emph{regular closed},
i.e.~the interior of $\mhx\r$ is closed in $\mhx\r$. Informally speaking, this 
excludes 
 low-dimensional features such as long bridges or needles.
 Moreover, note that we can typically not guarantee $\mhx\r = \mr$ for all $\r>0$, even if 
 $h$ is continuous or even $C^\infty$. Indeed, if $h$ has a strict local maximum at, say 
 $x_0\in  \Rd$, and we consider $\r:= h(x_0)$, then we have $x_0\in \mhx\r$ but $x_0\not\in \mr$. 
 Consequently, such levels $\r$ need to be excluded in our considerations. To this end, we introduce 
 the following notion of cluster peaks.

\begin{definition}
 Let $h:\Rd\to [0,\infty)$ be continuous with compact support. Then we call an $A\subset \Rd$
 a cluster peak of $h$, if there exists a $\rdcp>0$ such that 
 $A\in \cc{\mhx\rdcp}$ and $A\subset \{h=\rdcp\}$.  Moreover, we denote the set of all
 cluster peaks of $h$ by $\setcp$.
\end{definition}

Informally speaking, cluster peaks are connected sets at which local maxima occur, but in general 
these maxima are not strict. In addition, not every point at which a local non-strict maximum occurs needs to be 
part of a cluster peak as saddle points on a flat ridge demonstrate.
Intuitively, every connected component of some
level $\r$ should have at least one  cluster peak, and 
Lemma \ref{lemma:existence-of-cluster-peaks} confirms this intuition, at least for continuous densities 
with compact support. For such $h$, it further shows that 
$|\cc{\mhx\r}|\leq |\setcp|$ for all $\r>0$.

Besides the level sets for $\r>0$ we sometimes also need to 
consider the support $\overline{\{h>0\}}$ of a continuous density $h$ and as a
precautious note  we remark that while \eqref{Mr-diff-closure} ensures 
$\overline{\{h>0\}} \subset \supp P = \mx 0$, equality, i.e.~$\overline{\{h>0\}} = \supp P$
does in general not hold, even if the density is $C^\infty$.

Another difference between \cite{ChDaKpLu14a} and our work is that 
%
\cite{ChDaKpLu14a}  considers \emph{path connected} components instead of connected components.
For this reason, let us recall that an $A\subset X$ of a metric space $(X,d)$ is called 
path connected, if for all $x_0,x_1\in A$, there is a \emph{path $\g$ connecting $x_0$ and $x_1$ in $A$}, 
that is, there exists a 
continuous 
$\g:[0,1]\to A$ with $\g(0)=x_0$ and $\g(1) = x_1$. Obviously, path-connectivity 
defines an equivalence relation and the corresponding equivalence classes 
are called path connected components. In the following, we denote the set of path connected components 
of a set $A$ by $\ccp A$. Again, if $A\subset B$ are subsets of a metric space, then 
$\ccp A \comparable \ccp B$, see Lemma \ref{lem:path-comparable}.
Moreover, it is well-known that path connected sets are connected, and from this we 
almost immediately obtain $\ccp A \comparable \cc A$, see Lemmas \ref{lem:path-implies-connect} and 
\ref{lem:path-vers-connect-compos} for details.
Finally, in general we do \emph{not} have $\ccp A = \cc A$, but 
for \emph{open}, non-empty 
$A\subset \Rd$ we actually have 
$\ccp A = \cc A$, see Lemma \ref{lem:open-path-persists}.

With these preparations we can now formulate some assumptions on $P$ made regularly in this section.  

\begin{assumption}{C1}
 We have a Lebesgue absolutely continuous probability measure $P$ on $\Rd$ that has 
 a continuous Lebesgue density $h:\Rd\to [0,\infty)$ with compact support $X:= \overline{\{h>0\}}$. 
 In addition, $h$ has finitely many cluster peaks and $\cc{\mhx\r} = \ccp{\mhx\r}$
 for all $\r\in (0,\inorm h]$.
\end{assumption}

\begin{assumption}{C2}
 We have a probability measure $P$ on $\Rd$ that satisfies \assx  {C1}
 and for all $\r\in (0,\inorm h)$ that are not the height of a cluster peak the 
 level set $\mhx\r$ is regular closed.
\end{assumption}

Our analysis focuses on the identification and estimation of splits in the population cluster tree.
Similarly, \cite{WaLuRi19a} provides results for estimating such split levels, and later we will see that 
the results of \cite{ChDaKpLu14a} can be re-interpreted in such a way, too. For this reason, we 
now formally introduce the notion of split levels.

\begin{definition}\label{def:split-levels}
  Let $h:\Rd\to [0,\infty)$ be continuous with compact support. Then a $\rs>0$ is called a split level of $h$,
  if there exist a $\rss>\rs$ and $A_1,A_2\in \cc{\mhx \rss}$ such that for all $0<\r\leq \rss$ the 
  corresponding CRM $\z:\cc{\mhx\rss}\to \cc{\mhx\r}$ satisfies the following implications:
\begin{enumerate}
 \item If we have $\z(A_1) = \z(A_2)$, then $\r\leq \rs$.
 \item If we have $\z(A_1) \neq \z(A_2)$, then $\r\geq \rs$.
\end{enumerate}
In this case, we call $\rss$ a   witness level for $\rs$ and $(A_1,A_2)$ a pair of witness components for $\rs$ at 
level $\rss$.
Moreover, we say that $\rs$ is a simple split level, if for all witnessing levels $\rss$  for $\rs$ there is only 
one pair of witness components for $\rs$ at level $\rss$.
Finally, we denote the set of all split levels of $h$ by $\setsl$.
\end{definition}

Analogously to \cite{WaLuRi19a}, Definition \ref{def:split-levels} only considers 
 split levels $\rs>0$, but we would like to remark that if $X$ is not connected, it would
make perfect sense to view $\rs:= 0$ as a split level, too. Also, it is not surprising that
 if $h$ has finitely many cluster peaks, then it automatically has 
at most finitely many 
split levels, see  Lemma \ref{lemma:existence-of-cluster-peaks}. In addition,  note that if \assx{C2} is satisfied, $X$ is connected,  
and all split levels are simple and not the height of a cluster peak,
 then $P$
has a finite split tree. Finally, for later use we emphasize that pairs of witnesses enjoy a  certain persistence:
If $(A_1,A_2)$ is a pair of witness components for a split level $\rs$ at 
level $\rss$ and $\z:\cc{\mhx\rss}\to \cc{\mhx\r}$  is the CRM for some $\r\in (\rs,\rss)$, 
then $(\z(A_1), \z(A_2))$ is a pair of witness components for $\rs$ at level $\r$.
%
%

Finally, recall that for a 
continuous function $h:\Rd\to \R$, the uniform modulus of continuity is defined 
by
\begin{align*}
 \om(h,\d) := \sup_{\snorm{x-x'}\leq \d}|h(x) - h(x')|\, , \qquad \qquad \d>0.
\end{align*}
In general, we may have  $\om(h,\d) = \infty$. If, however, $h$ is uniformly continuous, then 
we have $\om(h,\d) <\infty$ for all $\d>0$ and $\lim_{\d\to 0}\om(h,\d) =0$. Moreover, for $\theta$-H\"older
continuous $h$ with $\theta$-H\"older constant $|h|_\theta$,  we obviously have $\om(h,\d) \leq |h|_\theta \,\d^\theta$
for all $\d >0$.

\subsection{Comparison to \cite{ChDaKpLu14a}}\label{sec:comparison_to_CDKL}

The goal of this section is to provide an in-depth comparison of our statistical guarantees 
to the ones obtained in 
\cite{ChDaKpLu14a}. To this end, we first recall the notion of separated sets used in \cite{ChDaKpLu14a},
as this notion is central for their analysis. We then present the central result of \cite{ChDaKpLu14a},
which in the remaining part of this section is then analyzed and compared in detail.

\begin{definition}\label{def:separation}
 Let $P$ be a distribution on $\Rd$ that has a continuous Lebesgue density $h$, and let $X:= \overline {\{h>0\}}$, 
 as well as 
  $\d>0$, $\e \in (0,1)$, and $\eps \geq 0$.
  Then two non-empty 
 $A_1,A_2\subset X$ are called $(\d,\e,\eps)$-separated, if there exists a set
 $S\subset X$ such that the following two conditions are satisfied:
\begin{enumerate}
 \item The set \emph{$S$ separates $A_1$ and $A_2$}, that is, 
  for all $x_1\in A_1$ and $x_2\in A_2$ and every continuous $\g:[0,1]\to X$ with $\g(0) = x_1$ and 
 $\g(1) = x_2$ there exists a $t\in [0,1]$ such that $\g(t) \in  S$.
 \item There is a \emph{gap in the density between the sets $S\pde$ and $A_1\pde \cup A_2\pde$} in the sense of 
 \begin{align}\label{sep-gap}
  0 \vee \sup_{x\in S\pde} h(x) < (1-\e) 
  \hdr \d  {A_1\cup A_2}
  - \eps\, ,
 \end{align}
 where the \emph{twin peaks function} $\hdr\mycdot  {A_1\cup A_2}:[0,\infty)\to [0,\infty)$ is defined by 
\begin{align*}
 \hdr {\d'}  {A_1\cup A_2} :=\inf_{x\in (A_1\cup A_2)\pdex{\d'}}h(x)\, , \qquad \qquad \d'\geq 0.
\end{align*}
\end{enumerate}
\end{definition}

Note that the original definition does \emph{not} demand the continuity of $h$. However, 
both the consistency result, see the discussion following \cite[Theorem III.4]{ChDaKpLu14a},
and the obtained rates,  see the discussion following \cite[Theorem VII.5]{ChDaKpLu14a},
require a continuous density, and hence we included this assumption already to the 
definition.
Furthermore,  we also like to mention  that 
\cite[Section III]{ChDaKpLu14a} only considers the case $\eps=0$,  
while later in their Theorem VII.5 the authors also need sets that are 
$(\d,\e,\eps)$-separated for some $\eps>0$. For this reason, we included $\eps$ into the definition.
Furthermore, note that the original definition of 
\cite{ChDaKpLu14a} does not take the maximum with $0$ on the left-hand side of 
\eqref{sep-gap}. However, 
the set $S$   may be empty, 
see Lemma \ref{lemma:base-components}. 
In this case, we have $\sup_{x\in S\pde}h(x) = -\infty$ and therefore 
\eqref{sep-gap} would be  satisfied for \emph{all} $\d>0$, $\e\in (0,1)$, and $\eps\geq 0$, 
if we had not included the maximum with $0$
on the left-hand side of \eqref{sep-gap}.
Finally, note that we always have $A_1\pde\cup A_2\pde = (A_1\cup A_2)\pde$, see 
e.g.~\cite[Lemma A.3.1]{SteinwartXXb1}, and hence the right-hand side of \eqref{sep-gap} does indeed coincide with that of the original definition.
For later use we also mention that  $\hdr \mycdot  {A_1\cup A_2}$ is obviously a decreasing function and 
additional properties of $\hdra\mycdot$ such as continuity 
can be found in Lemma \ref{lemma:twin-peaks-function}.

Let us now 
now describe the cluster algorithm of \cite{ChDaKpLu14a} and its 
 guarantees.
To this end, we
fix some $n,k\geq 1$, $\vs\in (0,1)$, $\d>0$, $\e\in (0,1)$, and  $\eps >0$. Moreover, 
we usually assume that $\e$ satisfies 
\begin{align}\label{min-eps}
 \e \geq C_{\vs,k,n} := C \cdot \log (2\vs^{-1})\cdot \sqrt{ \frac {d \log n}{k}} \,  ,
\end{align}
where 
$C$ is some universal constant that is not further specified in 
 \cite{ChDaKpLu14a}.\footnote{In \cite[Theorem VII.5]{ChDaKpLu14a} the authors actually omit the factor $2$ in the logarithm. However, in the proof of their  Lemma 4.1, they consider 
 our version of $C_{\vs,k,n}$, and since this lemma is key to all their guarantees including Theorem VII.5,
we decided to 
 streamline the presentation by including the factor $2$ in \eqref{min-eps}. In addition note that 
 \cite{ChDaKpLu14a} phrases \eqref{min-eps} as a condition on $k$ rather than on $\e$.} 
Furthermore, we define a minimal level 
\begin{align*}
 \r_{0,\d,\e} :=  \frac {2^d}{\vol_d}\cdot \frac {k}{n  \d^d} \cdot \frac{1+\e}{1-\e} + \frac {\eps}{1-\e} \, .
\end{align*}
Now  recall that 
the algorithms considered in \cite{ChDaKpLu14a} first construct a $k$-nearest neighbor density estimate 
$\hat h_{D,k}$ and then, for \emph{all} $\r\geq 0$, they 
build a graph $G_\rho$ with  vertices $D\cap \{\hat h_{D,k} \geq \r\}$.
Here, the edges are first set according to the local neighborhoods of the samples, see their Algorithms 1 and 2,
and then a pruning step  connects graph components in $G_\r$, if they belong to 
the same connected graph component at the lower level 
\begin{align*}
 \r_{\r, \vs,k,n,\eps} :=  
 \biggl|\r \cdot \frac{ 1 - C_{\vs,k,n}   }{1 + C_{\vs,k,n} } 
 - 
  \frac{ \eps}{1 + C_{\vs,k,n} }  \biggr|_+ \, ,
\end{align*}
where $|t|_+ := \max\{0,t\}$ for $t\in \R$. We refer to their algorithm in \cite[Figure 9]{ChDaKpLu14a} for details.
Here we note that \cite{ChDaKpLu14a} actually re-parametrizes the graphs, namely, the authors consider 
$\tilde G_r := G_{\rho(r)}$, where $\rho(r) = \frac k{n \vol_d r^d}$, and consequently,
 their formulas look 
different at first glance. 
In addition, note that this glueing strategy is different form ours as it is based on a vertical comparison, whereas ours 
is based on a horizontal approach.


With these assumptions and notations,  
Theorem VII.5 of \cite{ChDaKpLu14a} for their Algorithm 1 in combination with the described  pruning strategy
reads as follows:

\begin{customthm}{CDKL.VII.5}\label{thm:CDKL}
Let $P$ be a distribution
that has a continuous Lebesgue density $h$ 
and let $n,k\geq 1$, $\vs\in(0,1)$,  $\eps> 0$. Moreover we fix an $\e>0$ satisfying 
\eqref{min-eps}. 
Then with probability $P^n$ not smaller than $1-\vs$
the following two statements hold simultaneously:
\begin{enumerate}
 \item For all $\d>0$ and all path connected sets $A_1$ and $A_2$ that 
 are $(\d,2\e,\eps)$-separated and that  
 satisfies $\hdra\d   \geq \r_{0,\d,\e}$, 
 the sets $A_1\cap D$ and $A_2\cap D$ are contained in distinct connected graph components of 
 the graph $G_{\r'_\d}$ with level
 \begin{align*}
  \r'_\d := \frac {\hdra\d}{1+C_{\vs,k,n}}\, ,
 \end{align*}
 and each of them is 
 graph connected in $G_{\r'_\d}$.
 \item For all $\d>0$ with  $\eps \geq 2 \om(h,\d)$ and all $\r\geq \r_{0,\d,\e}$
  the set $A\cap D$
 is graph connected in $G_\r$ for  any  path connected component $A$ of $\{h\geq \r\}$.
\end{enumerate}
\end{customthm}

Before we compare this theorem to our results, let us first note that we did make a few 
changes to their original formulation:  
First, unlike in \cite[Theorem VII.5]{ChDaKpLu14a},
the sets $A_1$ and $A_2$ in part \emph{i)} are assumed to be path connected, since 
without this assumption it is easy to construct counterexamples, if, for example,
$\mr$ has three or more path connected components.
Second,  part \emph{ii)} 
 displays the 
 equivalent formulation considered in the proof of \cite[Lemma VII.4]{ChDaKpLu14a}, 
 since we believe that this formulation is a bit more intuitive.

 Now recall that our results are twofold in the sense that we provide rates for both
 estimating the split levels and estimating the corresponding clusters in measure.
 Clearly, the second type of guarantee cannot be provided by Theorem \ref{thm:CDKL}
 and the first type of guarantee does   not follow easily from Theorem \ref{thm:CDKL}, at least
 if one is interested in rates.
 Nonetheless, we will see below, that Theorem \ref{thm:CDKL} can actually be used to 
 derive learning rates  for  split estimation. This, however, requires some effort to 
 translate the notion of separation into a more suitable quantitative form.

 The first task in this direction is to free this notion from the burden of finding a
 \emph{good} separating set $S$.
To this end, let 
$h:\Rd\to [0,\infty)$ be  a continuous function with compact support
$X:=\overline{\{h>0\}}$, and  $A_1,A_2\subset X$ be two non-empty subsets.
Then, we define the \emph{separating valley function} 
$\hdsa \mycdot:[0,\infty)\to [0,\infty)$ by
 \begin{align*}
  \hdsa \d &:=  0 \vee \inf\biggl\{\,  \sup_{x\in S\pde} h(x) \,\,\Bigl|\, S \subset X\mbox{ separates $A_1$ and $A_2$  }    \biggr\}\, , \qquad\qquad \d\geq 0,
 \end{align*}
where we note that $\emptyset\pde = \emptyset$. Obviously, $\hdsa \mycdot$ is  an increasing function
and  
$\hdsa \d$
equals the minimal possible value on left-hand side 
of \eqref{sep-gap}. With this in mind, it is not surprising that   $A_1,A_2 \in \cc{\mhx \r}$ with $A_1\neq A_2$ for some $\r>0$
are $(\d,\e,\eps)$-separated 
if and only if 
\begin{align}\label{sep-gap-new}
 \hdsa \d < (1-\e)\hdra \d - \e\, ,
\end{align}
see Lemma \ref{lemma:separation-characterization} for details. Moreover, 
 Lemma \ref{lemma:separating-valley-function} 
lists additional properties of $\hdsa\mycdot$ such as continuity, and Theorem 
\ref{thm:valley-function-indentifies-split-level} shows that all split levels $\rs$
of a distribution satisfying \assx{C1} are of the form $\hdsa 0 = \rs$, where $A_1$ and $A_2$
satisfy the  assumptions above as well as $\hdsa 0>0$. In addition, it shows that such $A_1,A_2$
are a  pair of witnesses for $\rs$ at level $\r$. In other words, split levels can be fully  
 identified
with the help of the functions $\hdsa \mycdot$, and this observation will be a crucial aspect when  linking 
split level estimation with Theorem \ref{thm:CDKL}.

Besides find a separating set, part \emph{i)} of Theorem \ref{thm:CDKL} also allows some flexibility in the 
choice of $\d$ and $\e$. Here, a quick look shows that  $(\d,\e,\eps)$-separation becomes easier for smaller $\e$ and 
in addition, $\r_{0,\d,\e}$ is also decreasing in $\e$. Consequently, $\e$ should be chosen as small as possible, 
that is $\e:= c:= C_{\vs,k,n}$ by \eqref{min-eps}.  Unfortunately, determining the range of 
possible $\d$ in part \emph{i)} requires some more effort. To this end, let us assume that $P$ satisfies
\assx {C1} and that $\r>0$ is a level
for which there exists $A_1,A_2\in \cc{\mhx\r}$ with $A_1\neq A_2$. 
Let $\rs := \hdsa 0$ be the corresponding split level. Then Theorem \ref{thm:graph-level-range}
shows that $A_1$ and $A_2$ are $(\d,2c,\eps)$-separated for some $\d>0$ if and only if 
 \begin{align}\label{thm:graph-level-range-hxx}
 \r > \frac{\rs + \eps}{1-2c}\, ,
\end{align}
and if \eqref{thm:graph-level-range-hxx} is satisfied, then the 
set $\D_\r := \{ \d>0:  \mbox{$A_1$ and $A_2$ are $(\d,2c,\eps)$-separated}\}$ of 
allowed $\d$ is of the form 
$\D_\r = (0, \d_\r^*)$, where
\begin{align*}
 \d_\r^* &:= \min \biggl\{\d^*\geq0: \hdra{\d^*} = \frac{\hdsa{\d^*}+\eps}{1-2c}  \biggr\}\, .
\end{align*}
Note that in view of \eqref{sep-gap-new} this result is not overly surprising yet it requires some technical work.
Finally, we note that  $\r\mapsto \d_\r^*$ is an increasing map if the underlying pairs of connected components are witnesses for the same split level
$\rs$, see 
Lemma \ref{lem:drstar-is-monotone} for an exact formulation.

 With these preparations 
 we can now discuss the split level  guarantees provided by  Theorem \ref{thm:CDKL}. 
 To this end, we assume, like in \cite{ChDaKpLu14a},  that we have fixed a sequence $(k_n)_{n\geq 1}$  
 with 
 \begin{align*}
  \frac{k_n}{n} \to 0
  \qquad\qquad \mbox{ and } \qquad \qquad 
  \frac{ (\log n)^3}{k_n} \to 0\, .
 \end{align*}
Moreover, we fix some sequence $(\eps_n)_{n\geq 1}$ with 
 $\eps_n\to 0$ and write $\vs_n :=  2/n$. Then 
 \eqref{min-eps}
 reduces to 
 \begin{align*}
  c_n := C_{\vs_n,k_n,n} = C \cdot \sqrt{\frac{d\, (\log n)^3}{k_n}}\, ,
 \end{align*}
 and our assumptions ensure $c_n\to 0$. As already discussed before, we also choose 
%
%
$\e_n := c_n$. 
 For later use we finally write $C_d :=   {2^d}/{\vol_d}$ and pick an $n_0$ such that 
 $c_n < 1/3$ for all $n\geq n_0$.

 Let us now consider  Theorem \ref{thm:CDKL} for $n\geq n_0$.
 To this end, we assume
 that $P$   satisfies \assx {C1}.
  In addition, we fix some 
 split level $\rs>0$ and a witnessing level $\rss$ for $\rs$ with a corresponding witnessing pair 
 $A_1^*,A_2^*\in \cc{\mhx\rss}$. For a $\r\in (\rs, \rss)$ we write $A_{\r,i} := \z(A_i^*)$ for $i=1,2$,
 where $\z:\cc{\mhx\rss}\to \cc{\mhx\r}$ is the CRM. 
 Our final goal is to derive guarantees for estimating $\rs$ with the help of Theorem \ref{thm:CDKL}.

To this end, we first  investigate  the separation guarantees 
 for $A_{\r,1}$, $A_{\r,2}$.
Here, we write 
 \begin{align*}
  \ca L_n(\r) := \Bigl\{ \r': \exists \d>0 \mbox{ such that } A_{\r,1}, A_{\r,2} \mbox{ are $(\d,2\e_n,\eps_n)$-separated, $\r' := \frac{\hdrar \d}{1+c_n} \geq \frac{\r_{0,\d,\e_n}}{1+c_n}$} \Bigr\}
 \end{align*}
and note that  part \emph{i)} of Theorem \ref{thm:CDKL} provides a separation  guarantee
for $A_{\r,1}$, $A_{\r,2}$ if and only if $\ca L_n(\r)\neq \emptyset$.
Now \cite{ChDaKpLu14a} is mostly interested in a consistency notion that goes back to Hartigan \cite{Hartigan81a}
and that, roughly speaking, asks for  $P^n(\ca L_n(\r) \neq \emptyset) \to 1$. The latter is indeed true 
under the above assumption as shown in \cite{ChDaKpLu14a} and in our language this  quickly follows from 
 Theorem \ref{thm:graph-level-range}.
In the following, we are, however, interested in guarantees for estimating $\rs$. 
To this end, we consider
\begin{align*}
 \r_n := \inf\bigl\{ \r\in (\rs,\rss): \ca L_n(\r) \neq \emptyset  \bigr\}\, ,
\end{align*}
which describes smallest  level of the population cluster tree for which Theorem \ref{thm:CDKL} 
provides a separation guarantee. 
For later use we note that if $\ca L_n(\r)\neq \emptyset$ holds true for some 
$\r\in (\rs,\rss)$, then we also have $\ca L_n(\r')\neq \emptyset$ for all $\r'\in [\r,\rss)$.
%
Consequently, the set of $\r\in (\rs,\rss)$ with $\ca L_n(\r)\neq \emptyset$ is an interval as soon as this set is non-empty, and 
in this case we have $\ca L_n(\r)\neq \emptyset$ for all $\r\in (\r_n,\rss)$.

\begin{figure}
\includegraphics[width=0.45\textwidth]{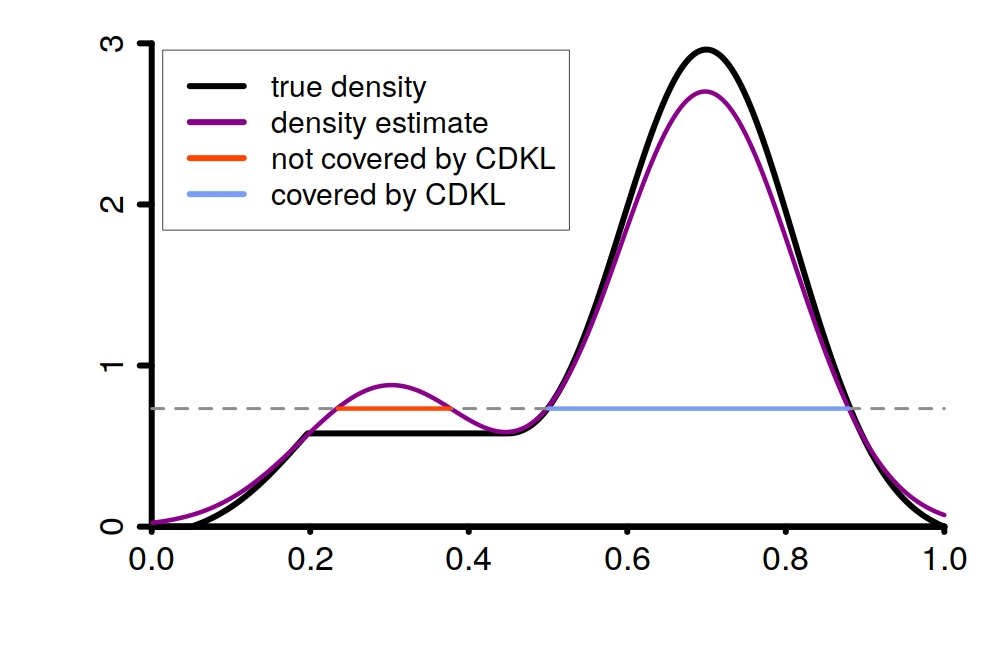}
\hspace*{0.03\textwidth}
\includegraphics[width=0.45\textwidth]{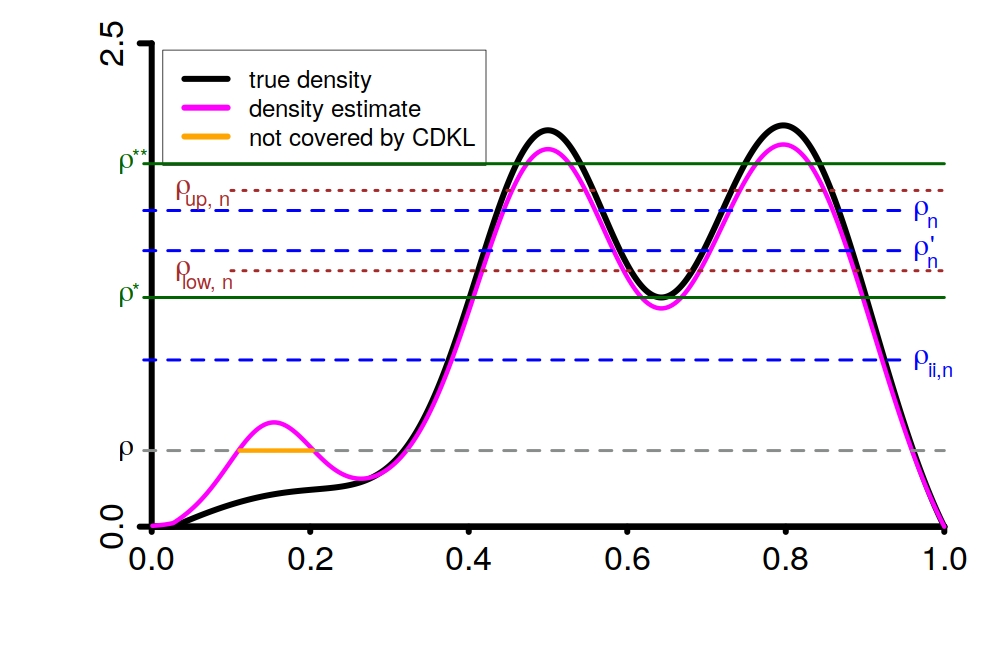}
\caption{\textbf{Left.} The left empirical cluster is not covered by Theorem \ref{thm:CDKL} since it does not intersect with a component of 
$\ccp{\mhx\r}$, while the right empirical cluster is covered.
\textbf{Right.} Guarantees provided by Theorem \ref{thm:CDKL} if $\riin < \rs$. Below the blue level $\riin$, no guarantees
are provided and hence spurious empirical clusters (orange) may appear. Between $\riin$ and $\rs$, part \emph{ii)} ensures that 
$D\cap \mhx \r$ is not split in the Graph $G_\r$, see Claim \ref{claim:no-false-cluster-guarantee}.
Still spurious clusters as in the left image are not excluded by Theorem \ref{thm:CDKL}. Theorem \ref{thm:CDKL} ensures that 
the first valid empirical  split $\hat \r_n$ is occurring between $\rs$ and $\r_n'$ and it corresponds 
to the population level $\r_n$ in the cluster tree of the true density $h$. For $n\geq n_0$ we have $\r_n'\leq \r_n$ by 
Claim \ref{claim1}.
Claim \ref{claim1} also provides a lower bound $\rnec$ on $\r_n'$ that describes the principal limitations 
of Theorem \ref{thm:CDKL} and an upper bound $\rsuf$ on $\r_n$ that can be used to obtain positive guarantees.
}
\label{figure:cdkl-guarantees}
\end{figure}

In general, the value $\r_n$ is not accessible to the user.
Instead, the user could use the split levels occurring in the empirical cluster tree 
as estimates for the true split levels. Clearly, such an approach is very similar to our ansatz.
Now the guarantees provided by Theorem \ref{thm:CDKL}
do not rule out the possibility of false empirical splits, if e.g.~$h$ has flat parts in the sense of 
$\mathring{\{h= \r\}}\neq \emptyset$ and the $k_n$-nearest neighbor density estimate $\hat h_{D,k_n}$
has fluctuations such that $D\cap \{\hat h_{D,k_n} \geq \r'\}$
has a graph component $A_D$ in the pruned graph  
with 
$A_D \cap \{h\geq \r'\}= \emptyset$
for some $\r>\r$. Note that Theorem \ref{thm:CDKL} does not rule out such $A_D$ since its
guarantees only consider graph components with samples in $D\cap \{h\geq \r'\}$, see also Figure \ref{figure:cdkl-guarantees}
for an illustration. 
However, it seems likely that an additional pruning step similar to the one in Line 3 of our
Algorithm \ref{cluster-algo-generic} may resolve this issue, and for this reason we will ignore 
such false empirical splits in the following discussion.

Now assume that a found split $\hat \r_n$ in the empirical cluster tree is valid for $\rs$.
To be more precise, assume that there exists a  $\r\in (\rs,\rss)$
such that $A_{\r,1}\cap D$ and $A_{\r,2}\cap D$ are contained in distinct connected
graph components of the graph $G_{\r_\d'}$ for some $\d>0$ according to part \emph{i)} 
and both sets are graph connected in $G_{\r_\d'}$. In this situation, let 
$\hat \r_n$ be the level of the graph at which the split occurs for the first time.
%
Then, by the very definition of $\r_n$ and the formulation of part \emph{i)} we see that 
\begin{align*}
 \hat \r_n \leq \inf \bigl\{\inf \ca L_n(\r) :\r\in (\r_n, \rss)\bigr\} =: \r_n'
\end{align*}
and that $\r_n'$ is the best upper bound of $\hat \r_n$ provided by part \emph{i)} 
and below we will 
see how part \emph{ii)} can be used to ensure $\hat \r_n\geq\rs$ for all sufficiently large $n$.
Since we obviously have $\inf \ca L_n(\r) \leq \r$, we further find $\r_n'\leq \r_n$.
Consequently, upper bounds on $\r_n$ will lead to positive guarantees 
for $\hat \r_n \to \rs$, whereas lower bounds on $\r_n'$ will determine principal 
limitations of the guarantees provided by Theorem \ref{thm:CDKL}. In the following, we will derive both
types of bounds. 

To this end, we will formulate an couple of intermediate claims. The first of these claims 
establishes auxiliary bounds on $\r_n'$ and $\r_n$.

 \begin{claim}\label{claim1}
  For $n\geq   n_0$ we define 
  \begin{align*}
  \dsuf^d := 2C_d \cdot \frac{k_n}{  n\cdot \rs}   
  \qquad \qquad \mbox{ and } \qquad \qquad 
  \dnec^d := C_d \cdot \frac {k_n}{n \cdot \rss   }
\end{align*}
as well as 
\begin{align*}
\rnec &:= \inf\biggl\{ \frac{\hdrar \d}{1+c_n}:  \r\in (\r_n,\rss), \d \in [\dnec,\d_\r^*)  \biggr\} \, ,\\
 \rsuf &:= \inf\biggl\{\r\in\Bigl(\frac{\rs + \eps_n}{1-2c_n},\rss\Bigr): \dsuf < \d_{\r}^*  \biggr\}\, .
\end{align*}
Then, for all $n\geq n_0$, we have $\rnec \leq \r_n' \leq \r_n \leq \rsuf$.
 \end{claim}

Our next claim investigates, which levels are covered by  part \emph{ii)} of Theorem \ref{thm:CDKL}.

\begin{claim}\label{claim:no-false-cluster-guarantee}
 For $n\geq n_0$ we define 
 \begin{align*}
 \diin & := \sup\bigl\{  \d>0 \mbox{ such that } 2\om(h,\d) \leq \eps_n  \bigr\}\, ,\\
   \riin  &:= \inf\bigl\{  \r_{0,\d, \e_n}: \d>0 \mbox{ such that } 2\om(h,\d) \leq \eps_n  \bigr\}\, .
 \end{align*}
Then we have $\riin =  \r_{0,\diin, \e_n}$. 
Moreover, for all $\r> \riin$ and all $A\in \ccp{\mhx\r}$  
Theorem \ref{thm:CDKL} ensures that $A \cap D$ is graph connected in $G_\r$.
Conversely, if 
$\rs>0$ is the smallest split level in the cluster tree 
and we have 
$|\cc {\mhx\r}|=1$ for all $\r\in (0,\rs]$, then Theorem \ref{thm:CDKL} does not guarantee
that $\mhx \r \cap D$ is graph connected in $G_\r$ for any $\r>0$ with $\r<\min\{\rs, \riin\}$.
%
\end{claim}

Now,
with the help of Claims \ref{claim1} and \ref{claim:no-false-cluster-guarantee} it is easy 
investigate consistency for 
 $\hat \r_n$. To this end, let us assume for a moment that
  $\rs$ is the smallest split level in the cluster tree 
and $h$ satisfies
$|\cc {\mhx\r}|=1$ for all $\r\in (0,\rs]$. In addition, assume that $n$ is sufficiently large in the sense of 
$\riin < \rs$, where we emphasize that this assumption can only be guaranteed for all possible values of $\rs$
if $\riin \to 0$.
Then 
splits in the graphs $G_\r$
 can only occur for $\r< \riin$ or $\r\geq \rs$. Since the former are
 false splits and every valid split $\hat \r_n$ satisfies $\hat \r_n\geq \rs$ because of the assumed  
 $\riin < \rs$, every split below $\riin$ should be ignored. 
 In the following claim investigating consistency, we therefore assume that splits below are indeed ignored
 despite the fact that 
  $\riin$ may in general be not 
accessible to us.


\begin{claim}\label{claim:consistency}
If we have $\eps_n \geq 2\om (h,\dsuf)$ for all sufficiently large $n$ 
then  the discussed estimator  $\hat \r_n$  
for $\rs$ is consistent, that is
\begin{align*}
 \hat \r_n \to \rs\, .
\end{align*} 
Conversely, if $\rs$ is the smallest split level in the cluster tree 
with
$|\cc {\mhx\r}|=1$ for all $\r\in (0,\rs]$
and Theorem \ref{thm:CDKL} ensures consistency for estimating  $\rs$,
then 
$\eps_n \geq 2\om (h,\dsuf/2)$ holds for all sufficiently large $n$.
\end{claim}

 Recall that both $\e_n$ and $k_n$ are set by the user, and therefore, the asymptotic behavior of 
 $\dsuf$ is also determined by the user. As consequence, the conditions 
 $\eps_n \geq 2\om (h,\dsuf)$ respectively $\eps_n \geq 2\om (h,\dsuf/2)$  actually 
 restrict the class of distributions for which Theorem \ref{thm:CDKL} ensures consistency.
 Or, to phrase it differently, achieving consistency with the help of Theorem \ref{thm:CDKL}
 requires to impose a worst-case behavior on $\om(h, \mycdot)$. 
 In contrast, our consistency results, see e.g.~Corollary \ref{cor:consis}, do not require 
 such a continuity assumption but instead, a bound on the thickness needs to be imposed.

 In this respect we note that the sufficient condition $\eps_n \geq 2\om (h,\dsuf)$ may  be superfluous 
for split levels further up in the tree: Indeed, this condition 
is used for ensuring $\riin \leq \r_{0,\dsuf, \e_n} < \rs$ and thus 
$\hat \r_n \geq \rs$ with the help of part \emph{ii)}, but 
since part \emph{i)} also states a connectivity guarantee, the application of 
 part \emph{ii)} may not be necessary for ensuring $\hat \r_n \geq \rs$.
Finally note that Claim \ref{claim:consistency}
does not state that the condition $\eps_n \geq 2\om (h,\dsuf/2)$ is necessary for consistency. It only states 
that a proof of consistency with the help of Theorem \ref{thm:CDKL} requires this assumption.

\begin{claim}\label{clam1:upper-bound}
 Let
$\om:[0,\infty)\to [0,\infty)$ be a function such that 
\begin{align}\label{disco:cdkl-pos-ass}
 \hdsar \d \leq \rs + \om(\d) 
 \qquad \mbox{ and } \qquad 
 \r - \om(\d) \leq \hdrar \d 
\end{align}
holds for all $\r\in (\rs,\rss)$ and $\d>0$.
Then we have 
\begin{align}\label{upper-bound-rn}
 \rsuf \leq \rs + 4 \om(\dsuf) + 3\eps_n + 6\rs c_n  
\end{align}
for all  $n\geq n_0$, for which the right-hand side of \eqref{upper-bound-rn} 
is smaller than $\rss$.
\end{claim}

Before we proceed we note that Condition \eqref{disco:cdkl-pos-ass}
is always satisfied for the modulus of continuity $\om(h,\mycdot)$
as it is shown in the Lemmas  \ref{lemma:separating-valley-function} and \ref{lemma:twin-peaks-function}. In particular, 
if $h$ is $\theta$-H\"older continuous with constant $|h|_\theta$, then 
$\om(\d) := |h|_\theta\, \d^\theta$ satisfies \eqref{disco:cdkl-pos-ass}.

With the help of Claim \ref{clam1:upper-bound} it is already possible to 
establish rates for $\hat \r_n\to \rs$, if, e.g.~$h$ is $\theta$-H\"older continuous.
To avoid unnecessary repetitions in the computations of these rates, it is, however,
more efficient, to establish a lower bound on $\rnec$, first.

\begin{claim}\label{clam1:lower-bound}
Let
$\nu:[0,\infty)\to [0,\infty)$ be a function 
such that 
\begin{align}\label{clam1:lower-bound-h1}
    \hdsar \d \geq  \rs + \nu(\d) 
\end{align}
holds for all $\r\in (\rs,\rss)$ and all $\d>0$ with $\hdsar \d\leq \r$.
Then, for all $n\geq n_0$, we have 
\begin{align}\label{clam1:lower-bound-h2}
 \rnec \geq  \rs + c_n \rs + \nu(\dnec) +\eps_n \, .
\end{align}
Moreover, if there exists a $\r\in (\rs,\rss)$ with $\hdsar \d>0$ for all $\d>0$, then 
there exists an increasing function $\nu$ satisfying \eqref{clam1:lower-bound-h1} with $\nu(\d) >0$ for all $\d>0$.
\end{claim}

We first note that the second part of Claim \ref{clam1:lower-bound} essentially shows that there exists 
a non-trivial function $\nu$ satisfying \eqref{clam1:lower-bound-h1} whenever there is no plateau 
at the split level $\rs$. Moreover note that for all functions  $\om$ and $\nu$
satisfying  \eqref{disco:cdkl-pos-ass}
and \eqref{clam1:lower-bound-h1}, respectively, we obviously have $\nu(\d) \leq \om(\d)$
for all $\d>0$. If, in addition,  $\nu$ is increasing and  $\nu(\d/4) \succeq \om(\d)$ for $\d\to 0$, then 
the bounds \eqref{upper-bound-rn} and \eqref{clam1:lower-bound-h2} have the same asymptotic behavior for 
$n\to \infty$ as soon as we choose a witness level $\rss \leq 2\rs$, which is, of course, always possible,
In this respect we note that in general we do not have $\nu(\d/4) \sim \om(\d)$.
For example, for general  Lipschitz continuous densities $h$ the best choice for $\om$ behaves like $\om (\d)\sim \d$,
while for any $\kappa\geq 1$ it is easy to construct such $h$ with best possible $\nu$ behaving like 
$\nu(\d)\sim \d^{\kappa}$

\begin{claim}\label{claim:the-three-guarantees}
 Assume that $h$ is $\theta$-H\"older continuous, i.e.~$\om(h,\d) \leq |h|_\theta \d^\theta$ for all $\d>0$,
 and that $\theta$ is sharp in the sense of $ K_h \d^\theta\leq \om(h,\d)$ for a fixed $K_h\in (0,|h|_\theta]$ and all $\d\in (0,1]$.
 Then, there exists constants $C_{\mathrm{low}}, C_{\mathrm{up}}\in (0,\infty) $ only depending on $\rs, d, |h|_\theta, K_h$, and $\theta$ such that  
 for all $n\geq n_0$ satisfying $\eps_n \geq 2\om (h,\dsuf/2)$ and  $\dsuf \leq 2$
we have 
\begin{align}\label{claim:the-three-guarantees-g1}
C_{\mathrm{low}} \cdot \biggl(\eps_n +  \frac{ (\log n)^{3/2}}{\sqrt {k_n}} \biggr)
   \leq   
   \rnec  - \rs 
   \leq         \rsuf  - \rs 
   \leq
   C_{\mathrm{up}} \cdot \biggl(\eps_n +  \frac{ (\log n)^{3/2}}{\sqrt {k_n}} \biggr)\, ,
\end{align}
where the upper bound on $\rsuf  - \rs$
only holds, if it is smaller than $\rss$. 
 Moreover, for 
all $n\geq n_0$ we have 
\begin{align}\label{claim:the-three-guarantees-g2}
  C_{\mathrm{low}} \cdot\biggl({\eps_n} + \frac {k_n}{n  } \cdot \eps_n^{- d/\theta}   \biggr)
 \leq 
\riin 
\leq 
   C_{\mathrm{up}} \cdot\biggl({\eps_n} + \frac {k_n}{n  } \cdot \eps_n^{- d/\theta}   \biggr) \, .
\end{align}
\end{claim}

To apply Claim \ref{claim:the-three-guarantees} we first note that the condition
$\eps_n \geq 2\om (h,\dsuf/2)$ cannot be dropped by Claim \ref{claim:consistency}.
In addition, we note that this condition
implies 
\begin{align*}
\eps_n 
\geq 
2\om (h,\dsuf/2) 
\geq 
2^{1-\theta} K_h \dsuf^\theta 
\geq 
2 K_h \cdot \Bigl(\frac{C_d}{\rs}\Bigr)^{\theta/d}   \cdot \Bigl(\frac{k_n}{n}\Bigr)^{\theta/d} \, .
\end{align*}
In view of the guarantees \eqref{claim:the-three-guarantees-g1}, it is then tempting to choose 
$\eps_n \sim (\frac{k_n}{n})^{\theta/d}$, but for this choice we would obtain  $\inf_{n\geq 1}\riin >0$
by \eqref{claim:the-three-guarantees-g2}. In other words, the range of levels $\r\in (0,\riin]$ for which 
we cannot exclude false splits with the help of Theorem \ref{thm:CDKL} would not vanish for $n\to \infty$.
Since the exclusion of such false splits is as important as the detection of 
true splits we now consider the ideal  case in which the guarantees for both have the same order, that is 
$\riin \sim \rsuf-\rs$.
Now note that \eqref{claim:the-three-guarantees-g1} and \eqref{claim:the-three-guarantees-g2} show that this case 
is satisfied if and only if 
\begin{align*}
 \frac {k_n}{n  } \cdot \eps_n^{- d/\theta} \sim  \frac{ (\log n)^{3/2}}{\sqrt {k_n}}\, .
\end{align*}
Obviously, the latter is equivalent to $k_n    \sim  n^{2/3} { \log (n)}{} \cdot \eps_n^{\frac{2d}{3\theta}}$
and inserting this into, e.g.~\eqref{claim:the-three-guarantees-g2}, then gives 
\begin{align*}
 \riin   
 \sim 
 \eps_n +   \frac {n^{2/3} { \log (n)}{} \cdot \eps_n^{\frac{2d}{3\theta}}}{n  } \cdot \eps_n^{- d/\theta}
 = 
 \eps_n + n^{-1/3} \log (n) \cdot \eps_n^{-\frac{d}{3\theta}}\, .
\end{align*}
A quick calculation now shows that the latter expression achieves the fastest rate of convergence if 
\begin{align*}
 \eps_n \sim n^{-\frac{\theta}{3\theta+d}}(\log n)^{\frac{3\theta}{3\theta+d}}\, .
\end{align*}
Now observe that 
\begin{align*}
 \om (h,\dsuf/2) \sim \Bigl(\frac{k_n}{n}\Bigr)^{\theta/d} \sim  n^{-\frac{\theta}{3d}} { (\log n)^{\frac \theta d  }}{} \cdot \eps_n^{\frac{2}{3}}
\end{align*}
and therefore $\frac{\theta}{3\theta+d} < \frac \theta d$ ensures that this behavior  of $\eps_n$
guarantees $\eps_n \geq 2\om (h,\dsuf/2)$  for all sufficiently large $n$. Consequently, Claim \ref{claim:the-three-guarantees}
yields 
\begin{align}\label{claim:the-three-guarantees-final}
 \riin \sim \rsuf-\rs \sim \rnec-\rs \sim \eps_n \sim n^{-\frac{\theta}{3\theta+d}}(\log n)^{\frac{3\theta}{3\theta+d}} \, ,
\end{align}
and since $\rnec-\rs \sim \eps_n$ these are the best rates one can obtain from Theorem \ref{thm:CDKL}.

Our next goal is to compare these rates to ours, where 
in the following we will ignore all logarithmic factors in the considered rates.
Let us begin by 
recalling from \cite[Lemma A.10.1]{SteinwartXXb1} that $\theta$-H\"older continuity 
implies a separation exponent $\k = \theta$. Modulo logarithmic factors,
our rates for estimating the split level $\rs$ are therefore of the form 
\begin{align*}
n^{- \frac {\g \k}{2\g \k + d}} =  n^{- \frac {\g \theta}{2\g \theta + d}} \, ,
\end{align*}
see Corollary 
\ref{rates-cor1}, where $\g\in (0,1]$ describes the thickness of the levels. Consequently, if we ignore logarithmic factors,
our rates 
are better if and only if $\g> \frac {d}{\theta + d}$. In particular, for the most natural case $\g = 1$, 
see e.g.~the detailed discussion in \cite[Appendix A.5]{SteinwartXXb1},
our rates are 
better in all dimensions.
In addition, if $h$ is a Morse function with compact support, 
then \cite[Proposition 2]{WaLuRi19a} shows that we 
actually have  
$\kappa = 2$, and since for such $h$ we also have $\theta = 1$, our results are better in the above sense if and only if  
$\g> \frac {d}{2 + 2d}$. Finally, our results also apply to clusters with 
very strong separation, that is $\k\to \infty$, and in the extreme case $\k=\infty$,
dimension and thickness independent rates of the form $n^{-1/2}$ are possible, see again 
Corollary \ref{rates-cor1}. In particular, there are discontinuous densities with $\k=\infty$, e.g.~step functions on rectangles
with mutually positive distances, and for such densities Theorem \ref{thm:CDKL} does not provide any guarantee at all.
Note that in all these cases, our rates can be achieved without knowing $\theta$ or $\k$, while 
\cite{ChDaKpLu14a} do not offer such adaptivity.

Finally, going back to the most comparable case $\k=\theta$, one might ask for the reason of the difference in rates.
Clearly, a possible explanation could focus on the difference in the pruning strategy:
While we follow an  horizontal approach by considering $\t$-connected components, 
\cite{ChDaKpLu14a} considers a vertical approach as described above. 
However, in \cite{SrSt12a} we also investigated a vertical pruning strategy for known $\theta$ and without
thickness assumption, i.e.~we considered exactly the scenario in which Theorem \ref{thm:CDKL} yields meaningful guarantees,
but unlike in Claim \ref{claim:the-three-guarantees}, respectively \eqref{claim:the-three-guarantees-final}, 
our rates are of the form 
$n^{- \frac{\theta}{2\theta + d}}$. In other words, taking a vertical pruning strategy 
does not necessarily lead to worse rates. On the other hand, without pruning, that is $\eps_n=0$, 
the bound \eqref{claim:the-three-guarantees-g1} also leads to $n^{- \frac{\theta}{2\theta + d}}$-rates
for optimized $k_n$, and consequently, the guarantees in part \emph{ii)} of Theorem \ref{thm:CDKL}
for the
particular pruning of \cite{ChDaKpLu14a}
are indeed responsible for the difference in rates. It is however, unclear to us, whether 
part \emph{ii)} is optimal, i.e.~whether the pruning of \cite{ChDaKpLu14a} or its analysis 
is sub-optimal.

Finally, we like to emphasize that  \cite{ChDaKpLu14a} also contains some 
lower bounds on possible clustering guarantees for $(\d,\e,\eps)$-separated components. 
Since we do not provide such lower bounds, we have excluded these results from our comparison.

\subsection{Proofs Required for the Supplement}\label{sec:comparison-proofs}

This section provides the proofs of all results and claims presented earlier in this supplement  as well as
some auxiliary results to establish them. 
We begin with the following well-known result, which is only included for the same of 
completeness.

\begin{lemma}\label{lem:path-implies-connect}
 Let $(X,d)$ be a metric space and $A\subset X$ be path connected. Then $A$ is connected.
\end{lemma}

\begin{proofof}{Lemma \ref{lem:path-implies-connect}}
 Let us assume that $A$ is not connected. Then there exist non-empty and relatively 
 open $A_0,A_1\subset A$
 with $A_0\cap A_1=\emptyset$ and $A_0\cup A_1 = A$. In particular, we find some $x_0\in A_0$
 and $x_1\in A_1$, and since $A$ is path connected, there then exists a path $\g:[0,1]\to A$
 with $\g(0) = x_0$ and $\g(1) = x_1$. Consequently, $\g^{-1}(A_0)$ and $\g^{-1}(A_1)$
 are non-empty, and since $\g$ is a continuous between the topological spaces $[0,1]$ and $A$, 
 these pre-images are also open. By construction, they also form a partition of $[0,1]$
 and hence $[0,1]$ is not connected, which is known to be false.
\end{proofof}

\begin{lemma}\label{lem:path-vers-connect-compos}
 Let $(X,d)$ be a metric space and $A\subset X$ be non-empty. Then we have 
 $\ccp A \comparable \cc A$.
\end{lemma}

\begin{proofof}{Lemma \ref{lem:path-vers-connect-compos}}
  Let us fix an $A'\in \ccp A$. We define 
 \begin{align*}
  \ca A := \bigl\{ A'\cap A'':     A''\in \cc A \mbox{ and } A'\cap A'' \neq \emptyset \bigr \}\, .
 \end{align*}
  Clearly, this gives $A' = \bigcup_{A'''\in \ca A}A'''$, and therefore our goal is to show that 
 $|\ca A| = 1$. 
 To this end, we fix an $A''' = A'\cap A''\in \ca A$. Then we note that 
 $A''$ is relatively closed in $A$ since $A''$ is a connected component of $A$, and by 
 $A'\subset A$ we then see that  $A'''$ is relatively closed in $A'$.
 Moreover, by construction the elements in $\ca A$ 
are also non-empty and mutually disjoint, and their union 
 gives $A'$. This implies $|\ca A| = 1$ since otherwise 
  $A'$ was not connected, and this would contradict Lemma \ref{lem:path-implies-connect}.
\end{proofof}

\begin{lemma}\label{lem:path-comparable}
 Let $(X,d)$ be a metric space and $A\subset B\subset X$ are non-empty sets. Then we have
 $\ccp A \comparable \ccp B$.
\end{lemma}

\begin{proofof}{Lemma \ref{lem:path-comparable}}
 Let us fix an $A'\in \ccp A$. We define 
 \begin{align*}
  \ca B := \bigl\{B'\in \ccp B: A'\cap B' \neq \emptyset \bigr\}\, .
 \end{align*}
 Clearly, this gives $A' \subset \bigcup_{B'\in \ca B}B'$, and our goal is to show that 
 $|\ca B| = 1$. To this end, we assume that $|\ca B| > 1$. Then, there exist
 $B_0,B_1\in \ca B$ with $B_0\neq B_1$, and our construction ensures the existence of 
 some $x_0\in A'\cap B_0$ and $x_1\in A'\cap B_1$. Since $A'$ is path connected, there then exists 
 path $\g$ connecting $x_0$ and $x_1$ in $A'$, and $A'\subset A\subset B$ implies, that $\g$
 also connects $x_0$ with $x_1$ in $B$. Consequently, $x_0$ and $x_1$ are contained in the same 
 path connected component of $B$, which contradicts $B_0\neq B_1$.
\end{proofof}

\begin{lemma}\label{lem:open-path-components}
 Let $A\subset \Rd$ be non-empty and open. Then every $A'\in \ccp A$ is open.
\end{lemma}

\begin{proofof}{Lemma \ref{lem:open-path-components}}
 Let us fix an $A'\in \ccp A$ and an $x\in A'$. Since $A$ is open, 
  there then exists an $\e>0$ such that $B (x,\e)\subset A$.
 For a fixed  $y\in B (x,\e)$ we now define $\g:[0,1]\to \Rd$ by
 $\g(t) := (1-t)x + ty$ for $t\in [0,1]$. Clearly, $\g$ is continuous with $\g(0) = x$ and $\g(1) = y$, 
 and $\g(t)\in B (x,\e)\subset A$ for all $t\in [0,1]$ shows that $\g$ connects $x$ and $y$ 
 in $A$. Consequently, $x$ and $y$ are in the same path connected component of $A$, and hence $y\in A'$.
 This shows $B (x,\e) \subset A'$, i.e., $A'$ is indeed open.
\end{proofof}

\begin{lemma}\label{lem:open-path-persists}
 Let $A\subset \Rd$ be open and non-empty. Then we have $\ccp A = \cc A$.
\end{lemma}

\begin{proofof}{Lemma \ref{lem:open-path-persists}}
 For a fixed  $A'\in \cc A$ we   define $\ca B:= \{B'\in \ccp A: B'\subset A'\}$. 
 By Lemma \ref{lem:path-vers-connect-compos} we already know $\ccp A \comparable \cc A$, and consequently 
 we obtain $A' = \bigcup_{B'\in \ca B}B'$. Moreover, all $B'\in \ca B$ are open by Lemma
 \ref{lem:open-path-components}, and they are thus also  relatively open in $A$.
 Since they are also non-empty and $A'$ is connected, we conclude that 
 $|\ca B| = 1$, which finishes the proof.
\end{proofof}

\begin{lemma}\label{lem:open-top-comparable}
 Let $A\subset \Rd$ be non-empty and $B\subset \Rd$ open with $A\subset B$.
 Then we have $\cc A\comparable \cc B$.
\end{lemma}

\begin{proofof}{Lemma \ref{lem:open-top-comparable}}
 For a fixed  $A'\in \cc A$ we   define $\ca B:= \{B'\in \cc B: A'\cap B' \neq \emptyset\}$. 
 This gives $A' \subset  \bigcup_{B'\in \ca B}B'$ and therefore, our goal is to show that 
 $|\ca B| = 1$. Let us assume the converse, so that we can pick a $B_0\in \ca B$
 such that $B_1 := \bigcup_{B'\in \ca B: B'\neq B_0}B'$ is non-empty. 
 By construction, we then see that $A_0:= A'\cap B_0\neq \emptyset$ and $A_1:= A'\cap B_1\neq \emptyset$
 form a partition of $A'$.
 Moreover, all $B'\in \ca B$ are open by Lemmas 
 \ref{lem:open-path-components} and \ref{lem:open-path-persists}, and consequently, $A_0$ and $A_1$ are open in $A'$ with respect to the 
 relative topology of $A'$. However, this means that $A'$ is not connected, which contradicts
 $A'\in \cc A$.
\end{proofof}

\begin{lemma}\label{lemma:base-components}
   Let $h:\Rd\to [0,\infty)$ be continuous and 
    $X:= \overline{\{h>0\}}$. Then, for all $\r>0$ with $\mhx\r \neq \emptyset$, we have 
   \begin{align*}
    \cc {\mhx\r} \comparable  \ccp X \, .
   \end{align*}
   Moreover, if   $\z:\cc {\mhx\r}\to \ccp X$ is the corresponding CRM
   and $A_1,A_2 \in \cc {\mhx\r}$ are distinct, then $S:= \emptyset$ separates $A_1$ and $A_2$ if and only if 
   $\z(A_1) \neq \z(A_2)$.
\end{lemma}

\begin{proofof}{Lemma \ref{lemma:base-components}}
 We begin by  observing that 
 Lemmas \ref{lem:open-top-comparable}, \ref{lem:open-path-persists}, and \ref{lem:path-comparable}   imply
  \begin{align*}
  \cc {\mhx\r} \comparable \cc {\mhxg 0} = \ccp {\mhxg 0} \comparable \ccp X \, .
 \end{align*}
 To establish the equivalence we  define $X_i := \z(A_i)$
for $i=1,2$, which in turn yields $A_1\subset X_1$ and $A_2\subset X_2$. 

Let us first assume that $X_1 = X_2$. Since $A_1$ and $A_2$ are non-empty and $X_1$ is path connected,
 there then exist $x_1\in A_1$, $x_2\in A_2$, and a path $\g:[0,1]\to X_1$ connecting $x_1$ and $x_2$.
 Clearly, $\g$ also connects $x_1$ and $x_2$ in $X$ and therefore we have $\g([0,1]) \cap S \neq \emptyset$
 for every $S$ separating $A_1$ and $A_2$. This shows that 
 $S:= \emptyset$ does not separate $A_1$ and $A_2$.
 
Let us now assume that 
 $X_1 \neq X_2$. We fix some $x_1\in A_1$ and $x_2\in A_2$. This gives $x_1\in X_1$ and $x_2\in X_2$,
 and since $X_1$ and $X_2$ are path connected components of $X$, it is then impossible 
 to find a path $\g$ connecting $x_1$ and $x_2$ in $X$. Consequently, the condition for separating 
 sets $S$ becomes void, and hence
 $S:= \emptyset$ indeed separates $A_1$ and $A_2$.
\end{proofof}

\begin{lemma}\label{lem:continuous-boundary}
 Let $h:\Rd\to [0,\infty)$ be continuous and  $\r>0$ with $\emptyset \neq \mhx \r\subsetneq  \Rd$. 
 Then   all $A\in \cc{\mhx\r}$ are closed and  we have both $\partial A\neq \emptyset$ and $h(x) = \r$ for all $x\in \partial A$.
\end{lemma}

\begin{proofof}{Lemma \ref{lem:continuous-boundary}}
To check that $A$ is closed it suffices to recall that $A$, as a connected component of $\mhx\r$ is 
closed in the relative topology of $\mhx\r$. Since $\mhx\r$ is closed in $\Rd$, we conclude that so is $A$,
i.e.~$\overline A = A$.

 Next, we quickly verify that $\partial A\neq \emptyset$. To this end, we
 assume that $\partial A=\emptyset$.
 Then we have $\emptyset = \partial A = \overline A\setminus \mathring A =
 A\setminus \mathring A$, and hence $A = \mathring A$. 
 In other words, $A$ is both open and closed, and hence  the partition 
 $\Rd =   A \cup  (\Rd\setminus A)$ consists of two open and non-empty sets.
 However, $\Rd$ is connected, and therefore such a  partioning  is impossible.
 
 Let us finally establish the behavior of $h$ on the boundary of $A$. 
 To this end, we first show 
 \begin{align}\label{lem:continuous-boundary-h1}
  A\cap \mhxg \r \subset \mathring A \, .
 \end{align}
To this end, we  fix an $x\in A\cap \mhxg \r$.
 By the continuity of $h$ there then exists a $\d>0$ such that $B(x,\d)\subset \mhxg\r$.
 Let us fix an $y\in B(x,\d)$. Since $B(x,\d)$ is path connected, there then exists a 
 path connecting $x$ and $y$ in $B(x,\d)$, and 
 using $B(x,\d)\subset \mhxg\r\subset \mhx\r$ we conclude that $x$ and $y$ are in the same 
 path connected component of $\mhx\r$. Lemma \ref{lem:path-vers-connect-compos}  
 then shows that they are also in 
  the same connected component of 
 $\mhx\r$, i.e. $y\in A$. In other words, we have shown 
 $B(x,\d)\subset A$, and this 
 gives the desired  $x\in \mathring A$.

 With the help of \eqref{lem:continuous-boundary-h1} and $\overline A = A$ we now observe that 
 \begin{align*}
  \partial A 
  = A \setminus \mathring A 
  \subset A \setminus \bigl( A\cap \mhxg \r  \bigr) 
  = A \setminus \mhxg \r
  \subset \{h \leq \r  \}\, .
 \end{align*}
Moreover, $\overline A = A$ also implies 
$\partial A \subset A\subset \mhx\r$, and combining both set 
inclusions then gives $\partial A \subset \{h = \r  \}$.
\end{proofof}

\begin{lemma}\label{lemma:existence-of-cluster-peaks}
 Let $h:\Rd\to [0,\infty)$ be continuous with compact support. Then the following statements hold true:
 \begin{enumerate}
  \item For all $\r>0$ and $A\in \cc{\mhx\r}$ there exists 
 a cluster peak $A'\subset A$.
 \item We have  $|\cc{\mhx\r}|\leq |\setcp|$ for all $\r\in(0,\inorm h]$.
 \item We have $|\setsl|+1\leq |\setcp|$.
 \end{enumerate} 
\end{lemma}

\begin{proofof}{Lemma \ref{lemma:existence-of-cluster-peaks}}
 \ada i Let us write $\rdcp := \sup_{x\in A}h(x)$. Since $A$ is closed in $\Rd$ and bounded, there then exists an 
 $x^\ddagger\in A$ with $h(x^\ddagger)=\rdcp$. Let $A'\in \cc{\mhx\rdcp}$ be the connected component with 
 $x^\ddagger\in A'$. Using $\cc{\mhx\rdcp}\comparable \cc{\mhx\r}$ we then find $A'\subset A$ and 
 in turn this implies $h(x) \leq \rdcp$ for all $x\in A'$. Moreover, the converse inequality is guaranteed by 
 $A'\in \cc{\mhx\rdcp}$.
 
 \ada {ii} By \emph{i)} we know that for each $A\in \cc{\mhx\r}$ there exists 
 an $A'\in \setcp$ with $A'\subset A$, and since the connected components of $\cc{\mhx\r}$ are mutually disjoint
 so are the found cluster peaks. This gives the assertion.
 
 \ada {iii} Clearly it suffices to show that  the existence of at least $n\geq 1$ 
 split levels $\r^*_1<\dots<\r^*_n$ implies the existence of at least $n+1$ cluster peaks.
 To this end, we begin with some preparations. Namely, for $i=1,\dots,n$ we fix $\r_i^{**}>\r^*_i$
 and $A_{1,i},A_{2,i}\in \cc{\mhx {\r_i^{**}}}$ according to the definition of split levels.
 Without loss of generality we additionally assume that $\r_i^{**}<\r^*_{i+1}$ for $i=1,\dots,n-1$.
 For later use we note that for $i_1< i_2$ this construction ensures $\r_{i_1}^{**} < \r_{i_2}^{**}$,
 and by Lemma \ref{lem:continuous-boundary} we thus find $A_{l_1,i_1}\neq A_{l_2,i_2}$
 for any choice of $l_1$ and $l_2$. Consequently, we have fixed $2n$ different connected components, altogether.
 Furthermore, we write $\r_0^{**} := \r_1^*/2$ and 
 \begin{align*}
  \ca C := \bigl\{ A\in \cc{\mhx{\r_0^{**}}}: \exists A_{l,i} \mbox{ with } A_{l,i} \subset A\bigr\}\, .
 \end{align*}
Clearly, we have $|\ca C|<\infty$ and therefore we can enumerate the elements of $\ca C$ by $A_{1,0},\dots,A_{k,0}$.
 Our first intermediate results shows  that for all $i=1,\dots,n$ there exists a $0\leq j<i$ and an
 $l$ such that 
 \begin{align}\label{lemma:existence-of-cluster-peaks-h1}
  A_{1,i}\cup A_{2,i} \subset A_{l,j}\, .
 \end{align}
  To verify \eqref{lemma:existence-of-cluster-peaks-h1}, let $\z:\cc{\mhx {\r_i^{**}}} \to \cc{\mhx {\r_{i-1}^{**}}}$ be the CRM. Since 
  our construction ensures $\r_{i-1}^{**}< \r_i^*$ the definiton of split levels then ensure  
  $\z(A_{1,i}) = \z(A_{2,i})$. We write $B:= \z(A_{1,i})$. If $i=1$, then 
  $A_{1,i} \cup A_{2,i} \subset B \in \cc{\mhx{\r_0^{**}}}$ implies $B\in \ca C$, and hence \eqref{lemma:existence-of-cluster-peaks-h1}
  is true. If $i>1$, then there exists an $A\in \cc{\mhx{\r_0^{**}}}$ with $B\subset A$, and 
  since the CRM property implies $A_{1,i}\cup  A_{2,i}\subset B\subset A$, we conclude that $A\in \ca C$
  and hence \eqref{lemma:existence-of-cluster-peaks-h1}
  is true in this case, too.
  
  In the following, we denote the largest $j$ for which \eqref{lemma:existence-of-cluster-peaks-h1} is true
  by $j_i$ and the corresponding $l$ is denoted by $l_i$. 
    For $1\leq m\leq n$ we further write 
    \begin{align*}
     \mathfrak I_m &:= \bigl\{ A_{l_i, j_i}: 1\leq i\leq m\bigr\} \\
     \mathfrak L_m &:= \bigl\{ A_{l,i}: 1\leq i\leq m, l=1,2\mbox{ and } A_{l,i}\not\in \mathfrak I_m\bigr\}\, . 
    \end{align*}
    Our next intermediate goal is to show that for all $1\leq i_1,i_2\leq m$ and $l=1,2$ the implication
\begin{align}\label{lemma:existence-of-cluster-peaks-h2}
 A_{1,i_1}\subsetneq A_{l,i_2} \qquad \Longrightarrow \qquad  A_{2,i_1}\subset A_{l,i_2}
\end{align}    
    holds. To this end, we note that $A_{1,i_1}\neq A_{l,i_2}$ immediately implies
    $i_1\neq i_2$. Moreover, the case $i_1<i_2$ implies $\r_{i_1}^{**} < \r_{i_2}^{**}$, but 
     but then the assumed $A_{1,i_1}\subset A_{l,i_2}$ gives $A_{l,i_2}\cap A_{1,i_1}\neq \emptyset$
     and thus $\cc{\mhx{\r_{i_2}^{**}}} \comparable \cc{\mhx{\r_{i_1}^{**}}}$ shows 
     the excluded $A_{l,i_2}\subset A_{1,i_1}$. Consequently, we have $i_2<i_1$.
     Let us write $i := i_1-1 \geq i_2$. Then our construction ensures $\r_{i_2}^* <\r_i^{**} < \r_{i_1}^*$
     and by the second inequality we find an $A\in \cc{\mhx{\r_{i}^{**}}}$ with 
     $A_{1,i_1}\cup A_{2,i_1}\subset A$. Moreover, the first inequality gives 
     $\cc{\mhx{\r_{i}^{**}}} \comparable \cc{\mhx{\r_{i_2}^{**}}}$ and therefore 
     $A_{1,i_1} \subset A\cap A_{l,i_2}$ implies $A\subset A_{l,i_2}$.

Our next intermediate goal is to show that for all $1\leq i_1,i_2\leq m$ and $l_1,l_2\in \{1,2\}$ the implication
\begin{align}\label{lemma:existence-of-cluster-peaks-h3}
 A_{l_1,i_1}\subsetneq  A_{l_2,i_2} \qquad \Longrightarrow \qquad A_{l_2,i_2} \in \mathfrak I_m
\end{align}
holds. To this end, we first note that by \eqref{lemma:existence-of-cluster-peaks-h2}
we already know that $A_{1,i_1}\cup  A_{2,i_1}\subset A_{l_2,i_2}$
and hence the definition of $j_{i_1}$ gives $j_{i_1} \geq i_2$.
If $i:= j_{i_1} > i_2$, we have $\cc{\mhx{\r_{i}^{**}}} \comparable \cc{\mhx{\r_{i_2}^{**}}}$
and thus there exists an $A \in \cc{\mhx{\r_{i_2}^{**}}}$ with 
$A_{l,i}\subset A$, where $l:= l_{i_1}$.  Then $A_{1,i_1}\subset A_{l,i}\subset A$ 
and $A_{1,i_1}\subset A_{l_2,i_2}$ implies $A = A_{l_2,i_2}$.
Consequently, we actually have $A_{l,i}\subset A_{l_2,i_2}$, and $i>i_2$ ensures 
$A_{l,i}\neq A_{l_2,i_2}$. In other words, we have $A_{l,i}\subsetneq  A_{l_2,i_2}$,
where our construction ensures $i = j_{i_1}<i_1$, i.e.~we have replaced the index $i_1$ on
the left hand side of \eqref{lemma:existence-of-cluster-peaks-h3} by a smaller value $i$.
Clearly, we can continue this until we have reached $j_i = i_2$.
Let us therefore assume without loss of generality that  
 $j_{i_1} = i_2$. Then, however, we obviously have $l_2 = l_{i_1}$, and hence the desired $A_{l_2,i_2} \in \mathfrak I_m$
immediately follows.

Our final intermediate goal is to show that,  for all $1\leq i_1,i_2\leq m$ and $l_1,l_2\in \{1,2\}$, the implication
\begin{align}\label{lemma:existence-of-cluster-peaks-h4}
 A_{l_1,i_1},  A_{l_2,i_2} \in \mathfrak L_m \mbox { and } A_{l_1,i_1} \neq  A_{l_2,i_2}
 \qquad \Longrightarrow \qquad 
 A_{l_1,i_1} \cap  A_{l_2,i_2} = \emptyset
\end{align}
holds true.
To this end, we first recall that $A_{l_1,i_1} \neq  A_{l_2,i_2}$
implies $l_1\neq l_2$ or $i_1\neq i_2$. Now, if $i_1 = i_2$, then $l_1\neq l_2$
immediately gives $ A_{l_1,i_1} \cap  A_{l_2,i_2} = \emptyset$.
Moreover, if $i_1 \neq  i_2$ we assume without loss of generality that $i_2<i_1$ and $l_1 = 1$.
Then there exists an $A\in \cc{\mhx{\r_{i_2}^{**}}}$ with $A_{1, i_1}\subset A$.
In the case $A\neq A_{l_2,i_2}$ we have $A\cap A_{l_2,i_2} = \emptyset$ and thus 
$ A_{l_1,i_1} \cap  A_{l_2,i_2} = \emptyset$ follows. Moreover, the case $A= A_{l_2,i_2}$
implies $A_{l_1,i_1} \subsetneq  A_{l_2,i_2}$, which is impossible by 
\eqref{lemma:existence-of-cluster-peaks-h3}.

Now observe that by \eqref{lemma:existence-of-cluster-peaks-h4} and 
\emph{i)} each $A\in \mathfrak L_m$ contains a different cluster peak, and hence it 
suffices to show that $n + 1\leq |\mathfrak L_n|$. We will establish the latter by inductively 
showing that $m + 1\leq |\mathfrak L_m|$ holds for all $m=1,\dots,n$.
In the case $m=1$ we have $\mathfrak I_1 = \{A_{l_1, 0}\}$, and hence we have $|\mathfrak L_1| = 2$.
Let us now assume that we already know $m \leq |\mathfrak L_{m-1}|$ for some $2\leq m\leq n$.
Then the definition of $\mathfrak I_m$ ensures $A_{1,m}, A_{2,m} \not\in \mathfrak I_m$.
Consequently, in the case $i_m = 0$  we have 
\begin{align*}
 \mathfrak L_m = \mathfrak L_{m-1} \cup \bigl\{ A_{1,m}, A_{2,m}\bigr\}
\end{align*}
and in the case $i_m > 0$ we have 
$\mathfrak L_m = (\mathfrak L_{m-1}\setminus \{A_{l_m,i_m}\}) \cup \bigl\{ A_{1,m}, A_{2,m}\bigr\}$
by \eqref{lemma:existence-of-cluster-peaks-h3}, \eqref{lemma:existence-of-cluster-peaks-h2}, 
and the definition of $i_m$. In both cases, we thus have $m \leq |\mathfrak L_{m-1}| +1 \leq |\mathfrak L_{m}|$.
\end{proofof}

\begin{lemma}\label{lemma:inf-on-tube-function}
 Let $h:\Rd\to \R$ be continuous with compact support $X:= \overline{\{h\neq 0\}}$, and let
 $A\subset X$ be non-empty. Then the function $H_A:[0,\infty) \to \R$ defined by 
 \begin{align}
  H_A(\d) := \inf_{x\in A\pde} h(x)\, , \qquad\qquad\d\geq 0, 
 \end{align}
is decreasing and the following statements hold true:
\begin{enumerate}
 \item We have $H_{\overline A} = H_A$.
 \item For all $\d_1,\d_2\geq 0$ with $\d_1\neq \d_2$ we have $|H_A(\d_1)-H_A(\d_2)| \leq \om(h, |\d_1 - \d_2|)$.
 \item There exists a $\d_0\geq 0$ such that for all $\d\geq \d_0$ we have $H_A(\d) = \inf_{x\in \Rd} h(x)$.
\end{enumerate}
\end{lemma}

\begin{proofof}{Lemma \ref{lemma:inf-on-tube-function}}
\ada i We first note that for $\d>0$ we have $\overline A\pde = A\pde$.
In addition $\overline B = B\pdex 0$ for any non-empty set $B\subset \Rd$ shows that $\overline A\pdex0 = A\pdex0$.

\ada {ii} By \emph{i)}  we may assume that $A$ is closed. In addition, it suffices to consider
the case  $0\leq \d_1<\d_2$. 
 The definition of $H_A$ then gives $H_A(\d_2) \leq H_A(\d_1)$, and hence it remains to show 
 \begin{align}\label{lemma:inf-on-tube-function-h1}
  H_A(\d_1) - H_A(\d_2)
  \leq 
  \om(h, \d_2 - \d_1) \, .
 \end{align}
To this end, we note that $A\pdex{\d_2}$ is closed and bounded, and hence there exists an $x_2\in A\pdex{\d_2}$
with $h(x_2) = H_A(\d_2)$. In the case $x_2\in A\pdex{\d_1}$, 
we then find $H_A(\d_1) \leq h(x_2)=H_A(\d_2)$. This shows $H_A(\d_2) = H_A(\d_1)$, and hence 
\eqref{lemma:inf-on-tube-function-h1} is true. Let us therefore consider the case $x_2\not \in A\pdex{\d_1}$.
Since $A$ is closed, it is compact, 
and thus there exists an $x_0\in A$ with $\snorm{x_0-x_2}\leq \d_2$. Moreover, the assumed 
$x_2\not \in A\pdex{\d_1}$ 
ensures $\snorm{x_0-x_2} > \d_1$.
We define $x(t) := (1-t) x_0 + tx_2$ for $t\geq 0$ and $t_0 := \d_1 / \snorm{x_2-x_0}$.
A simple calculation then shows 
\begin{align*}
 d(x(t_0), A) \leq \snorm{x(t_0)- x_0} = t_0\,\snorm{x_2- x_0} = \d_1\, ,
\end{align*}
that is $x(t_0) \in A\pdex{\d_1}$. Moreover, $\snorm{x_0-x_2} > \d_1$ gives $t_0<1$, and thus we find
\begin{align*}
 \snorm{x_2 - x(t_0)} 
 = |1-t_0|\, \snorm{x_0 - x_2} 
 = \snorm{x_0 - x_2} - \d_1 
 \leq \d_2  - \d_1\, .
\end{align*}
Combining these considerations, we find 
\begin{align*}
 H_A(\d_1) - H_A(\d_2)
 \leq 
 h(x(t_0)) - h(x_2) 
 \leq 
 \om(h, \d_2  - \d_1)\, .
\end{align*}
and letting $\e\to 0$ then yields \eqref{lemma:inf-on-tube-function-h1}.

\ada {iii} Since $h$ is continuous with compact support, there exists an $x_0\in \Rd$ with 
\begin{align*}
 h(x_0) =  \inf_{x\in \Rd} h(x)\, .
\end{align*}
Moreover, $A$ is assumed to be non-empty, and therefore there exists a $\d_0\geq 0$ with $x_0\in A\pdex{\d_0}$. Combining 
both gives the assertion.
\end{proofof}

\begin{lemma}\label{lemma:min-mutual-dist}
Let $h:\Rd\to [0,\infty)$ be continuous with compact support $X:= \overline{\{h>0\}}$.
 Then, for all  $\r>0$ with $2\leq |\cc{\mhx\r}|<\infty$ and 
 \begin{align}\label{lemma:min-mutual-dist-h1}
  \t^* := \min\bigl\{d(A,A'): A,A'\in \cc{\mhx\r} \mbox{ with } A \neq A'\bigr\}\,.
 \end{align}
 we have $\t^* >0$.
\end{lemma}

\begin{proofof}{Lemma \ref{lemma:min-mutual-dist}}
 By the compactness of $X$ all connected components of $\cc{\mhx\r}$ are compact and hence we have 
$d(A,A')>0$ for all $A,A'\in \cc{\mhx\r}$ with $A\neq A'$. Our assumption 
$|\cc{\mhx\r}|<\infty$ then implies $\t^* >0$.
\end{proofof}

\begin{lemma}\label{lemma:separating-valley-function}
Let $h:\Rd\to [0,\infty)$ be continuous with compact support $X:= \overline{\{h>0\}}$. 
 Moreover, for a   $\r>0$ with $2\leq |\cc{\mhx\r}|<\infty$
 we fix some $A_1,A_2\in \cc{\mhx\r}$ with $A_1\neq A_2$ and define 
 \begin{align*}
  \rda &:=  0 \vee \inf\biggl\{\,  \sup_{x\in S} h(x) \,\,\Bigl|\, S \subset X\mbox{ separates $A_1$ and $A_2$  }    \biggr\}\, .
 \end{align*}
 Then the following statements hold true:
 \begin{enumerate}
  \item There exists an $S\subset X$ that separates $A_1$ and $A_2$, and we have $0\leq\rda < \r$.
  \item For all $\d_1,\d_2\geq 0$ with $\d_1\neq \d_2$ we have $|\hdsa{\d_1}-\hdsa{\d_2}| \leq \om(h, |\d_1 - \d_2|)$.
    \item We have $\rda = \hdsa 0$, and if $\rda>0$, then $\lim_{\d\to \infty} \hdsa \d = \inorm h$.
    \end{enumerate}
\end{lemma}

\begin{proofof}{Lemma \ref{lemma:separating-valley-function}}
Let $\z:\cc {\mhx\r}\to \ccp X$ be the CRM according to 
Lemma \ref{lemma:base-components}. Moreover, we define $X_i := \z(A_i)$
for $i=1,2$, which in turn yields $A_1\subset X_1$ and $A_2\subset X_2$. 

\ada i In the case  $X_1 \neq X_2$, Lemma \ref{lemma:base-components} already shows that $S:= \emptyset$ separates $A_1$ and $A_2$,
and this also gives $\rda = 0<\r$. It thus remains to consider 
the case $X_1= X_2$. To this end, we define $\t^*>0$ by \eqref{lemma:min-mutual-dist-h1}, and 
%
%
%
%
in addition, we write
\begin{align*}
S:= \bigl\{x\in X: 2d(x,A_1) =  \t^*\bigr\} \, .
\end{align*}
Our first intermediate goal is to show that $S$ separates $A_1$ and $A_2$.
To this end, we fix some $x_1\in A_1$ and $x_2\in A_2$ and a continuous $\g:[0,1]\to X$
with $\g(0)= x_1$ and $\g(1) = x_2$.
Next, we define $f:[0,1]\to \R$ by 
$f(t) := d(\g(t), A_1)$ for $t\in [0,1]$. Clearly, $f$ is continuous with $f(0) = 0$
and $f(1) = d(x_2,A_1) \geq \t^*$. Consequently, the intermediate value theorem gives a $t_0\in (0,1)$
with $f(t_0) = \t^*/2$. This implies $\g(t_0) \in S$, i.e.~$S$ does indeed separate $A_1$ and $A_2$.

Our next goal is to show that $\rda < \r$. To this end, we note that $S$ is non-empty by Lemma \ref{lemma:base-components},
and hence it suffices to show that 
$\sup_{x\in S}h(x) < \r$. Here we note that $S$ is compact since $X$
is compact by assumption and $S$ is closed in $X$. Consequently, there exists an $x^* \in S$ with 
$h(x^*) = \sup_{x\in S}h(x)$. To derive the desired $h(x^*) < \r$ it thus suffices to 
prove $S\cap \mhx \r = \emptyset$. Let us assume the converse. Then there exists an
$x\in S\cap \mhx \r$, and consequently also an $A\in \cc{\mhx\r}$ with $x\in A$.
Clearly, $A=A_1$ is impossible since this would imply
$0=d(x,x) \geq d(x,A_1) = \t^*/2 >0$ by the definition of $S$. Moreover, the case $A\neq A_1$ is also impossible, since 
in this case $x\in S$ implies $d(x,A_1) = \t^*/2$, while $x\in A$ implies 
$d(x,A_1) \geq d(A,A_1) \geq \t^*$.
Combining the considerations above  we then find the desired $\rda < \r$. 
Moreover, 
the estimate $\rda \geq 0$ is obvious.

\ada {ii} If $S:= \emptyset$ separates $A_1$ and $A_2$,
we have  $\hdsa \d = 0$ for all $\d\geq 0$, and hence there is nothing to prove. 
In the remaining case, we define
\begin{align*} 
 \tilde H_S(\d) := \inf_{x\in S\pde} -h(x) = - \sup_{x\in S\pde} h(x) \, , \qquad \d\geq 0
\end{align*}
for each $S\subset X$ that separates $A_1$ and $A_2$. Since all these $S$ are non-empty, we then find 
\begin{align}\label{lemma:separating-valley-function-h1}
 \hdsa \d =  \inf\bigl\{\,  - \tilde H_S(\d) \,\,\Bigl|\, S \subset X\mbox{ separates $A_1$ and $A_2$  }    \bigr\} \, , \qquad \d\geq 0.
\end{align}
Let us now assume without loss of generality that $0\leq \d_1<\d_2$. For an arbitrary $\e>0$ we then find 
an $S\subset X$ that separates $A_1$ and $A_2$ and that satisfies $-\tilde H_S(\d_1) \leq \hdsa{\d_1}+\e$.
This gives 
\begin{align*}
 0 
 \leq \hdsa{\d_2} - \hdsa{\d_1}
 \leq - \tilde H_S(\d_2) + \tilde H_S(\d_1) + \e
 \leq \om(h, |\d_1 - \d_2|)+ \e\, ,
\end{align*}
where in the first step we used the monotonicity of $\hdsa\mycdot$ and in the last step we applied 
Lemma \ref{lemma:inf-on-tube-function} to the function $\tilde H_S$.


\ada {iii} 
If $S:= \emptyset$ separates $A_1$ and $A_2$,
we have  $\rda = 0 =\hdsa \d$ for all $\d\geq 0$. In the remaining case, 
we again have \eqref{lemma:separating-valley-function-h1}. 
Moreover, for each $S\subset X$ separating $A_1$ and $A_2$ we find
\begin{align*}
 - \tilde H_S(0) 
 =  - \tilde H_{\overline S}(0)  
 = \sup_{x\in \overline S\pdex 0} h(x)
 = \sup_{x\in \overline S} h(x) 
 = \sup_{x\in S} h(x)\, ,
\end{align*}
where in the first step we applied Lemma \ref{lemma:inf-on-tube-function} to the function $\tilde H_S$ and in the last 
step we used the continuity of $h$.
Combining this with \eqref{lemma:separating-valley-function-h1} and the definition of $\rda$ we then obtain $\rda  =\hdsa \d$.
Finally, if $\rda>0$, then each $S\subset X$ separating $A_1$ and $A_2$ is non-empty. For $\d> \diam X$, we then find 
$X\subset S\pde$ and hence
$\hdsa\d = \inorm h$.
%
%
\end{proofof}

\begin{lemma}\label{lemma:twin-peaks-function}
Let $h:\Rd\to [0,\infty)$ be continuous with compact support $X:= \overline{\{h>0\}}$. 
  Moreover, for a   $\r>0$ with $2\leq |\cc{\mhx\r}|<\infty$
 we fix some $A_1,A_2\in \cc{\mhx\r}$ with $A_1\neq A_2$.
 Then the following statements hold true:
 \begin{enumerate}
  \item We have $\hdra 0 = \r$ and $\hdra \d    < \r$ for all  $\d>0$.
  \item For all $\d_1,\d_2\geq 0$ with $\d_1\neq \d_2$ we have $|\hdra{\d_1}-\hdra{\d_2}| \leq \om(h, |\d_1 - \d_2|)$.
  \item There exists a $\d_0>0$ such that $\hdra\d = 0$ for all $\d\geq \d_0$.
 \end{enumerate}
%
%
%
 \end{lemma}

\begin{proofof}{Lemma \ref{lemma:twin-peaks-function}}
\ada i Since $A_1$ and $A_2$ are closed, we have 
\begin{align*}
  \hdra 0
  = 
  \inf_{x\in (A_1 \cup A_2)\pdex 0} h(x) 
  =
  \inf_{x\in A_1} h(x)      \wedge \inf_{x\in A_2} h(x)
  = \r\, ,
\end{align*}
where in the last step we used Lemma \ref{lem:continuous-boundary}.
Moreover, for $\d>0$, the monotonicity of $\hdra\mycdot$ gives 
$\hdra \d \leq \hdra 0 = \r$.
%
%
Let us now assume that  $\hdra \d  = \r$ was true for some $\d>0$.
By the definition of $\hdra\d$ we then see that
\begin{align}\label{lemma:twin-peaks-function-h1}
 h(x) \geq \r\, , \qquad \qquad x\in A_1\pde \cup A_2\pde 
\end{align}
would be true. For 
 $\t^*>0$  defined by \eqref{lemma:min-mutual-dist-h1},
 there then exists an $\e>0$ with both $\e<\d$ and $\e<\t^*$.
Moreover, 
 by Lemma \ref{lem:continuous-boundary} we find an $x^*\in \partial A_1$, and by
the definition 
of  topological boundaries, there then exists an $x\in B(x^*,\e) \setminus A_1$.
This implies $x\in A_1\pde$, and thus we have $h(x)\geq \r$ by 
\eqref{lemma:twin-peaks-function-h1}. Therefore, there exists an $A'\in \cc{\mhx\r}$ with $x\in A'$, and $x\not \in A_1$ implies 
$A'\neq A_1$. 
Using  $\partial A_1\subset A_1$ we thus obtain
\begin{align*}
 \t^* \leq d(A_1,A') \leq d(x^*,x) \leq \e\, ,
\end{align*}
which contradicts  $\e<\t^*$.
%
%

\emph{ii)} and \emph{iii).} Both assertions follow from Lemma \ref{lemma:inf-on-tube-function} by using
the identity
$\hdra \mycdot = H_{A_1\cup A_2}(\mycdot)$.
\end{proofof}


\begin{lemma}\label{lemma:separation-characterization}
Let $h:\Rd\to [0,\infty)$ be continuous with compact support $X:= \overline{\{h>0\}}$. 
  Moreover, for a   $\r>0$ with $2\leq |\cc{\mhx\r}|<\infty$
 we fix some $A_1,A_2\in \cc{\mhx\r}$ with $A_1\neq A_2$.
Then the following statements are equivalent for $\d>0$, $\e \in (0,1)$, and $\eps \geq 0$:
\begin{enumerate}
 \item The sets 
  $A_1$ and $A_2$ are   $(\d,\e,\eps)$-separated.
 \item We have $\hdsa\d < (1-\e) \hdra\d - \eps$.
%
\end{enumerate}
 \end{lemma}

\begin{proofof}{Lemma \ref{lemma:separation-characterization}}
 \atob i {ii} This implication is trivial, since $\hdsa\d$ is a lower bound on the left-hand side of \eqref{sep-gap}.

 
 \atob {ii} i 
By Lemma \ref{lemma:separating-valley-function}
there exists an $S_0\subset X$ that separates $A_1$ and $A_2$. If $S_0=\emptyset$, then 
we have $\hdsa\d = 0$, and since for this $S_0$, the left-hand side of \eqref{sep-gap}
equals $0$, the assumed inequality $\hdsa\d < (1-\e) \hdra\d - \eps$ implies 
 \eqref{sep-gap}. 
Furthermore, if $S_0\neq \emptyset$, then the definition of $\hdsa\d$ shows  
\begin{align}\label{lemma:separation-characterization-h2}
 \hdsa\d = \inf\biggl\{\,  \sup_{x\in S\pde} h(x) \,\,\Bigl|\, S \subset X\mbox{ separates $A_1$ and $A_2$  }    \biggr\} \geq 0\, .
\end{align}
In addition, by $\hdsa\d < (1-\e) \hdra\d - \eps$ there exists a 
$\varrho>\hdsa\d$ with $\varrho < (1-\e) \hdra\d - \eps$ and from
\eqref{lemma:separation-characterization-h2} we then conclude that 
 there exists an $S\subset X$ that separates $A_1$ and $A_2$ and that satisfies 
 \begin{align*}
  \sup_{x\in S\pde} h(x) \leq \varrho\, .
 \end{align*}
Using $\varrho < (1-\e) \hdra\d - \eps$ we then find \eqref{sep-gap} for this $S$, 
i.e.~\emph{i)} is established.
%
\end{proofof}


\begin{lemma}\label{lemma:separation-persistence}
 Let $h:\Rd\to [0,\infty)$ be continuous with compact support $X:= \overline{\{h>0\}}$. 
 Moreover, for a   $\r>0$ with $2\leq |\cc{\mhx\r}|<\infty$
 we fix some $A_1,A_2\in \cc{\mhx\r}$ with $A_1\neq A_2$. Then the following statements hods true:
 \begin{enumerate}
  \item For all ${\r'}>0$, for which there exists a
  $B\in \ccp {\mhx{\r'}}$ such that  $A_1\cup A_2\subset B$, we have  ${\r'} \leq \hdsa 0$.
 
  \item For all 
 $0<\r' < \r$, for which there are  path connected $B_1,B_2\in \cc{\mhx{\r'}}$
 with $B_1\neq B_2$,  $A_1\subset B_1$, and $A_2\subset B_2$, we have 
 \begin{align*}
  \hdsa \d = \hdsb \d
 \end{align*}
 for all $\d\geq 0$ with $\hdsb\d\leq\r'$.
 \end{enumerate}
\end{lemma}

\begin{proofof}{Lemma \ref{lemma:separation-persistence}}
\ada i  Let us fix an arbitrary $S\subset X$ that separates $A_1$ and $A_2$.
Moreover, let $x_1\in A_1$ and $x_2\in A_2$. By assumption, there then exists a continuous $\g:[0,1]\to B$
such that $\g(0) = x_1$ and $\g(1) = x_2$. Since $S$ separates $A_1$ and $A_2$, we then find 
a $t\in [0,1]$ with $\g(t) = S$, and since we also have $\g(t) \in B\subset \mhx {\r'}$, we conclude that 
$\sup_{x\in S}h(x) \geq h(\g(t))\geq {\r'}$. Taking the infimum over all separating $S$ we thus find 
$\rda \geq {\r'}$, where $\rda$ is defined in Lemma \ref{lemma:separating-valley-function}.
Since in this lemma we have also shown $\rda = \hdsa 0$, the assertion is proven.

\ada {ii}
Since any $S$ that separates $B_1$ and $B_2$ also separates $A_1$ and $A_2$ 
we immediately obtain the inequality $\hdsa \d\leq \hdsb\d$ for all $\d\geq 0$.

To prove the converse inequality we first consider the case, in which
$S:= \emptyset$ separates $B_1$ and $B_2$. In this case we simply have 
$0 \leq \hdsa \d\leq \hdsb\d = 0$ for all $\d\geq 0$, and hence there is nothing left to be proven.
In the remaining case, in which $S:= \emptyset$ does not separate $B_1$ and $B_2$,
we know from Lemma \ref{lemma:base-components} that 
 there is a single 
path connected component $X_0$ of $X$ with $B_1\cup B_2\subset X_0$.
Since this implies $A_1\cup A_2 \subset X_0$ we conclude by Lemma \ref{lemma:base-components} 
that $S:= \emptyset$ does not separate $A_1$ and $A_2$, either.
Let us now assume that 
 $$\hdsa \d< \hdsb\d$$
holds for some $\d\geq 0$ with $\hdsb\d\leq\r'$.
 Since $S:=\emptyset$ does not separate $A_1$ and $A_2$ we then know that 
 there exists a non-empty $S\subset X$ that separates $A_1$ and $A_2$
such that 
\begin{align}\label{prop:separation-continuous-h3}
  \sup_{x\in S\pde}h(x) < \hdsb\d \, .
\end{align}
By the definition of $\hdsb\mycdot$
we then observe that we have found a contradiction as soon as we have shown 
that $S$ 
also separates $B_1$ and $B_2$. 
Therefore, let us now fix some  
$y_1\in B_1$ and $y_2\in B_2$ and a path  $\g:[1,2]\to X$ connecting $y_1$ and $y_2$ in $X$,
i.e.~$\g(1) = y_1$ and $\g(2) = y_2$.
Let us further  fix some $x_1\in A_1$ and $x_2\in A_2$.
Since $x_1,y_1\in B_1$ and $B_1$ is path connected, there then exists a path 
$\g_1:[0,1]\to B_1$ connecting 
$x_1$ and $y_1$ in $B_1$, i.e.~$\g_1(0) = x_1$ and $\g_1(1) = y_1$.
Analogously,  we find a path $\g_2:[2,3]\to B_2$ that connects $y_2$ with $x_2$ in $B_2$, 
i.e.~$\g_2(2) = y_2$ and $\g_2(3) = x_2$.
Consequently, $\tilde \g:[0,3]\to X$ defined by 
\begin{align*}
 \tilde \g(t) :=
 \begin{cases}
  \g_1(t) & \mbox{ if } t\in [0,1]\\
  \g(t) & \mbox{ if } t\in [1,2]\\
  \g_2(t) & \mbox{ if } t\in [2,3]
 \end{cases}
\end{align*}
is a path connecting $x_1$ and $x_2$ in $X$.
Since $S$ separates $A_1$ and $A_2$, there then exists a $t_0\in [0,3]$ with $\tilde \g(t_0) \in S$.
By \eqref{prop:separation-continuous-h3} and our assumption on $\d$ we then find
$h(\tilde \g(t_0)) < \hdsb\d \leq \r'$,
%
and hence we have $\tilde \g(t_0) \not\in B_1\cup B_2$.
However, our construction ensures $\tilde \g(t) \in B_1\cup B_2$ for all $t\in [0,1]\cup [2,3]$, and 
thus we find $t_0 \in (1,2)$. This gives $\g(t_0) = \tilde \g(t_0) \in S$.
 In other words, we have shown that $S$
does separate $B_1$ and $B_2$.
\end{proofof}

\begin{theorem}\label{thm:valley-function-indentifies-split-level}
Let $P$ be a probability measure that satisfies \assx {C1} and let $\rs>0$ be a 
split level of $h$ with witness level $\rss>\rs$. Then for all pairs of 
witness components $(A_1,A_2)$ for $\rs$ at level $\rss$ we have 
\begin{align*}
 \hdsa 0 = \rs \, .
\end{align*}
Conversely, if we have a $\r>0$ and some $A_1,A_2\in \cc{\mhx\r}$ with $\hdsa 0>0$,
then $\rs := \hdsa 0$ is a split level with witnessing level $\r$ and 
$(A_1,A_2)$ is a pair of witnesses for $\rs$ at level $\r$.
\end{theorem}

\begin{proofof}{Theorem \ref{thm:valley-function-indentifies-split-level}}
Let us fix a $\r\in (0,\rs)$ and the   CRM $\z:\cc{\mhx\rss}\to \cc{\mhx\r}$.
Then we have $\z(A_1) = \z(A_2) \in \cc{\mhx\r} = \ccp{\mhx\r}$, and consequently
part \emph{i)} of Lemma \ref{lemma:separation-persistence}  applied to $B:= \z(A_1)$
shows 
%
%
%
%
  $\r \leq \hdsa 0$.
Since this holds for all $\r\in (0,\rs)$, we conclude that $\rs \leq \hdsa0$.

Our next goal is to show the converse inequality. To this end, we fix a
$\r'\in (\rs,\r)$ and consider the CRM $\z: \cc{\mhx\r} \to \cc{\mhx{\r'}}$.
By our assumptions, the sets $B_i := \z(A_i) \in   \cc{\mhx{\r'}}$ are path connected for $i=1,2$,
and  by our assumptions, we also have $B_1\neq B_2$. The CRM property 
further ensures $A_i\subset B_i$, and hence part \emph{ii)} of Lemma \ref{lemma:separation-persistence}
together with Lemma \ref{lemma:separating-valley-function} applied to the level $\r'$
shows 
\begin{align*}
 \hdsa 0 = \hdsb0 < \r'\, .
\end{align*}
Since this holds for all $\r'\in (\rs,\r)$, we find $\hdsa0 \leq \rs$. 

Finally, let us assume that  we have a $\r>0$ and some $A_1,A_2\in \cc{\mhx\r}$ with $\rs := \hdsa 0>0$.
For a fixed $0<\r'<\r$ we consider the CRM $\cc{\mhx\r}\to \cc{\mhx{\r'}}$ and write 
$B_i := \z(A_i)$ for $i=1,2$.
In the case $B_1\neq  B_2$, Lemma \ref{lemma:separating-valley-function}  
in combination with part \emph{ii)} of Lemma \ref{lemma:separation-persistence} then shows 
\begin{align*}
\r' > \hdsb0 = \hdsa 0 =  \rs\, .
\end{align*}
In the case $B_1=  B_2$,  part \emph{i)} of Lemma \ref{lemma:separation-persistence}
yields $\r'\leq \hdsa 0 = \rs$. 
\end{proofof}

\begin{theorem}\label{thm:graph-level-range}
Let $P$ be a probability measure that satisfies \assx {C1} and  $\r>0$ be a level
for which there exists $A_1,A_2\in \cc{\mhx\r}$ with $A_1\neq A_2$. We write 
$\rs := \hdsa 0$, and for 
$\eps\geq 0$ and $c\in (0, 1/2)$ we further define 
\begin{align*}
 \D_\r &:= \bigl\{ \d>0:  \mbox{$A_1$ and $A_2$ are $(\d,2c,\eps)$-separated} \bigr\}\, ,\\
 \d_\r^* &:= \inf \biggl\{\d^*\geq0: \hdra{\d^*} = \frac{\hdsa{\d^*}+\eps}{1-2c}  \biggr\}\, .
\end{align*}
Then we have $\D_\r\neq \emptyset$ if and only if 
 \begin{align}\label{thm:graph-level-range-hxx-proof}
 \r > \frac{\rs + \eps}{1-2c}\, .
\end{align}
Moreover, if \eqref{thm:graph-level-range-hxx-proof} is true, then $\D_\r = (0, \d_\r^*)$ and the infimum 
in the definition of $\d_\r^*$ is attained. 
\end{theorem}

\begin{proofof}{Theorem \ref{thm:graph-level-range}}
Before we begin the proof, let us recall that 
by Lemma \ref{lemma:twin-peaks-function} we know that the function
$\hdra \mycdot$ is continuous and decreasing with $\hdra 0 = \r$,
 and that it satisfies both $\hdra \d < \r$ for all $\d>0$ and 
  $\hdra{\d'} = 0$ for some $\d'>0$. In addition, Lemma \ref{lemma:separating-valley-function}
 shows that $\hdsa \mycdot$ is continuous and increasing
 with $\lim_{\d\to \infty} \hdsa \d = \inorm h$, and Theorem \ref{thm:valley-function-indentifies-split-level}
 ensures $\hdsa 0=\rs$. 
 Consequently, the function $f:[0,\infty)\to \R$
 defined by 
 \begin{align*}
  f(\d) := \hdra{\d} - \frac{\hdsa{\d}+\eps}{1-2c} 
 \end{align*}
is continuous and decreasing and satisfies  $f(0) =  \r - \frac{\rs + \eps}{1-2c}$
as well as $f(\d) < f(0)$ for all $\d>0$ 
and $f(\d'') < 0$
for some $\d''\geq \d'$. 

Let us now fix a $\d>0$. 
Then Lemma \ref{lemma:separation-characterization} shows that we have $f(\d) >0$ if 
and only if $A_1$ and $A_2$ are $(\d,2c,\eps)$-separated. In other words we have 
\begin{align}\label{thm:graph-level-range-hxxx}
 \bigl\{\d>0 : f(\d) > 0\bigr\} = \D_\r\, .
\end{align}
Consequently, if there exists a $\d\in \D_\r$, then we have $f(\d) >0$ and the monotonicity 
of $f$ then gives $f(0)>0$, that is \eqref{thm:graph-level-range-hxx-proof} is true.
Conversely, if \eqref{thm:graph-level-range-hxx-proof} is true, then we have $f(0) >0$ and hence the continuity
of $f$ gives us a $\d>0$ with $f(\d)>0$, that is $\d\in \D_\r$.

For the proof of the final two assertions, we first note that we have 
\begin{align*}
 \d_\r^* = \inf\{\d^*\geq 0: f(\d^*) = 0\}\, .
\end{align*}
Now if \eqref{thm:graph-level-range-hxx-proof} is true, then we have $f(0) >0$ and hence the 
intermediate theorem gives us a $\d^*>0$ with $f(\d^*) = 0$, that is $\{\d^*\geq 0: f(\d^*) = 0\} \neq \emptyset$.
Moreover, this set is closed by the continuity of $f$, and combining both properties 
we easily find 
$\d_\r^* \in \{\d^*\geq 0: f(\d^*) = 0\}$.
Finally, the monotonicity of $f$ gives
\begin{align*}
 \bigl\{\d>0 : f(\d) > 0\bigr\} = (0, \d_\r^*)
\end{align*}
and combining this with \eqref{thm:graph-level-range-hxxx} yields $\D_\r = (0, \d_\r^*)$.
\end{proofof}

\begin{lemma}\label{lem:drstar-is-monotone}
 Let $P$ be a probability measure that satisfies \assx {C1} and let $\rs>0$ be a split level with witnessing level $\rss$.
 Moreover, for $\r_1,\r_2\in (\rs,\rss]$, let 
$(A_{\r_1,1}, A_{\r_1,2})$ and $(A_{\r_2,1}, A_{\r_2,2})$ be pairs of witnesses for $\rs$ at the levels 
$\r_1$ and $\r_2$, respectively. Then, $\d_{\r_1}^*$ and  $\d_{\r_2}^*$ defined 
for  $(A_{\r_1,1}, A_{\r_1,2})$ and $(A_{\r_2,1}, A_{\r_2,2})$ as in Theorem \ref{thm:graph-level-range}, satisfy
\begin{align}\label{drstar-monotone}
 \r_1\leq \r_2
 \qquad \qquad 
 \Longrightarrow
 \qquad \qquad
 \d_{\r_1}^* \leq \d_{\r_2}^*\, .
\end{align}
\end{lemma}

\begin{proofof}{Lemma \ref{lem:drstar-is-monotone}}
 Let us assume that there was some $\r_1\leq \r_2$ with  $\d_{\r_1}^* > \d_{\r_2}^*$.
Then there exists a $\d \in (\d_{\r_2}^*, \d_{\r_1}^*)$, and the definition of $\d_{\r_1}$ in 
combination with the monotonicity of $\hdsax \mycdot{\r_1}$ and $\hdrax \mycdot{\r_1}$ gives.
\begin{align*}
 \frac{\hdsax \d{\r_1} + \eps_n}{1-2c_n}  < \hdrax \d{\r_1}  \, .
\end{align*}
Similarly, we obtain 
\begin{align*}
 \hdrax \d{\r_2} \leq \frac{\hdsax \d{\r_2}  +\eps_n}{1-2c_n}\, .
\end{align*}
However,  
$\r_1\leq \r_2$ implies $\hdsax \d{\r_1} \geq \hdsax \d{\r_2}$ as well as
$\hdrax \d{\r_1} \leq \hdrax \d{\r_2}$ for all $\d>0$ by the very definition of these quantities,
and combining these inequalities we then find a contradiction.
\end{proofof}

%

 \begin{proofof}{Claim \ref{claim1}}
  We first verify $\r_n \leq \rsuf$. To this end, 
  we   note that if there is no $\r$ satisfying 
     \begin{align}\label{rho-range}
  \frac{\rs + \eps_n}{1-2c_n} < \r < \rss
 \end{align}
   and $\dsuf < \d_{\r}^*$
  then we have $\rsuf = \infty$ and hence there is nothing to prove. 
  Let us thus assume that there exists
  a $\r$ satisfying   \eqref{rho-range}
 and $\dsuf < \d_{\r}^*$. Then there also exists a 
  $\d\in [\dsuf,\d_\r^*)$, and by using $\e_n<1/3$ we find
\begin{align*}
 \r_{0,\d,\e_n} 
 =  
 C_d\cdot \frac {k_n}{n  \d^d} \cdot \frac{1+\e_n}{1-\e_n} + \frac {\eps_n}{1-\e_n} 
 < 
 C_d \cdot \frac {k_n}{n  \dsuf^d}  \cdot \frac{2}{1-2c_n}  + \frac {\eps_n}{1-2c_n} 
&=
\frac{\rs+\eps_n}{1-2c_n}\\
&\leq 
\frac{\hdsar {\d_\r^*}+\eps_n}{1-2c_n} \\
&= 
\hdrar {\d_\r^*}\\
&\leq 
\hdrar {\d}\, .
\end{align*}
In other words, we have found $\ca L_n(\r) \neq \emptyset$, that is $\r_n \leq \r$. Taking the infimum over the $\r$ specified above then 
yields $\r_n \leq \rsuf$.

For the proof of the remaining $\rnec \leq \r_n'$
we first note that $\r_n\geq \rss$   implies $\r_n'=\infty$, and hence there is nothing left to prove.
Let us thus consider the case $\r_n < \rss$. Here, we fix an arbitrary $\r\in (\r_n,\rss)$.
Then we have $\ca L_n(\r) \neq \emptyset$ and by 
Theorem \ref{thm:graph-level-range} we thus conclude that  \eqref{rho-range}
is satisfied. Consequently, Theorem \ref{thm:graph-level-range} further shows that 
 \begin{align}\label{alter-Lnr}
  \ca L_n(\r) = \Bigl\{ \r': \exists \d \in (0,\d_\r^*)  \mbox{ such that    $\r' := \frac{\hdrar \d}{1+c_n} \geq \frac{\r_{0,\d,\e}}{1+c_n}$} \Bigr\}\, .
 \end{align}
For all $m\geq 1$, 
there then exists a $\d_m\in (0, \d_\r^*)$ with $\r_{0,\d_m,\e_n} \leq \hdrar {\d_m}$ and $\frac{\hdrar {\d_m}}{1+c_n} \leq \inf \ca L_n(\r) + 1/m$.
This implies
%
\begin{align*}
C_d \cdot \frac {k_n}{n  \d_m^d}
\leq 
 C_d \cdot \frac {k_n}{n  \d_m^d} + \eps_n 
 < 
  \r_{0,\d_m,\e_n} 
  \leq 
  \hdrar {\d_m} 
  \leq
  \r
  \leq \rss\, ,
\end{align*}
that is $\d_m \geq \dnec$. By the definition of $\rnec$, we therefore obtain $\rnec\leq \frac{\hdrar {\d_m}}{1+c_n}$.
Letting $m\to \infty$, we thus find 
%
$\rnec \leq \inf\ca L_n(\r)$ and
taking the infimum over all $\r\in (\r_n,\rss)$ then yields $\rnec \leq \r_n'$.
 \end{proofof}

  \begin{proofof}{Claim \ref{claim:no-false-cluster-guarantee}}
  We first show the identity $\riin =  \r_{0,\diin, \e_n}$. To this end, let us fix a $\d\in (0,\diin)$.
  By the monotonicity of $\om(h,\mycdot)$ we then have $2\om(h,\d) \leq \eps_n$, which in turn gives
  $\riin \leq \r_{0,\d, \e_n}$. Taking the supremum over all such $\d$ then yields 
  $\riin \leq \r_{0,\diin, \e_n}$. To establish the converse inequality, we pick a $\r>\riin$. Then 
  there exists a $\d>0$ satisfying both $2\om(h,\d) \leq \eps_n$ and $\r_{0,\d, \e_n}<\r$.
  This gives $\d \leq \diin$ and thus $\r_{0,\diin, \e_n} \leq \r_{0,\d, \e_n}<\r$. Taking the infimum 
  over all such $\r$ then yields $\r_{0,\diin, \e_n} \leq \riin$.
  
  Now let $\r>\riin$. Then there exists a $\d>0$ with $2\om(h,\d) \leq \eps_n$ and $\r\geq  \r_{0,\d, \e_n}$, and therefore
  part \emph{ii)} of Theorem \ref{thm:CDKL} ensures that $\mhx \r \cap D$ is graph connected in $G_\r$.
  
  Conversely, let us fix a $\r>0$ with $\r<\min\{\rs, \riin\}$.
  Then we first note part \emph{i)} of Theorem \ref{thm:CDKL} 
  only considers levels $\r \geq \r_n \geq \rs$ with corresponding 
  Graphs $G_{\r'}$ with 
   $\r' \geq \r_n' \geq \rnec \geq \r_n \geq \rs$, see Claim \ref{claim1} for the latter lower bounds, 
   and therefore the assumed $\r<\rs$ excludes that part \emph{i)} of Theorem \ref{thm:CDKL} provides 
   a guarantee for $\mhx \r \cap D$. Moreover, $\r<\riin$ immediately shows that 
   part \emph{ii)} of Theorem \ref{thm:CDKL} cannot provide a guarantee, either.
 \end{proofof}

 \begin{proofof}{Claim \ref{claim:consistency}}
  Let us first assume that there exists an $n_\rs\geq n_0$ such that 
  $\eps_n \geq 2\om (h,\dsuf)$ for all  $n\geq n_\rs$.
  Without loss of generality we may additionally assume that we have $c_n \leq 1/6$ and
  $\eps_n \leq \rs /6$ for all $n\geq n_\rs$. Let us fix an $n\geq n_\rs$.
Using $\e_n = c_n$ we then find
\begin{align*}
 \r_{0,\dsuf,\e_n} 
 = C_d\cdot \frac {k_n}{n \cdot \dsuf^d} \cdot \frac{1+\e_n}{1-\e_n} + \frac {\eps_n}{1-\e_n}
 = \frac{\rs}{2}\cdot \frac{1+\e_n}{1-\e_n} + \frac {\eps_n}{1-\e_n}
 \leq 
 \frac 9{10} \cdot \rs\, .
\end{align*}
For all $\r\in [0.9 \cdot \rs,\rs)$, we then have $\r\geq \r_{0,\dsuf,\e_n}$, and part \emph{ii)}
of Theorem \ref{thm:CDKL} thus shows that $A_\r\cap D$ is connected in the graph $G_\r$ at level $\r$, where
$A_\r$ is the unique 
 connected component $A_\r\in \cc{\mhx\r}$ containing $A_1^*$ and $A_2^*$. 
 This gives $\hat \r_n \geq \r$ and by taking the supremum over all $\r\in [0.9 \cdot \rs,\rs)$
 we then find $\hat \r_n \geq \rs$.
Moreover, for $\r\in (\rs,\rss)$ there exists an $n_\r\geq n_0$ such that for all $n\geq n_\r$ we have both
$\dsuf < \d_\r^*$ and $\r> \frac{\rs + \eps_n}{1-2c_n}$. This shows $\rsuf \leq \r$, and hence 
we obtain $\hat \r_n \leq \r_n' \leq \rsuf\leq  \r$ for all $n\geq n_\r$.
Combining the lower and upper bound on $\hat \r_n$ then gives $\hat \r_n \to \rs$.

Let us now assume that Theorem \ref{thm:CDKL} ensures $\hat \r_n \to \rs$ for the smallest split $\rs>0$. 
Then there exists a $\r\in (0,\rs]$ and an $n_\r\geq n_0$ such that  
 Theorem \ref{thm:CDKL} guarantees $\hat \r_n \geq \r$ for all 
$n\geq n_\r$, that is, $A_\r\cap D$ is connected in the graph $G_\r$, where $A_\r\in \cc{\mhx\r}$ is the unique 
component containing both $A_{\rss,1}$ and $A_{\rss,2}$.
Since $\rs$ is the smallest split level
and we assumed
$|\cc {\mhx\r}|=1$ for all $\r\in (0,\rs]$,
the guarantee  for $A_\r\cap D$ 
needs to be ensured by part \emph{ii)} of Theorem \ref{thm:CDKL}, and in this case the guarantee actually holds for 
all $\r'\in [\r,\rs]$ by the formulation of part \emph{ii)}.
Consequently, for fixed $n\geq n_\r$, 
there  exists a $\d_n>0$ with $\eps_n\geq 2\om(h, \d_n)$ and 
$\r_{0,\d_n,\e_n} \leq \r$. This gives
\begin{align*}
 C_d \cdot \frac{k_n}{n \cdot \d_n^d} \leq \r_{0,\d_n,\e_n} \leq \r\leq \rs\, ,
\end{align*}
that is $\d_n^d \geq  C_d \cdot \frac{k_n}{n \cdot \rs} = \dsuf^d/2\geq \dsuf^d /2^d$. This implies 
$\d_n\geq \dsuf/2$, and hence we find the desired $\eps_n\geq 2\om(h, \d_n)\geq 2\om(h, \dsuf/2)$.
 \end{proofof}

\begin{proofof}{Claim \ref{clam1:upper-bound}}
We fix an $n\geq n_0$ and write $\r := \rsufnu := \rs + 4 \om(\dsuf) + 3\eps_n + 6\rs c_n$.
Recalling that for $n\geq n_0$ we have
$c_n < 1/3$, we then find 
\begin{align}\nonumber
  \frac{\hdsar \dsuf +\eps_n}{1-2c_n}
 \leq 
 \frac{\rs + \om(\dsuf) + \eps_n}{1-2c_n}
 &= 
  \rs + \frac{2c_n \rs}{1-2c_n} + \frac{\om(\dsuf) + \eps_n}{1-2c_n}\\ \label{clam1:upper-bound-h1}
 &<
 \rs  + 6\rs c_n + 3\om(\dsuf) + 3\eps_n \, .
\end{align}
Using $\hdsar \dsuf \geq \rs$ we then immediatly obtain $\rsufnu >  \frac{\rs+\eps_n}{1-2c_n}$.
Moreover, we have 
\begin{align*}
 \rs  + 6\rs c_n + 3\om(\dsuf) + 3\eps_n 
 = 
 \r - \om(\dsuf)
  \leq 
 \hdrar \dsuf
\end{align*}
and combining this with \eqref{clam1:upper-bound-h1} gives $\frac{\hdsar \dsuf +\eps_n}{1-2c_n} <  \hdrar \dsuf$,
which in turn implies $\dsuf < \d_\r^*$. Consequently, if we additionally assume $\rsufnu = \r< \rss$, then 
the definition of $\rsuf$ immediately gives the desired $\rsuf \leq \rsufnu$.
\end{proofof}

 \begin{proofof}{Claim \ref{clam1:lower-bound}}
Let us fix an $n\geq n_0$.
If $\rnec = \infty$, there is clearly nothing to prove and hence we now
focus on the case $\rnec < \infty$.
Here, we fix an arbitrary 
$\e>0$.
By the definition of $\rnec$ there then exist
 a 
$\r\in (\r_n,\rss)$ and a $\d\in [\dnec, \d_\r^*)$ such that 
\begin{align}\label{clam1:lower-bound-h5}
 \r' :=  \frac{\hdrar \d}{1+c_n}  \leq \rnec + \e\, .
\end{align}
Using $\d<\d_\r^*$ we then find
\begin{align*}
 \hdsar \d \leq \frac{\hdsar \d+\eps_n}{1-2c_n} <  {\hdrar \d} \leq \r\, ,
\end{align*}
and therefore $\r$ and $\d$ satisfy the assumption \eqref{clam1:lower-bound-h1}.
By employing
$0<(1+c_n)(1-2c_n) \leq 1 -c_n$   we thus obtain 
\begin{align*}
 \frac{\hdrar \d}{1+c_n} 
 > 
 \frac{\hdsar \d+\eps_n}{(1+c_n)(1-2c_n)}
 \geq 
 \frac{\rs + \nu(\d) +\eps_n}{(1+c_n)(1-2c_n)}
 &\geq 
 \frac{\rs}{(1-c_n)} + \nu(\d) +\eps_n\\
 &\geq 
 \rs + c_n \rs + \nu(\d) +\eps_n\, .
\end{align*}
Combining this with \eqref{clam1:lower-bound-h5} yields
$\rs + c_n \rs + \nu(\d) +\eps_n  \leq \rnec + \e$ and letting $\e\to 0$ 
then gives the assertion.

Let us now assume that there exists a $\r\in (\rs,\rss)$ with $\hdsar \d>0$ for all $\d>0$.
We define $\nu:[0,\infty) \to [0,\infty)$ by 
\begin{align*}
 \nu(\d) := \inf\Bigl\{ \hdsax \d{\r'}: \r'\in (\rs,\rss) \mbox{ with } \hdsax \d{\r'} \leq \r'  \Bigr\} - \rs   
\end{align*}
for all $\d\geq 0$. By Theorem \ref{thm:valley-function-indentifies-split-level} and the monotonicity 
of $\hdsar \mycdot$ 
we know that $\nu(\d)\geq 0$, and 
 Condition \eqref{clam1:lower-bound-h1} is satisfied by construction.
Let us show that $\nu(\d) >0$ for all $\d>0$. To this end, we assume the converse, 
that is, there exists a $\d_0>0$ with $\nu(\d_0)= 0$. Since $\nu$ is an increasing function, we may assume without 
loss of generality that $\hdsar {\d_0} \leq \r$ by the continuity of $\hdsar \mycdot$ and 
Theorem \ref{thm:valley-function-indentifies-split-level}.
By our assumption we further know $\hdsar {\d_0}>0$ and hence $\nu(\d_0)= 0$ ensures the existence 
of a $\r'\in (\rs,\rss)$ with $\hdsax{\d_0}{\r'} < \hdsar {\d_0}$.
Now, $\r'\leq \r$ is impossible since in this case
every $S$ separating $A_{\r',1}$ and $A_{\r',2}$ also separates $A_{\r,1}$ and $A_{\r,2}$
and this gives $\hdsar {\d_0} \leq \hdsax{\d_0}{\r'}$. However, 
$\r'> \r$ is also impossible, since in this case 
Lemma \ref{lemma:separation-persistence} with the role of $\r$ and $\r'$ interchanged shows 
$\hdsar {\d_0} = \hdsax{\d_0}{\r'}$.
\end{proofof}

\begin{proofof}{Claim \ref{claim:the-three-guarantees}}
 We first note that our assumption, together with the definition of $\dsuf$ yields
 $K_h \dsuf^\theta\leq 2^{1-\theta} K_h \dsuf^\theta \leq 2\om (h,\dsuf/2) \leq \eps_n$.
  Let us now define $\om(\d) := |h|_\theta \d^\theta$ for all $\d>0$. 
Using $K_h\leq |h|_\theta$ we then find 
\begin{align*}
 4 \om(\dsuf) + 3\eps_n
 \leq 
 4  |h|_\theta \dsuf^\theta + 3\eps_n
 \leq 
 4  \frac{|h|_\theta}{K_h} \eps_n + 3\eps_n
 \leq 
 7  \frac{|h|_\theta}{K_h} \eps_n  \, .
\end{align*}
By Lemmas  \ref{lemma:separating-valley-function} and \ref{lemma:twin-peaks-function}
in combination with  $\om(h,\mycdot) \leq \om(\mycdot)$ we then know that 
 \eqref{disco:cdkl-pos-ass} is satisfied, and therefore Claim \ref{clam1:upper-bound}
 gives 
 \begin{align*} 
 \rsuf 
 \leq 
 \rs + 4 \om(\dsuf) + 3\eps_n + 6\rs c_n  
 &\leq 
  \rs + 7  \frac{|h|_\theta}{K_h}\,  \eps_n + 7 \rs C\, \sqrt d\cdot  \frac{ (\log n)^{3/2}}{\sqrt {k_n}}\\
&\leq 
\rs + 7 \, \biggl( \frac{|h|_\theta}{K_h} +  C\, \rs\, \sqrt d     \biggr) \, \biggl(\eps_n +  \frac{ (\log n)^{3/2}}{\sqrt {k_n}} \biggr)\, ,
\end{align*}
whenever the right-hand side of this inequality is smaller than $\rss$.
The lower bound on  $\rnec  - \rs$ directly follows from Claim \ref{clam1:lower-bound} 
by considering the function $\nu \equiv 0$, which is possible by Theorem \ref{thm:valley-function-indentifies-split-level}.
 
To establish the upper bound for $\riin$, we first observe that 
\begin{align*}
 \inf\bigl\{  \r_{0,\d, \e_n}: \d>0 \mbox{ such that } 2\om(h,\d) \leq \eps_n  \bigr\}
 \leq 
 \inf\bigl\{  \r_{0,\d, \e_n}: \d>0 \mbox{ such that } 2|h|_\theta \d^\theta \leq \eps_n  \bigr\}\, .
\end{align*}
Moreover, the infimum on the right hand side is attained at the $\d_n>0$ satisfying $2|h|_\theta \d_n^\theta = \eps_n$.
A simple transformation then yields
\begin{align*}
 \riin 
 \leq 
 \r_{0,\d_n, \e_n}
 =
  C_d\cdot \frac {k_n}{n  \d_n^d} \cdot \frac{1+\e_n}{1-\e_n} + \frac {\eps_n}{1-\e_n} 
\leq
  2 C_d \cdot (2\,|h|_\theta)^{1/\theta}  \cdot \frac {k_n}{n  } \cdot \eps_n^{- d/\theta}  + 2 {\eps_n}  
\end{align*}
where in the last step we used $\e_n < 1/3$ for all $n\geq n_0$.
Analogously, for the proof of the lower bound, we use that 
 \begin{align*}
 \inf\bigl\{  \r_{0,\d, \e_n}: \d>0 \mbox{ such that } 2K_h \d^\theta \leq \eps_n  \bigr\}
 \leq
 \inf\bigl\{  \r_{0,\d, \e_n}: \d>0 \mbox{ such that } 2\om(h,\d) \leq \eps_n  \bigr\}\, ,
\end{align*}
and for the $\d_n>0$ satisfying $2K_h \d_n^\theta = \eps_n$ we then obtain 
\begin{align*}
 \riin 
 \geq 
 \r_{0,\d_n, \e_n}
 =
  C_d\cdot \frac {k_n}{n  \d_n^d} \cdot \frac{1+\e_n}{1-\e_n} + \frac {\eps_n}{1-\e_n} 
  \geq
   C_d \cdot (2\,K_h)^{1/\theta}  \cdot \frac {k_n}{n  } \cdot \eps_n^{- d/\theta}  +  {\eps_n}  \, .
\end{align*}
With the help of these estimates we can now determine suitable values for $C_{\mathrm{low}}$ and $C_{\mathrm{up}}$.
\end{proofof}

\end{document}